\documentclass[mnsc,nonblindrev]{informs3} %

\OneAndAHalfSpacedXI

\usepackage{natbib}
 \bibpunct[, ]{(}{)}{,}{a}{}{,}%
 \def\bibfont{\small}%

\TheoremsNumberedThrough     %
\ECRepeatTheorems

\EquationsNumberedThrough    %

\MANUSCRIPTNO{MS-0001-1922.65}

\usepackage[utf8]{inputenc}
\usepackage{bm}
\usepackage{bbm}
\usepackage{amsmath}
\usepackage{psfrag,amssymb,thmtools}
\usepackage{mathrsfs}
\usepackage{hyperref}
\hypersetup{colorlinks=true,urlcolor=blue,linkcolor=blue,citecolor=blue}    \usepackage{multirow}
\usepackage{xspace}
\usepackage{url}
\usepackage{dsfont}
\usepackage{natbib}
\usepackage{graphicx,graphics}
\usepackage{subfigure}
\usepackage{wrapfig}
\usepackage{xcolor}
\usepackage{enumitem}
\usepackage{booktabs} 
\usepackage{nicefrac} 
\usepackage{yfonts}
\usepackage[euler]{textgreek}
\usepackage[normalem]{ulem}
\usepackage[most]{tcolorbox}
\usepackage[english]{babel}
\usepackage{makecell}
\usepackage{soul}
\usepackage[textsize=tiny]{todonotes}
\usepackage[toc,page,header]{appendix}
\usepackage{minitoc}
\usepackage[subfigure]{tocloft}%
\usepackage{etoc}
\usepackage{array}
\usepackage{arydshln}

\usepackage[linesnumbered, ruled, vlined]{algorithm2e}
\usepackage{algpseudocode}

\newcommand{\cN}{\mathcal{N}}

\newcommand{\cK}{\mathcal{K}}

\newcommand{\cD}{\mathcal{D}}
\newcommand{\cX}{\mathcal{X}}

\newcommand{\R}{\mathbb{R}}
\newcommand{\N}{\mathcal{N}}
\newcommand{\K}{\mathcal{K}}

\newcommand{\Tra}{^{\sf T}} %
\newcommand{\V}[1]{{\bm{\mathbf{\MakeLowercase{#1}}}}} %
\newcommand{\M}[1]{{\bm{\mathbf{\MakeUppercase{#1}}}}} %
\newcommand{\norm}[1]{\left\lVert#1\right\rVert}

\newcommand{\CCETC}{dynamic matching algorithm}
\newcommand{\jb}[1]{\text{job applicants}}
\newcommand{\app}[1]{\text{applicant}}
\newcommand{\com}[1]{\text{company}}

\usepackage{xcolor}
\usepackage[textsize=tiny]{todonotes}

\newcounter{counter}[section]

\newtheorem{thm}{Theorem}[section]
\newtheorem{lem}{Lemma}[section]

\newtheorem{cor}{Corollary}[section]
\newtheorem{ass}{Assumption}[section]
\newtheorem{defn}{Definition}[section]

\newtheorem{exmp}{Example}[section]

\newtheorem{clm}{Claim}[section]
\newtheorem{rem}{Remark}[section]

\begin{document}

\RUNTITLE{Dynamic Matching Bandit For
Two-Sided Online Markets}

\TITLE{Dynamic Matching Bandit For
Two-Sided Online Markets}

\ARTICLEAUTHORS{%
\AUTHOR{Yuantong Li}
\AFF{Department of Statistics and Data Science, University of California Los Angeles, CA 90025, \EMAIL{yuantongli@ucla.edu}} %
\AUTHOR{Chi-Hua Wang}
\AFF{Department of Statistics and Data Science, University of California Los Angeles, CA 90025, \EMAIL{chi-huawang@ucla.edu}}
\AUTHOR{Guang Cheng}
\AFF{Department of Statistics and Data Science, University of California Los Angeles, CA 90025, \EMAIL{sun244@purdue.edu}}
\AUTHOR{Will Wei Sun}
\AFF{Daniels School of Business, Purdue University, IN 47907, \EMAIL{guangcheng@ucla.edu}}
} %

\ABSTRACT{%
Two-sided online matching platforms are employed in various markets. However, agents' preferences in the current market are usually implicit and unknown, thus needing to be learned from data. With the growing availability of dynamic side information involved in the decision process, modern online matching methodology demands the capability to track shifting preferences for agents based on contextual information. This motivates us to propose a novel framework for this dynamic online matching problem with contextual information, which allows for dynamic preferences in matching decisions. Existing works focus on online matching with static preferences, but this is insufficient: the two-sided preference changes as soon as one side's contextual information updates, resulting in non-static matching. 
In this paper, we propose a dynamic matching bandit algorithm to adapt to this problem. The key component of the proposed dynamic matching algorithm is an online estimation of the preference ranking with a statistical guarantee. 
Theoretically, we show that the proposed \CCETC{} delivers an agent-optimal stable matching result with high probability. In particular, we prove a logarithmic regret upper bound $\mathcal{O}(\log(T))$ and construct a corresponding instance-dependent matching regret lower bound.  
In the experiments, we demonstrate that \CCETC{} is robust to various preference schemes, dimensions of contexts, reward noise levels, and context variation levels, and its application to a job-seeking market further demonstrates the practical usage of the 
proposed method.
}%

\KEYWORDS{Dynamic matching, Contextual bandits, Online decision making, Regret analysis, Two-sided market} %

\maketitle

\section{Introduction}
\label{sec:intro}
Two-sided online matching platforms are utilized in various marketplaces, including college admissions \citep{gale1962college, roth2008deferred}, ride-sharing \citep{lokhandwala2018dynamic, shi2023multiagent}, medical doctor placement \citep{roth1984evolution}, dating markets \citep{gusfield1989stable, knuth1997stable, zap2018swiping}, and job-seeking \citep{mine2013reciprocal, almalis2014content, gugnani2020implicit, vafa2022career}. In modern job matching platforms, the two sides are represented by recruiters and job-seekers. The platform's objective is to recommend job-seekers to recruiters to determine if these recommendations meet the companies' talent demands. Recruiters provide a matching score for each recommended job-seeker, which the platforms use as feedback to enhance their recommendation mechanisms. However, optimizing this recommendation process is significantly complicated by two intrinsic factors: (1) \textit{competing characteristic}—the supply of job seekers and demand from companies create competition within the market; (2) \textit{dynamic and two-sided preferences}—preferences are not static and are two-sided, with recruiters and job-seekers each having their own criteria and preferences. Recruiters' preferences vary based on the dynamic fitness of candidate profiles for current positions. Similarly, job-seekers have fixed preferences regarding potential employers, roles, locations, salaries, and other job-related aspects. These challenges significantly complicate the formulation of an effective dynamic matching problem. The platform must continuously adapt its algorithms and strategies to cater to the changing preferences and the competitive nature of the job market. This adaptation requires a sophisticated understanding of market dynamics and the ability to dynamically adjust recommendations based on online feedback and evolving preferences on both sides of the job market.

The two-sided preference structure has been recently studied in the literature as \textit{static} but not \textit{dynamic}. \cite{liu2020competing} operates under the assumption of static preferences and a one-sided preference structure (the preference from job-seekers to companies is known), which proves impractical in settings with a high volume of job-seekers. In such environments, it becomes prohibitively expensive and time-consuming for recruiters to continually rank job-seekers due to dynamic preferences.%
Similarly, \cite{li2023double} assumes knowledge of a single-sided preference structure and provides an extensive study on \textit{static} complementary preferences, overlooking the dynamic nature of job-matching, such as the changing talent pool. While these prior efforts advance the understanding of matching in static talent market environments and deliver efficient algorithm designs, challenges arise when engineers implement these algorithms in environments with \textit{dynamic preferences}. For instance, as job-seekers regularly update their skills, experiences, and wage expectations, companies dynamically change their preferences over these job-seekers \citep{guo2016resumatcher}.

\begin{figure}
    \centering
    \includegraphics[scale=.37]{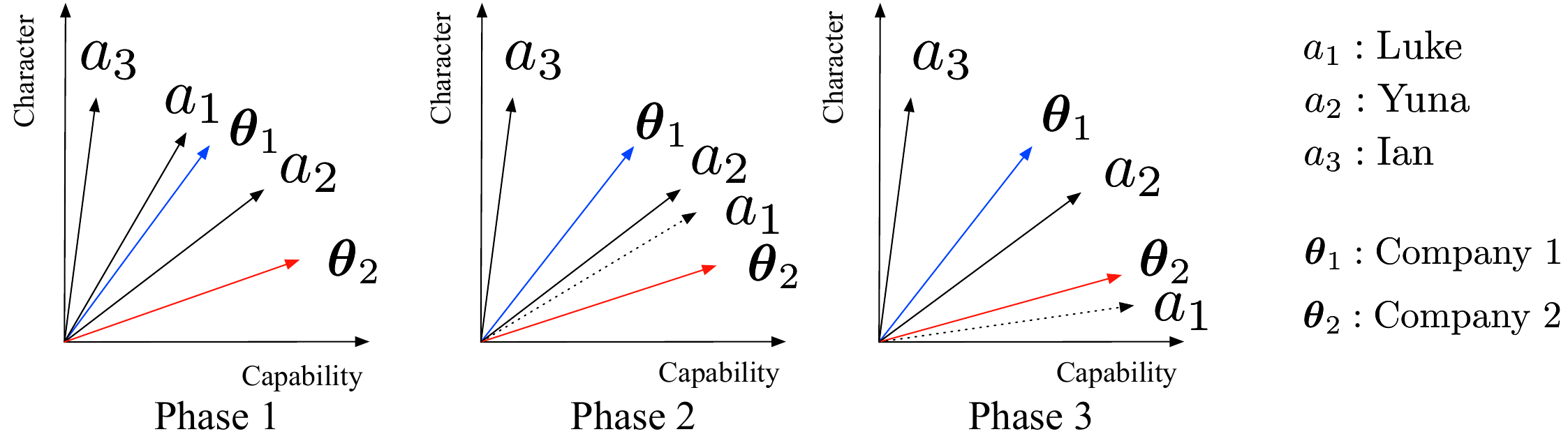}
    \caption{Arm $a_{1}$'s profile changes with an angular velocity, which results in different optimal matching results. Phase 1's optimal matching: (company 1, $a_{1}$), (company 2, $a_{2}$), Phase 2's optimal matching: (company 1, $a_{2}$), (company 1, $a_{1}$), and Phase 3's optimal matching: (company 1, $a_{2}$), (company 2, $a_{1}$).}
    \label{fig:fig_4_cosine_fig}
\end{figure} 

This concept of dynamic preference is illustrated in Figure \ref{fig:fig_4_cosine_fig}. The scenario includes two companies (Company 1 and Company 2) and three job applicants (\(a_{1}, a_{2}, a_{3}\)). The profiles of these job applicants are depicted along two dimensions: capability level (represented on the \(x\)-axis) and character level (on the \(y\)-axis). The true preference parameters of Company 1 and Company 2 are denoted as \(\{\theta_{1}, \theta_{2}\} \in \mathbb{R}^{2}\). The elements within \(\theta_{1}\) and \(\theta_{2}\) represent the respective companies' preference magnitudes for the capability and character traits of the job applicants. It is assumed that all job applicants uniformly prefer Company 1 over Company 2. In this scenario, job applicant \(a_1\)'s profile transitions from Phase 1 to Phase 3, while the profiles of \(a_2\) and \(a_3\) remain unchanged. The preference of a company for a job applicant is determined by the fitness (inner product) \(\langle \theta_i, x_a \rangle\), where \(x_a\) represents the profile of job applicant \(a\) for \(a \in \{1, 2, 3\}\). The higher this fitness, the more preferable the job applicant is to the company. An interesting observation from this example is that as \(a_1\)'s profile updates, the company's preference for job applicants shifts, and correspondingly, the optimal matching changes. Such a dynamic nature of preferences and its impact on optimal matchings highlight the primary challenge in the dynamic online matching market.

The primary goal of the matching platform is to continuously pair companies with the most suitable job applicants, thereby optimizing the overall matching outcome. However, achieving this objective presents a significant challenge: platforms often struggle to accurately estimate companies' true preferences in an ever-changing pool of job applicants. Furthermore, the matching process is complicated by the concept of \textit{noisy bandit feedback}. Specifically, a company only receives a noisy feedback—namely, a noisy observation of the level of satisfaction—from the job applicant with whom it is currently matched, while the counterfactual (other applicants not matched) outcomes remain unobserved \citep{lattimore2020bandit}. This interdependency implies that the feedback received at any given step not only reflects the outcome of the current match but also influences and shapes subsequent matching decisions. This interdependent nature of feedback and decision-making introduces an additional layer of complexity to the dynamic matching process, underscoring the need for adaptive algorithms capable of navigating these complexities effectively.

\subsection{Major Contributions}  

In this study, we leverage a critical observation: the optimality of matching decisions in a dynamic environment depends on the sufficient exploration of two-sided preferences. This insight emerges from an elegant integration of online penalized regression with bandit learning strategies, which aims to achieve optimal matching decisions. This integration leads us to propose a novel two-sided matching algorithm in a dynamic environment. We quantify the uncertainties over learned preference parameters to identify a sufficient exploration horizon that enables us to make optimal matching decisions. Consequently, a successful two-sided matching algorithm will yield optimal decisions once the sample size surpasses this sufficient exploration horizon.

We refer to our novel two-sided online matching algorithm as the dynamic matching algorithm (see Section \ref{section: policy}). The \CCETC{} offers three major advantages: it centralizes all matching decisions within the platform, addresses the continuously changing dynamics in preference learning, and produces optimal dynamic matching decisions. These attributes ensure the validity and robustness of our algorithm in practical two-sided matching scenarios. Theoretically, we establish an upper bound on agent regret and a corresponding theoretical lower bound in a two-agent and three-arms scenario to demonstrate the optimality of our algorithm. Experimentally, we evaluate the performance of \CCETC{} using both synthetic and real datasets.

In summary, our work advances the algorithmic matching literature with the following three major methodological contributions:

\begin{enumerate}
\item Conceptually, we formulate the two-sided online matching problem as a Dynamic Matching Problem (DMP) (see Section \ref{sec: 2-problem}). The DMP encapsulates the ever-changing nature of the talent pool in the job-matching market (see Figure \ref{fig:fig_4_cosine_fig}) and highlights the intrinsic challenges associated with preference learning in dynamic recommendation environments.

\item Methodologically, we introduce a novel \CCETC{} (see Section \ref{section: policy}, Algorithm \ref{algo:CC-ETC}) that addresses the DMP through a bandit algorithm design. The dynamic matching algorithm initially estimates the dynamic preferences for agents (companies) using a penalized statistical estimation method to construct complete ranking lists over arms (job-seekers). After collecting these rankings, the platform employs the classic deferred-acceptance (DA) algorithm \citep{gale1962college} to provide the matching object for all participants (agents and arms). The design of our multi-agent dynamic matching algorithm extends the single-agent bandit algorithm framework \citep{lattimore2020bandit}.
Furthermore, we demonstrate that existing online matching bandit algorithms based on the Upper Confidence Bound (UCB) approach fail in the DMP context and suffer from a linear regret (see Figure \ref{fig: ucb incapable}), due to the non-shrinking upper confidence bounds for specific arms inherent in the dynamic matching problem's characteristics. This phenomenon is further demonstrated through a simple simulation example (Section \ref{sec: ETC vs UCB}). Our algorithm circumvents this issue by employing a theoretically-guided optimal exploration sample size.

\item Empirically, we demonstrate that our algorithm exhibit robustness across diverse arm-to-agent preference uncertainties, in scenarios with rapid temporal changes, preference structures, contextual dimensions, and participant sizes in Section \ref{sec: simulation}. Furthermore, \CCETC{} also showcases its versatility and practical applicability in a dynamic and complex real-world job market, utilizing LinkedIn data, as discussed in Section \ref{sec: real data}.
\end{enumerate}

In addition to the methodological contributions listed above, we also discuss our theoretical contributions in the following:

\begin{enumerate}    
    \item \textbf{Connection Between Statistical Learing and DMP.} 
    In Claims \ref{claim: ranking condition} and \ref{claim: estimation condition} of Section 
    \ref{sec: zero regret}, we show that a fully correct ranking or an unbiased estimation of the preference parameter are the sufficient conditions to achieve an agent-optimal matching. Our work is the first to elucidate the roles that build the bridge between the statistical learning method and the DMP. Additionally, we introduce a novel conceptualization of the DMP as essentially a dual-layered mixture of ranking and estimation challenges in Section \ref{sec: foundation of DMP}.

    \item \textbf{Stable Matching.} We prove the matching stability of the \CCETC{} at each time step with high probability, as highlighted in Theorem \ref{thm: stability of ccetc}. A key characteristic is that at any given moment, and with a complete ranking list available, no participant shows a willingness to deviate from the current recommended matching assigned by \CCETC{} in favor of another participant. This aspect of matching stability is crucial in the dynamic matching problem, as it underscores the efficacy and robustness of the algorithm in maintaining satisfactory recommended matchings throughout the matching process.  

    \item \textbf{Regret Upper Bound.} We establish that the \CCETC{}  achieves a logarithmic expected cumulative regret over time \(T\) (Corollary \ref{coro: upper order}). We show that the complexity of the dynamic matching problem is directly proportional to the job-seeker feature dimension, number of participants, and matching feedback noise level, and inversely proportional to the gap between different job-seekers. Achieving this regret upper bound presents considerable challenges due to the time-variant dynamic preferences, which makes our proof more complex compared to scenarios with fixed preferences between agents and arms over time \citep{liu2020competing}.
    To navigate this regret upper bound, we employ novel non-asymptotic concentration results based on the online penalized regression \citep{li2021online} to quantify the union-bound of probability of ``invalid ranking" (Lemma \ref{lem: invalid prob}).

     \item \textbf{Instance-Dependent Regret Lower Bound.} We utilize a two-agent, three-arm example to explore the instance-dependent regret lower bound. Specifically, we decompose the instance-regret based on the correctness of other agents' rankings and evaluate the probabilities of correct and incorrect ranking events (Section \ref{sec: theory-lowerbound}). By analyzing these events, we can assess the regret on a case-by-case basis and aggregate the regret lower bound across all six identified cases. This lower bound analysis  indicates that the logarithmic regret bound achieved by our dynamic matching algorithm is tight.

\end{enumerate}

\subsection{Related Work}
\label{sec2: related works}

Our work advances the study of preference-based two-sided market matching, and bandit exploration policy design. 

\textbf{Matching in Two-Sided Markets}. 
We first discuss the matching in discrete and continuous two-sided markets when the preference from both sides are known to the platform.
\cite{gale1962college} studied the two-sided matching markets as a pioneer and proposed the deferred-acceptance algorithm (also known as the GS Algorithm), which achieved the stable matching. 
This algorithm \citep{roth2008deferred} has been widely used to match hospitals with residents \citep{roth1986allocation} and students with public schools \citep{atila2005new, abdulkadirouglu2005boston} in New York City and Boston. They focused on discrete two-sided matching models without money transfer. 
\cite{kojima2018designing, nguyen2019stable,aziz2021matching, li2023double} focused on the two-sided market with side constraint, e.g., different races should have the same admitting proportions in the college admission.
However, these results assume that preferences from both sides are known to the platform, which is fundamentally different from our setting, where agents on the one side of the market's preferences are \textit{unknown} and need to be learned through historical interactions.

In practice, there usually exists a centralized platform helping agents to match with each other and one side preference is unknown.  To tackle this problem in the bandit framework, researchers transform the matching objects into bandit notation and assume that one side of market participants can be represented as agents (preferences are unknown) and the other side participants of the market can be viewed as arms (preferences are known), and transform this problem into a  bandit matching problem.
\cite{liu2020competing, liu2021bandit} stand out as pioneering works that delve into scenarios where agents must acquire their preferences through bandit techniques within both centralized and decentralized platforms. %
\cite{jagadeesan2021learning} considers that both sides' preferences are represented by utility functions over contexts where the matching is accompanied by monetary transfers. 
Their objective is different from ours which is defined with respect to the subset instability. 
\cite{sankararaman2020dominate,sarkar2021bandit,dai2021learningA,kong2022thompson, maheshwari2022decentralized, li2023double} considered the bandit matching problem in either centralized or decentralized markets. %
However, these works do not consider the arms' dynamic contextual information and hence are not capable of tackling the dynamic matching problem. \cite{muthirayan2023competing, ghosh2024competing}  explore the bandit matching model similar to DMP under the assumption of non-stationary mean matching scores, but with limited variability.
\cite{cen2022regret} even considers the case when both users and providers do not know their true preferences a priori and incorporate costs and money transfers among agents to faithfully model the competition among agents and discuss the fairness in the matching. In addition, 
\cite{min2022learn} considers the uncertain utility of matching two agents in the episodic reinforcement learning setting.

Research in operations, such as \citep{hu2015dynamic, ma2020algorithms, kalvit2022dynamic, shi2022optimal,ma2023fairness} examine online matching markets to optimize rewards and manage costs. Our approach differs significantly by considering two-sided preferences with one side unknown, unlike their model which involves one-sided, observable preferences with a known or noisy reward matrix. Additionally, while their studies focus on uncertainties from the random future supply and demand affecting time-dependent rewards and costs, our model introduces randomness through noisy matching feedback.

\textbf{Bandit Exploration Strategy}. 
Bandit algorithms \citep{lattimore2020bandit} and reinforcement learning \citep{sutton2018reinforcement}
are modern strategies to solve sequential decision making problems. They have received attentions in business and scientific applications including dynamic pricing \citep{chen2022privacy,wang2023online, miao2023personalized, varma2023dynamic}, online decision making \citep{yuan2021marrying, chen2021statistical, shi2022statistical, baek2023ts, shi2023dynamic}, dynamic treatment regimes \citep{luckett2019estimating, qi2020multi}, and revenue management \citep{aouad2022nonparametric, cheung2022inventory}. Our work also utilizes bandit exploration strategy to handle the unknown preference. However, our work focuses on a fundamentally different dynamic matching problem in two-sided markets.

\noindent\textbf{Notations}.
We denote $[N] = [1,2,..., N]$. Define the capital $X \in \R^{d}$ be the $d$-dimensional random vector. Let $x \in \R^{d}$ represents a $d$-dimensional vector, $x^{(r)}$ represents the $r$-th element of vector $x$, and the bold $\M{X}\in \R^{d\times d}$ represents a real valued matrix. Let $\M{I}_{d} = \text{diag}(1,1,...,1) \in \R^{d\times d}$ represent a $d\times d$ diagonal identity matrix. Denote $\left\lceil x \right\rceil$ as the minimum integer greater than $x$. We denote $T$ as the time horizon.

\section{Dynamic Matching Problem}
\label{sec: 2-problem}

This section formulates the Dynamic Matching Problem (DMP).

\subsection{Environment}
\label{sec: Problem formulation}
We use matching of job applicants and companies as the running example throughout the paper. There are three primary roles in this environment: the organizer (recommendation platform), job applicants, and companies. The goal of the organizer is to recommend the optimal job applicant to companies within this dynamic, online, competitive environment.
We begin by introducing three essential elements in the DMP.

\textsc{(I) Participants.} In this centralized platform, there are $N$ companies (agents) denoted by $\mathcal{N} = \{p_1, p_2,..., p_N\}$, and $K$ job applicants (arms) denoted by $\mathcal{K} = \{a_1, a_2, ... a_K\}$. We assume that the number of companies ($N = |\N|$) is fewer than the number of job applicants ($K = |\K|$).\footnote{Here we also allow job applicants joining and leaving. It is important to note that these job applicants are not static entities within this platform; their composition may vary over time.} %

\textsc{(II) Two-sided Preferences.}  
For DMP, there are two types of preferences: arms to agents' preferences, and agents to arms' preferences.

\textit{Arms to agents' fixed and known preference \(\pi:\mathcal{K}\mapsto\mathcal{N}\)}: We assume that there exist fixed preferences from job applicants to companies, and these preferences are known to the centralized platform. For instance, job applicants are typically required to submit their preferences for different companies via the platform. Let $\pi_{j,i} \in [N]$ represent the ranking for company $p_i$ from the perspective of job applicant $a_j$, and $\pi_{j} = \{\pi_{j,1}, ..., \pi_{j,N}\}$ denote the complete set of company rankings for arm $a_j$. Here, $\pi_{j}$ is a permutation of $[N]$, and without loss of generality we assume that there are no ties in rankings. The shorthand notation $p_i >_j p_{i'}$ indicates that job applicant $a_j$ prefers company $p_i$ over company $p_{i'}$. This known arm-to-agent preference is a mild and common assumption in current online matching literature \citep{liu2020competing, liu2021bandit, li2023double}.

\textit{Agents to arms' dynamic and unknown preference $r(t):\mathcal{N}\mapsto\mathcal{K},  t \in [T]$}. Preferences from companies to job applicants are dynamic and are unknown to the platform due to the large scale of $K$. 
Denote $r_{i,j}(t)$ as the ranking for the job applicant $a_{j}$ from the perspective of company $p_{i}$ and $r_{i}(t) = \{r_{i,1}(t), ..., r_{i,K}(t)\}$ represents the ranking for all \jb{} at time $t$ which is a permutation of $[K]$.
We again assume that there are no ties in rankings.
The notation $r_{i,j}(t) < r_{i,j'}(t)$ indicates that company $p_i$ prefers job applicant $a_j$ over job applicant $a_{j'}$ at time $t$. Similarly, $a_j >_i^t a_{j'}$ means that at time $t$, company $p_i$ prefers job applicant $a_j$ over job applicant $a_{j'}$. The key distinction between the DMP and existing two-sided matching literature \citep{gale1962college, liu2020competing, liu2021bandit, li2023double} is that $\{r_{i}(t)\}_{i \in [N]}$ are both unknown and dynamic.

\textsc{(III) Stable Matching and Optimal Matching.} We introduce several key concepts in the two-sided matching field \citep{roth2008deferred}. 

\begin{defn}[Blocking]
\label{def: blocking}
   A matching $m$ is \textit{blocked by agent} $p_{i}$ if $p_{i}$ prefers being single to being matched with $m(p_{i})$, i.e. $p_{i} >_{i} m(p_{i})$. A matching $m$ is \textit{blocked by a pair of agent and arm} $(p_{i}, a_{j})$ if they each prefer each other to the partner they receive at $m$, i.e. $a_{j} >_{i} m(p_{i})$ and $p_{i} >_{j} m^{-1}(a_{j}).$
\end{defn}

\begin{defn}[Stable Matching]
A matching $m$ is stable if it isn't blocked by any individual or pair of agent and arm applicant.
\end{defn}

Stable matching in a two-sided market ensures that no pair of agent and arm prefers another partner over their current match. This stability is crucial because it fosters efficiency and reduces costs, leading to more satisfied participants and a robust marketplace. (1) \textit{Efficiency} is achieved as all participants are optimally matched, with no blocking pairs present, ensuring that no participant can improve their situation without disadvantaging others. (2) \textit{Reduced transaction costs} arise because stable matching prevents the need for repeated re-negotiations, saving time, effort, and resources. Consequently, stability contributes to the smooth and efficient operation of matching markets, providing predictable and cost-effective outcomes for all involved. 

To account for the potential non-uniqueness of stable matching, we introduce further definitions to delineate agent-optimal matching:

\begin{defn}[Valid Match]
With true preferences from both sides, arm $a_{j}$ is called a \textit{valid match} of agent $p_{i}$ if there exist a stable matching according to those rankings such that $a_{i}$ and $p_{j}$ are matched.
\end{defn}

\begin{defn}[Agent-Optimal Match]
Arm $a_{j}$ is an \textit{optimal match} of agent $p_{i}$ if it is the most preferred valid match.  
\end{defn}
 
Given true preferences from both sides, the DA algorithm shown in Appendix \ref{app-sec:da algo} \citep{gale1962college} provides a stable matching and is always optimal for members of the proposing side. 
We use $\overline{m}_{t}(i)$ to represent the \textit{agent-optimal matching arm} for agent $p_{i}$ and $\overline{m}_{t} = \{\overline{m}_{t}(1), ...., \overline{m}_{t}(N)\}$ represent the agent-optimal matching from $\N$ to $\K$ at time $t$.

\subsection{Matching Protocol}
At time $t$, the platform recommends a job applicant $a_j$ from $\mathcal{K}$ for company $p_i$ according to the current matching policy $m_t(\cdot)$. This recommendation is based on the contextual information $x_j(t) \in \mathbb{R}^d$ of the job applicant $a_j$,  which may include demographics, geography, or capabilities, etc. In response, company $p_i$ evaluates the recommended arm $a_{j}$ by providing a \textit{noisy matching score} $y_{i,j}(t)$ written as:
\begin{equation}
\label{eq: linear model}
    \begin{aligned}
        y_{i,j}(t) = \mu_{i,j}(t) + \epsilon_{i,j}(t), \forall i \in [N], j \in [K], t \in [T],
    \end{aligned}
\end{equation}
where $\mu_{i,j}(t) = \theta_{i, *}\Tra x_{j}(t)$ represents the \textit{true matching score}, $\epsilon_{i,j}(t)$ is a subgaussian noise (Assumption \ref{ass: noise ass}), and $\theta_{i, *} \in \mathbb{R}^d$ denotes the \textit{true preference parameter} for company $p_i$, indicating preference priority across different contexts. Additionally, for company $p_i$, we define $\overline{\Delta}_{i,j}(t)$ as the \textit{score gap} between the optimal matching arm $ \overline{m}_{t}(i)$ and the currently recommended arm $a_{j}$ at time $t$:
\begin{equation}
\label{eq: score gap}
    \overline{\Delta}_{i,j}(t) =\mu_{i, \overline{m}_{t}(i)}(t) - \mu_{i, j}(t).
\end{equation}
In contrast to single-agent bandit problems, where the score gap is always positive, in our dynamic matching problem, this score gap can be positive, negative, or even zero.
Detailed discussion of this gap can be found in Section \ref{sec: zero regret}.

\textsc{Regret.} 
Based on model (\ref{eq: linear model}), we define the \textit{agent-optimal regret} for $p_{i}$ as
\begin{equation}
\label{eq: gap}
    R_{i}(T) = \sum_{t=1}^{T}\mu_{i,\overline{m}_{t}(i)}(t) - \mu_{i,m_{t}(i)}(t).
\end{equation}
This agent-optimal regret represents the difference between the capability of a policy $\V{m}(i) \overset{\Delta}{=} \{m_{1}(i), m_{2}(i),..., m_{T}(i)\}$ in hindsight and the agent-optimal stable matching oracle policy $\overline{\V{m}}(i) \overset{\Delta}{=} \{\overline{m}_{1}(i), \overline{m}_{2}(i),..., \overline{m}_{T}(i)\}$.

\textsc{Social Welfare Gap}. We define social welfare gap as the sum of the absolute value of agent-optimal regret $R_{i}(T)$ across all agents,
$$\textsc{Social Welfare Gap} = \sum_{i=1}^{N}|R_{i}(T)|.$$
It indicates the difference between the total optimal matching score that could have been achieved under ideal conditions and the actual outcome achieved under the current strategy. Since in DMP, social welfare gap is always non-negative and is easier to compare among different policies, which can be used to provide crucial insights into the efficiency of the matching process.

\section{Challenges and Resolutions}

The challenges of the DMP stem from the ever-changing contextual information of job-seekers, which lead to dynamic preferences. To accurately evaluate these dynamic preferences, the platform must learn from historical data, influenced by the policy it employs. An ideal matching algorithm should effectively balance the trade-off between exploring these contextual information and exploiting them to minimize the agent-optimal regret.

\subsection{Pitfall: Incapable Exploration of UCB in DMP}
\label{sec: ETC vs UCB}
Upper Confidence Bound (UCB) method \citep{lattimore2020bandit} is a widely used exploration method used in existing online decision making tasks. 
In this subsection, we demonstrate why directly applying UCB to balance exploration—by adaptively shrinking the upper confidence bound to quickly find the optimal arm—and exploitation—by frequently pulling the optimal arm to minimize the agent-optimal regret—is infeasible in our dynamic matching problem.
We show that centralized UCB suffers a \textit{linear} agent-optimal regret in the following DMP example.

Let $\cN = \{p_{1}, p_{2}, p_{3}\}$ and $\cK = \{a_{1}, a_{2}, a_{3}\}$, with true preferences at time $t$ given below:
\begin{figure}
    \centering
    \includegraphics[scale=0.5]{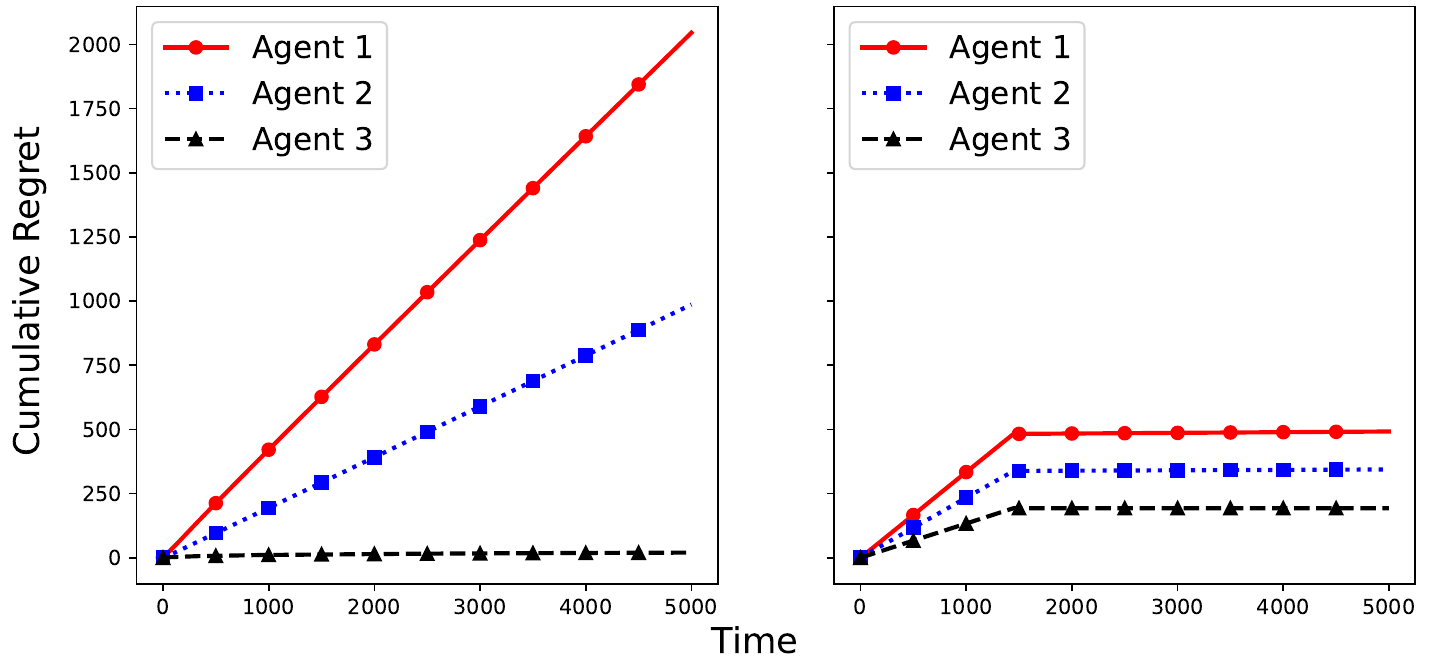}
    \caption{Left: UCB algorithm, Right: our algorithm. Incapable exploration of UCB method.}
    \label{fig: ucb incapable}
\end{figure}
\begin{equation*}
\begin{aligned}
    &p_{1}: a_{1} > a_{2} > a_{3}   \hspace{2cm} a_{1}: p_{2} > p_{3} > p_{1}\\
    &p_{2}: a_{2} > a_{1} > a_{3}   \hspace{2cm} a_{2}: p_{1} > p_{2} > p_{3}\\
    &p_{3}: a_{3} > a_{1} > a_{2}   \hspace{2cm} a_{3}: p_{3} > p_{1} > p_{2}.\\
\end{aligned}    
\end{equation*}
\noindent{}
Based on the above preference design, the agent-optimal stable matching is $(p_{1}, a_{1})$, $(p_{2}, a_{2})$, $(p_{3}, a_{3})$. However, if the platform  wrongly estimates $p_{3}$'s preference as $a_{1} > a_{3} > a_{2}$ based on the UCB estimator, the output stable matching is $(p_1, a_{2}), (p_{2}, a_{1}), (p_{3}, a_{3})$.
As a result, $p_{1}$ and $p_{2}$ suffer positive regrets since their optimal matching arms are $a_{1}$ and $a_{2}$.
In this case, $p_{3}$ will never have the opportunity to correct its mistake $a_{1} > a_{3}$, as it will never be matched with $a_{1}$ where arm $a_{1}$ has a higher upper confidence bound. \textit{Therefore, the upper confidence bound for $a_{1}$ will never shrink, maintaining the preference $a_{1} > a_{3}$}. Consequently, this leads to $p_{1}$ and $p_{2}$ experiencing linear regrets. We empirically demonstrate this phenomenon in Figure \ref{fig: ucb incapable} and the detailed setting is available Section \ref{app: ucb vs ts} in the appendix.

However, as shown in Figure \ref{fig: ucb incapable}, our algorithm to be introduced in Section \ref{section: policy} can avoid this situation through a dedicated design to balance the exploration and exploitation.
The advantage of our algorithm is that it can utilize the historical matching data to acquire a good estimate of $\theta_{i}^{*}$ and $r_{i}(t)$ with a high probability (Lemma \ref{lem: invalid prob}).

\begin{rem}
The above example illustrates that the mechanism to achieve the optimal matching within the DMP is fundamentally different from the single agent bandit problem since the best fitness (optimal) matching arm is not always the top-1 arm (with the highest matching score) for agent due to the competitive characteristics.
\end{rem}

Based on the previous finding, our goal is to design a matching policy $\{m_{t}(i)\}_{i=1, t=1}^{N, T}$ recommending arms for agents. It appears essential for our algorithm to (i) \textit{learn} the true agent-specific preference parameter $\theta_{i,*}$ to uncover the underlying true preference model, and (ii) \textit{design} an exploration strategy based on bandit matching feedback. This strategy efficiently explores potential matching pairs by extracting dynamic ranking information, thereby assisting the algorithm in minimizing agent-optimal matching. Specifically, we have the following two challenges.

\subsection{Challenge 1: Dynamic Preference Learning} 
Learning companies' preferences given dynamic \jb{}' profiles is challenging since there are  numbers of possible matchings between companies and \jb{}. Recovering true preference parameters from noisy matching scores requires modeling the relationship between companies and \jb{}. We resolve this challenge by considering the parametric model \eqref{eq: linear model} to capture the relationship between the matching score and \jb{}' profiles. 
Therefore, the main task becomes estimating the underlying preference parameter by adaptively and sequentially conducting matching experiments to have a good statistical property of these estimators.
Such an estimate is important for inferring a true preference scheme and informing future matching decisions.

\subsection{Challenge 2: Bandit Feedback} 
The platform also needs to balance the exploration (collecting enough job applicants' profiles and companies' matching information) to estimate companies' true preference parameters and the exploitation (providing the optimal matching for companies) tradeoff at each matching time point. Compared to the single-agent bandit problem, the multi-agent competing matching problem is more challenging since the platform needs to handle the multi-agent exploration and exploitation simultaneously. We resolve this challenge by using a new dynamic matching algorithm to balance the multi-agent exploration-exploitation trade-off.

\section{Dynamic Matching Algorithm}
\label{section: policy}
In this section, we propose the dynamic matching algorithm to learn all agents' preference parameters $\{\theta_{i,*}\}_{i=1}^{N}$ and to minimize agent-optimal regret $R_{i}(T)$. 
The \CCETC{} functions as an online statistical estimation method, which achieves optimal matching at most of time. This characteristic underscores the algorithm's efficacy in balancing the trade-offs between estimation accuracy and sample efficiency within dynamic matching problem.

Dynamic matching algorithm includes two major steps, the \textit{learning step}, and the \textit{exploitation step}. 
In the learning step, the platform recommends $a_{j}$ to $p_{i}$ randomly.
After the learning step ends, platform estimates agents' preference parameters $\{\theta_{i,*}\}_{i=1}^{N}$, constructs estimated preference ranking $\{\widehat{r}_{i}(t)\}_{i=1}^{N}$, and collects arms preference $\{\pi_{j}\}_{j=1}^{K}$ in Stage 2 of Figure \ref{fig: algorithm flow}. Then the platform operates the DA algorithm \ref{algo:gs-algo} in the appendix with previous estimated preference ranking in Stage 3 of Figure \ref{fig: algorithm flow} to recommend arms to agents in Stage 4 of Figure \ref{fig: algorithm flow}. Finally, agents provide matching score $\{y_{i,j}(t)\}_{i=1, j=1}^{N,K}$ to the platform in Stage 5 of Figure \ref{fig: algorithm flow}. The detailed \CCETC{} is summarized in Algorithm \ref{algo:CC-ETC}. Below we discuss these two major steps in details.

\begin{figure}
    \centering
    \includegraphics[scale=0.25]{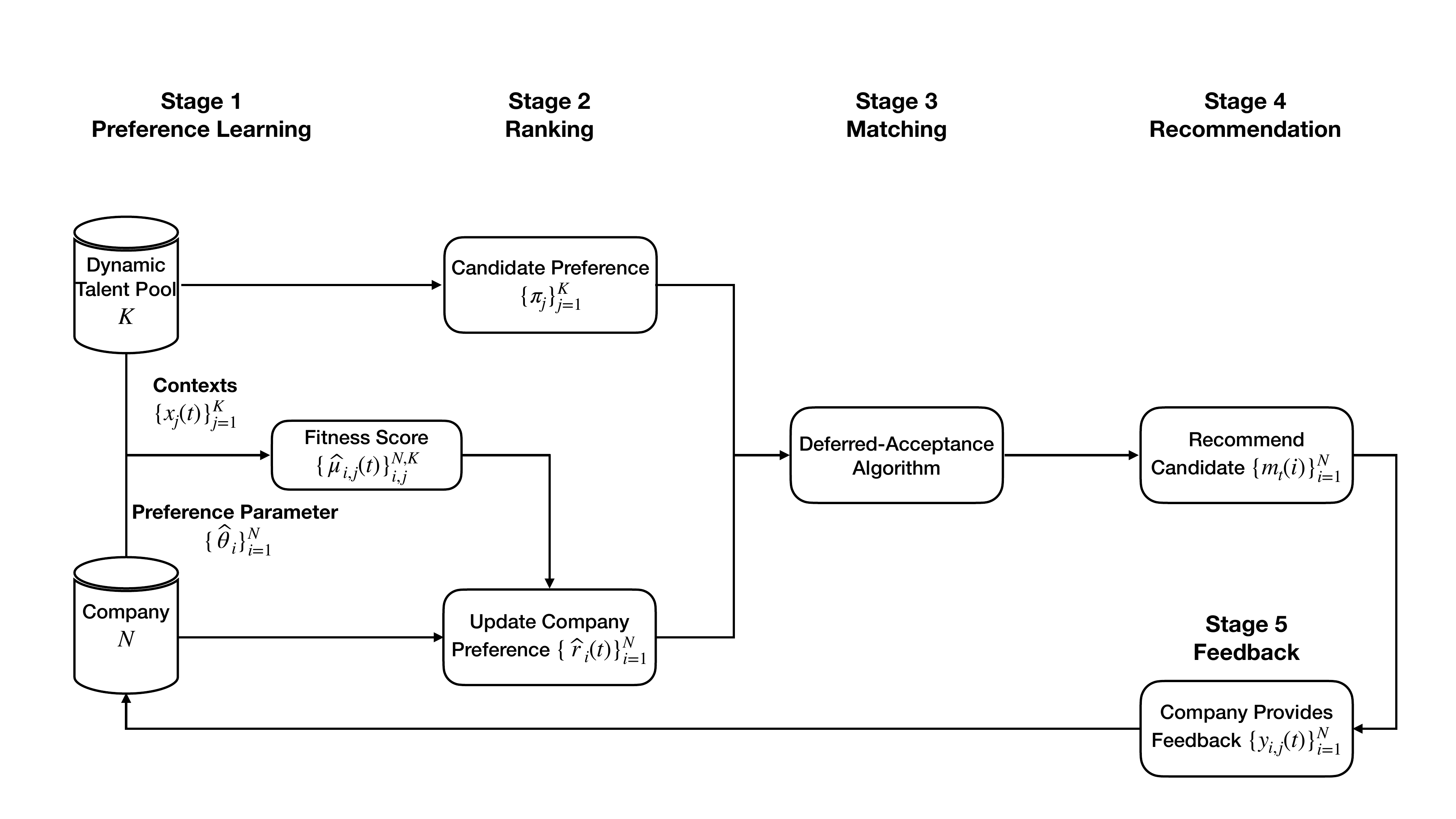}
    \caption{A generic design of dynamic matching platform.}
    \label{fig: algorithm flow}
\end{figure}

\begin{algorithm}[t]
\label{algo:CC-ETC}
\SetAlgoLined
	\DontPrintSemicolon
	\SetAlgoLined
	\SetKwInOut{Input}{Input}\SetKwInOut{Output}{Output}
	Input: Time horizon $T$; exploration loop $h$; ridge parameters $\lambda_{i}, \forall i \in [N]$; preference $\pi_{j}, \forall j \in [K]$.\\
	Learning Step: Get all companies' estimated true parameters: $(\widehat{\theta}_{1}(h),...,\widehat{\theta}_{N}(h))$ =  \texttt{Learning}($\N$, $\K$, $\pi_{j\in [K]}$, $\lambda_{i\in [N]}$, $h$) from Algorithm \ref{algo:exploration}.\\
	Exploitation Step: Get the matching result: \texttt{Exploitation}(T, $\N, \K$, $\pi_{j\in [K]}$, $\widehat{\theta}_{i\in [N]}(h)$) from Algorithm \ref{algo:exploitation}.  \\
	\caption{\texttt{Dynamic Matching Algorithm}}
\end{algorithm}

\subsection{Learning Step}

Let $h$ denote the learning length of dynamic matching algorithm. The key challenge is to find a sufficient learning length, which is a lower bound on $h$ such that the resulting algorithm secures a sub-linear regret. Determine the lower bound of $h$ is a challenging task due to many factors in DMP. We overcome this challenge by utilizing concentration results of the online ridge regression \citep{li2021online} to control probability of invalid ranking such that the agent will enjoy valid ranking with high probability. 
The theoretical choice of $h$ is provided in 
Corollary \ref{coro: upper order} in Section \ref{sec: theory-upperbound}.

After $h$ rounds, the platform collects the historical matching data $\mathbb{D}_{i}(h) = \{\M{X}_{i}(h), \V{y}_{i}(h)\}_{i=1}^{N}$, where $\M{X}_{i}(t) = [x_{i}(1), x_{i}(2), ..., x_{i}(t)]^{\Tra} \in \R^{t\times d}$ denotes $p_{i}$'s historical matched arms' profiles and $\V{y}_{i}(t) = [y_{i}(1), y_{i}(2), ..., y_{i}(t)]\Tra \in \R^{t}$ represents $p_{i}$'s historical noisy matching scores. With data $\mathbb{D}_{i}(h)$, the platform estimates $\{\theta_{i,*}\}_{i=1}^{N}$ through minimizing the mean square error with an $l_{2}$ penalty. Specifically, the objective function is
\begin{equation}
    \label{eq: loss function}
    \begin{aligned}
        \underset{\theta_{i}\in \R^{d}}{\min} \norm{\V{y}_{i}(h) - \M{X}_{i}(h) \theta_{i}}_{2}^{2} + \lambda_{i}\norm{\theta_{i}}_{2}^{2}, \quad \forall i \in [N],
    \end{aligned}
\end{equation}
where $\lambda_{i} > 0$ is the penalty parameter. 
The corresponding \textit{online ridge estimator} for company $p_{i}$ is
\begin{equation}
\label{eq:ridge est}
    \begin{aligned}
        \widehat{\theta}_{i}(h) =
        \big(
        \M{X}_{i} (h)\Tra \M{X}_{i} (h) + \lambda_{i} \M{I}_{d}
        \big)^{-1} \M{X}_{i} (h)\Tra \V{y}_{i} (h), \quad \forall i \in [N].
    \end{aligned}
\end{equation}
The learning step is available in Algorithm \ref{algo:exploration}. 
From lines 3-7, the platform sequentially updates the collected contextual information $\M{\Sigma}_{i}(t)$ and matching scores' information $\M{S}_{i}(t)$. In the end, \CCETC{} obtains the estimated preference parameter $\{\widehat{\theta}_{i}(h)\}_{i=1}^{N}$.

\begin{algorithm}[h]
\label{algo:exploration}
\SetAlgoLined
	\DontPrintSemicolon
	\SetAlgoLined
	\SetKwInOut{Input}{Input}\SetKwInOut{Output}{Output}
	Input: Number of companies $N$; number of \jb{}  $K$; preference $\pi_{j}, \forall j \in [K]$; ridge parameters $\lambda_{i}, \forall i \in [N]$; learning length $h$.\\
	Initialization: $\M{\Sigma}_{i}(0) = \lambda_{i} \M{I}_{d}$, $\M{S}_{i}(0) = \textbf{0}_{d}$, $\widehat{\theta}_{i}(0) = \textbf{0}_{d}$, for $\forall i \in [N]$.\\
	\For{$t \in \{1, ..., h\}$}{
    \For{$i \in \{1, ..., N\}$}{
    	\textsc{Match Arm}: Recommend job applicant $m_{t}(i)$ to company $p_{i}$.\\
        \textsc{Collect Response}: 
        Receive matching score $y_{i}(t)$ from company $p_{i}$.\\
        \textsc{Update Information}: Update the collected information for company $p_{i}$. $\M{\Sigma}_{i}(t) = \M{\Sigma}_{i}(t-1) + x_{m_{t}(i)}(t)x_{m_{t}(i)}(t)\Tra,  \M{S}_{i}(t) = \M{S}_{i}(t-1) + x_{m_{t}(i)}(t) y_{i}(t).$
        	}
    }
    \For{$i \in \{1, ..., N\}$}{
    	\textsc{Estimate Parameters}: Estimate preference parameter $\widehat{\theta}_{i}(h) = \M{\Sigma}_{i}^{-1}(t) \M{S}_{i}(t)$.
    	}
	\caption{\texttt{Learning Step}}
\end{algorithm}

\subsection{Exploitation Step}
In the Exploitation Step (Algorithm \ref{algo:exploitation}),
given the estimated preference parameter $\widehat{\theta}_{i}(h)$ from the learning step, platform constructs the estimated preference rankings $\{\widehat{r}_{t}(i)\}_{i=1}^{N}$ as follows. At $t = h + 1$, the platform estimates all arms' matching score for agent $p_{i}$ as
\begin{equation}
\label{policy: mean index}
    \begin{aligned}
        \widehat{\mu}_{i,j}(t) =\langle \widehat{\theta}_{i}(h), x_{j}(t) \rangle, \quad \forall i \in [N], j \in [K].
    \end{aligned}
\end{equation}
According these estimated matching scores $\widehat{\mu}_{i,j}(t)$, the platform ranks all arms in descending order. Denote the ranking list as $\widehat{r}_{i,[K]}(t) = \{\widehat{r}_{i,1}(t), ..., \widehat{r}_{i,K}(t)\}$ for agent $p_{i}$.
The platform then collects the estimated preferences of agents towards arms, $\{\widehat{r}_{i,[K]}(t) \}_{i=1}^{N}$, along with the arms' true preferences towards agents, $\{\pi_{j}\}_{j=1}^{K}$, which are assumed to be known in the DMP (see Section \ref{sec: Problem formulation}). Following this, the platform executes the DA algorithm.
Subsequently, the platform recommends job applicants $\{m_{t}(i)\}_{i=1}^{N}$ to each agent. In response, the companies provide their matching scores $\{y_{i}(t)\}_{i=1}^{N}$ to the platform, as illustrated in Stage 5 of Figure \ref{fig: algorithm flow}.

\begin{algorithm}[h]
\label{algo:exploitation}
\SetAlgoLined
	\DontPrintSemicolon
	\SetAlgoLined
	\SetKwInOut{Input}{Input}\SetKwInOut{Output}{Output}
	Input: Time horizon $T$; number of companies $N$; number of \jb{}  $K$; estimated true parameters $\widehat{\theta}_{i}(h), \forall i \in [N]$; preference $\pi_{j}, \forall j \in [K]$.\\
	\For{$t \geq h + 1$}{
	    \For{$i \in \{1, ..., N\}$}{
	        \textsc{Rank Candidates}: Estimate scores $\widehat{\mu}_{i,j}(t) = \widehat{\theta}_{i}(h)\Tra x_{j}(t), \forall j \in [K]$. Rank all \jb{} in descending order by $\{\widehat{\mu}_{i,j}(t)\}_{j=1}^{K}$ and get the preference ranking list $\widehat{r}_{i,[K]}(t)$.\\
         }
	   \textsc{Match}: With two-sided preferences $\{\widehat{r}_{i,[K]}(t)\}_{i=1}^{N}$ and $\{\pi_{j}\}_{j=1}^{K}$, platform computes stable matching $m_{t}$ via DA Algorithm \ref{algo:gs-algo}.\\
        \textsc{Receive Response}: Company $\cN$ provide their matching score $\{y_{i}(t)\}_{i=1}^{N}$.
        
    }
	\caption{\texttt{Exploitation Step}}
\end{algorithm}

\begin{rem}[Doubling Trick for Unknown $T$ for Dynamic Matching Algorithm] If $T$ is unknown, the platform can employ the \textit{doubling trick} \citep{auer1995gambling,besson2018doubling}. This approach involves initially setting a small $T$, and if more decisions are required beyond this horizon, the platform restarts the algorithm with a doubled horizon $T := 2T$ and  restart the learning step followed by the exploitation step, which suffers the same order regret upper bound as the dynamic matching algorithm with known $T$. 
\end{rem}

\begin{rem}[Computational Complexity]
    The computational costs for \CCETC{} consists of the learning step and the exploitation step. The learning step has an one-time estimation with cost $\mathcal{O}(d^{3})$ and matching cost $\mathcal{O}(NK)$. At each exploitation step, it has the ranking cost $\mathcal{O}(K\log K)$ and matching cost $\mathcal{O}(NK)$. So the total computational cost for $T$ steps' DMP is $\mathcal{O}((T-h)(K\log K + NK) + d^3 + NK)$. If $T$ is large, $d$ is small and $N \geq \log K$, the computational cost for DMP is $\mathcal{O}(TNK)$.
\end{rem}

\begin{rem}[Comparison with $\epsilon$-greedy Algorithm]
\label{sec: why etc}
The strategy of a learning step followed by an exploitation step is a typical approach in bandit learning \citep{lattimore2020bandit}, particularly when historical data is available. This method is widely used in applications such as website optimization \citep{garivier2016explore} and dynamic pricing \citep{fan2024policy}. It shares a similar exploration-exploitation tradeoff with the $\epsilon$-greedy algorithm \citep{auer2002finite}, but differs in the timing of exploration. Specifically, our approach conducts explorations initially, while $\epsilon$-greedy employs a randomized strategy with gradually reduced exploration over time. In our real data study, we compare the $\epsilon$-greedy method and our \CCETC{} in Section \ref{sec: real data}.
\end{rem}

\section{Connection Between Statistical Learning and DMP}
\label{sec: zero regret}
In this section, we mainly focus on the underlying relationship between the statistical learning and the dynamic matching problem. First in Section \ref{subsec:opt_matching_rank_inconsistent}, we explore two types of measure to characterize the correctness of ranking — \textit{correct ranking} and \textit{valid ranking} — that lead to optimal matching. In Section \ref{sec: unb and bias}, we show that both unbiased and biased estimations can achieve the optimal matching, and later we provide the motivation of our algorithm's design based on these findings. Finally, in Section \ref{sec: foundation of DMP}, we discuss the foundational terms determining the complexity of the DMP.
\begin{figure}[h] %
    \centering
    \begin{tikzpicture}[node distance=2cm and 3cm, auto] %
        \tikzstyle{box} = [rectangle, minimum width=3cm, minimum height=1.5cm, text centered, draw=black, fill=white!30, thick, line width=0.3mm]
        \tikzstyle{arrow} = [thick,->,>=stealth]

        \node [box] (box1) {\makecell{Unbiased Parameter \\ Estimation}};
        \node [box, right=of box1] (box2) {\makecell{Correct \\ Ranking}};
        \node [box, right=of box2] (box3) {\makecell{Optimal \\ Matching}};
        \node [box, below=of box2] (box4) {\makecell{Valid \\ Ranking}};
        \node [box, below=of box1] (box5) {\makecell{\makecell{Biased Parameter \\ Estimation}}};

        \draw [arrow] (box1) -- (box2) node[midway, above] {} node[midway, below] {Lemma \ref{claim: estimation condition}};
        \draw [arrow] (box2) -- (box3) node[midway, above] {} node[midway, below] {Claim \ref{claim: ranking condition}};
        \draw [arrow] (box2) -- (box4) node[midway, above] {} node[midway, below] {};
        \draw [arrow] (box4) -- ++(+5,0) node[midway, below] {Example \ref{exam: valid ranking}}  node[midway, above] {Lemma \ref{lem: valid claim}}-- ++(0,0)  -- ++(1,0) -- (box3) ; %
        \draw [arrow] (box5) -- (box4) node[midway, above] {} node[midway, below] {};
        \draw [arrow] (box5) -- (box2) node[midway, above] {} node[midway, below=5pt, sloped] {Example \ref{exmp: biased estimation}};

    \end{tikzpicture}
    \caption{Flow of sufficient conditions for optimal matching.}
    \label{fig:three-boxes}
\end{figure}
Ensuring all agents achieve the best matching outcome via the DA algorithm heavily relies on generating accurate ranking lists. Yet, operating within an online matching setup introduces the chance of producing partly accurate rankings due to limited data. These inaccuracies can profoundly affect matching results, leading to suboptimal outcomes for agents.

In the following part, we demonstrate that the key quantity for assessing the accuracy of ranking lists in the context of DMP is not merely the number of correctly ranked positions but rather the concept of a \textit{valid ranking}, which is a more precise and comprehensive measure that directly influences the ability to achieve optimal matching outcomes. 

\subsection{Correct Ranking and Valid Ranking}
\label{subsec:opt_matching_rank_inconsistent}

In the toy example provided below, we illustrate an intriguing scenario where achieving the optimal matching result is possible even when there are no correct ranking positions. %

\begin{exmp}\label{eg:optima_matching_inconsist_rank}
    Suppose the platform provides correct rankings for all agents except $p_{i}$, and assume the optimal matching arm for $p_{i}$ is at rank $j$. All ranks from $1$ to $j-2$ are permuted (i.e., $\widehat{r}_{i, k} \neq r_{i,k}$ and $\widehat{r}_{i, k} \in \{1, 2, ..., j-2\}$ for all $k \in [j-2]$). Similarly, all ranks from $j+1$ to $K$ are permuted ($\widehat{r}_{i, k} \neq r_{i,k}$ and $\widehat{r}_{i, k} \in \{j+1, j+2, ..., K\}$ for all $k \in [j+1, K]$). Additionally, the platform swaps the arm at rank $j-1$ with the arm at rank $j$ (the optimal matching arm). Despite this arrangement, agent $p_{i}$ can still achieve an optimal match. This is because all arms ranked before $j-1$ will be rejected based on the preferences from the other side (arm side), as per the DA algorithm, even when the positions of the arm at rank $j-1$ and the optimal matching arm at rank $j$ are switched.
\end{exmp}
The above example illustrates that the number of correct rankings is not the prime key determinant in achieving optimal matching. We provide the following claim to summarize.
\begin{clm}
\label{claim: ranking condition}
Correct ranking is a sufficient but not necessary condition for the optimal matching.
\end{clm}
Building on the above insight, we propose that the relative position of a wrongly ranked arm to the optimal arm is crucial in determining the achievement of optimal matching. Consequently, we introduce the term valid ranking to quantify this concept.
To better present the concept of valid ranking, we first classify arms based on its relative position over the optimal arm.
\begin{defn}[Types of Arms]
\label{def: classification of arm}
Arms can be classified into two types.
\begin{itemize}
    \item \textit{Sub-optimal matching arms set}: $\mathcal{K}_{i, \text{sub}}(t)=\{a_{j} |\overline{\Delta}_{i,j}(t) > 0, j\in [K]\}$, which is similar to the definition in the single-agent bandit problem;
    \item \textit{Super-optimal matching arms set}: $\mathcal{K}_{i,\text{sup}}(t)=\{a_{j} | \overline{\Delta}_{i,j}(t) < 0, j\in [K]\}$, which is a unique definition for the dynamic matching problem.
\end{itemize}
Here the score gap $\overline{\Delta}_{i,j}(t) = \mu_{i, \overline{m}_{t}(i)}(t) - \mu_{i, j}(t)$ is defined in Eq. \eqref{eq: score gap}.
\end{defn}

\begin{defn}[Valid and Invalid Ranking]
\label{def: invalid ranking}
Ranking $\widehat{r}_{i,[K]}(t)$ is \textit{valid} when, if an arm $a_{j}$ from the super-optimal matching arms set is ranked lower than the optimal matching arm $\overline{m}_{t}(i)$ (i.e, $\widehat{r}_{i,j}(t) > \widehat{r}_{i, \overline{m}_{t}(i)}(t)$), then it must follow that the score $\mu_{i,j}(t) > \mu_{i, \overline{m}_{t}(i)}(t)$. Conversely, if an agent ranks arms from the sub-optimal matching arms set higher than the agent-optimal arm, then it is considered \textit{invalid}.
\end{defn}
Valid ranking necessitates that the arms from the sub-optimal group $\mathcal{K}_{i, \text{sub}}(t)$ are not ranked higher than the optimal arms $a_{\overline{m}_{t}(i)}$ for agent $p_{i}$ at time $t$, rather than requiring fully correct ranking. This perspective contrasts with focusing solely on the number of correct rankings. We conclude that if agent $p_i$ maintains a valid ranking and all other agents also possess valid rankings, then all agents can achieve optimal matching. This suggests that maintaining all rankings as valid is inherently simpler, thus simplifying both the learning objectives and the matching process.
\begin{lem}
\label{lem: valid claim}
If all agents maintain valid rankings, they all obtain the agent-optimal matching.
\end{lem}
The detailed proof of Lemma \ref{lem: valid claim} is available in Section \ref{Appendix: Lemma 6 proof} of Appendix.
To illustrate Lemma \ref{lem: valid claim}, we consider a simplified scenario with two agents and three arms. 
\begin{figure}[t]
\centering
\includegraphics[scale=.25]{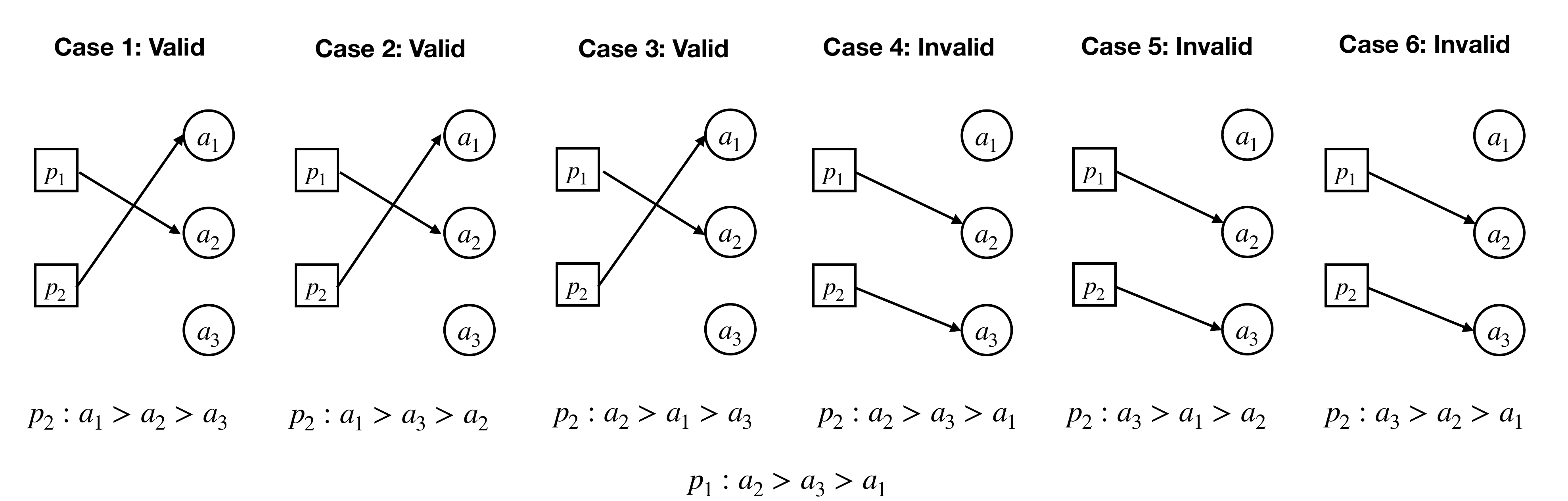}
\caption{The corresponding matching results for $p_{1}$ and $p_{2}$ if $p_{1}$ has valid ranking $a_{2} > a_{3} > a_{1}$ and $p_{2}$ has six possible rankings. Valid ranking for both and optimal matching: Case 1, 2, and 3. 
Single invalid ranking and non-optimal matching: Case 4, 5, and 6.}
\label{fig:6-cases}
\end{figure}

\begin{exmp}
\label{exam: valid ranking}
We assume agent $p_{1}$ has the valid ranking $a_{2} > a_{3} > a_{1}$ and agent $p_{2}$ has any one of the six possible rankings (3! permutations), and preferences from agents to arms and arms to agents are 
\begin{equation*}
    \begin{aligned}
        p_{1}: a_{2} > a_{1} > a_{3}, \qquad p_{2}: a_{2} > a_{1} > a_{3}\\
        \pi_{1}: p_{1} > p_{2}, \qquad  \pi_{2}: p_{1} > p_{2}, \qquad \pi_{3}: p_{1} > p_{2}.
    \end{aligned}
\end{equation*}
Given this preference setup, the agent-optimal matching for agents is $\{(p_{1}, a_{2}), (p_{2}, a_{1})\}$. 
In addition, all cases of the final matching results shown in Figure \ref{fig:6-cases} can be categorized as follows:

\noindent\textbf{Cases 1,2,3} (Optimal matching).
\textit{If agent $p_{2}$ has valid ranking as in Cases 1,2,3, the matching result is still $(p_{1}, a_{2})$ and $(p_{2}, a_{1})$. As long as $p_{2}: a_{3} > a_{1}$ (omit $a_{2}$) and $p_{1}$ has valid ranking, no agents suffers a regret.}

\noindent\textbf{Cases 4,5,6} (Non-optimal matching).
\textit{If agent $p_{2}$ has invalid ranking as in Cases 4,5,6, the matching result is no longer $(p_{1}, a_{2})$ and $(p_{2}, a_{1})$. Since $p_{2}: a_{1} > a_{3}$ (omit $a_{2}$) and even $p_{1}$ has a valid ranking, $p_{2}$ still suffers a regret.}
\end{exmp}

We observe that even if agent $p_{2}$ does not have a correct ranking, it can still achieve an optimal match in Cases 1, 2, and 3. Analyzing these six cases provides insights into how agents can achieve optimal matching results despite having incorrect rankings. %

\subsection{Unbiased Estimation and Biased Estimation}
\label{sec: unb and bias}

Lemma \ref{lem: valid claim} underscores the importance of the valid ranking property in attaining optimal matching. Essentially, obtaining a valid ranking relies on having accurate estimates of preference parameters, ensuring the correct ranking as the sample size grows. Naturally, an unbiased estimator serves as a favorable choice in this context.
\begin{clm}
\label{claim: estimation condition}
    Unbiased estimation is a sufficient condition for the optimal matching.
\end{clm}
However, we present an example demonstrating that correct/valid ranking can be achieved even with biased estimations of the preference parameters.
\begin{exmp}
\label{exmp: biased estimation}
(1) \textit{Correct ranking}. If $\widehat{\theta}_{i} = \theta_{i} + b$, where $b\in \mathbb{R}^{d}$, this results in a biased matching score $\widehat{\mu}_{i,j} = \widehat{\theta}_{i}^\top x_{j} = \theta_{i}^\top x_{j} + b^\top x_{j} = \mu_{i,j} + b'$ for all $j \in [K]$. Despite this bias, the correct ranking $\widehat{r}_{i} = r_{i}$ can still be maintained because the small shift $b'$ is consistently applied across all arms. (2) \textit{Valid ranking}. This approach involves applying a personalized bias $b_{i}'$ to these estimators, ensuring that as long as we maintain a valid ranking, it can produce an optimal matching. 
\end{exmp} 

While it's clear that an unbiased estimator can lead to optimal matching, the pursuit of such an estimator often involves significant computational expenses, rendering it impractical. A more feasible strategy is to employ a biased version of online estimation that remains sufficiently close to $\theta_{i,*}$. Despite its bias, this approach is still capable of recovering the correct or valid ranking, thus achieving optimal matching. This rationale underpins our algorithm design, which utilizes the online ridge regression method, as detailed in Section \ref{section: policy}.

\subsection{Foundations of DMP}
\label{sec: foundation of DMP}
In addressing the DMP, it's crucial to differentiate between online ranking and online estimation challenges, as they fundamentally guide the strategy of an algorithm's design and implementation. The key quantity that distinguishes between these two types of problems is the ``rate of error decay'' in relation to the sample size.
\cite{chen2022optimal} proposed that online ranking problems tend to be less challenging than online estimation problems in terms of sample complexity. This is primarily because the number of incorrect rankings in online ranking problems decays exponentially with the increase in sample size, while in online estimation problems, the decay is polynomial. This distinction suggests that the ranking method can swiftly approach the optimal matching in simpler scenarios. However, in more complex problems, the ranking method may struggle due to the influence of feedback noise.

Indeed, the primary challenge that impedes the straightforward application of either the online ranking or online estimation approaches lies in the information about the DMP's complexity available to determine which algorithm is optimal. 
This complexity leads to the characterization of DMP as typically presenting an online \textit{dual-layered mixture of ranking and estimation challenges}.
Firstly, DMP results in a divergence of difficulty experiences due to its multi-agent nature; at any given decision point, some agents are primarily dealing with a ranking problem while others grapple with an estimation problem. This variation is influenced by the interplay between the preference parameters $\theta_{i}$ for each agent $i \in [N]$ and the contextual attributes $x_{j}(t)$ for each arm $j \in [K]$.
Secondly, the dynamic and online nature of DMP means that the difficulty dynamically shifts for each agent between ranking and estimation challenges, corresponding to continuously evolving contexts.

This insight underscores the need for flexible estimation strategies and decision-making within the DMP framework. Therefore, we have designed our algorithm from the perspective that the worst-case scenario is one where all agents are confronted with dynamic online estimation problems. It ensures that our algorithm is robust and capable of adapting to the most challenging conditions, providing reliable performance even under significant variability and uncertainty in agent preferences and market dynamics.

\section{Regret Optimality of Dynamic Matching Algorithm}
\label{sec: theory-upperbound}
In this section, we outline the properties that our algorithm possesses.  We first state assumptions in Section \ref{sec: regu condi}. Subsequently, in Section \ref{sec: main upper bound}, we present the agent-optimal logarithmic regret upper bound of \CCETC{}, accompanied by an outline of the proof that includes critical steps in decomposing regret and its key lemma. Additionally, we demonstrate in Section \ref{sec: stability of CCETC} that our algorithm produces stable matching with high probability. Furthermore, in Section \ref{sec: theory-lowerbound}, we provide an instance-dependent regret lower bound along with its proof outline.

\subsection{Regularity Conditions}
\label{sec: regu condi}
We first assume the noise follows the subguassian distribution.
\begin{ass}[Subgaussian Noise]
\label{ass: noise ass}
    The noise $\epsilon_{i,j}(t)$'s are drawn independently from a $\sigma$-subgaussian distribution for $t \in [T], i \in [N], j \in [K]$. That is, for every $\alpha \in \R$, it is satisfied that 
$\mathbb{E}[\exp(\alpha \epsilon_{i,j}(t))] \leq \exp(\alpha^2\sigma^2/2)$. 
\end{ass}

Next, we assume that the context $x_{j}(t)$ distribution of the arm $a_{j}$ is from distribution $\cD_{\cX_{j}}$ and the joint distribution of all arms $\cD_{\cX} = \cD_{\cX_{1}} \times ... \times \cD_{\cX_{K}}$ is independent product of individual context distribution $\{\cD_{\cX_{j}}\}_{j=1}^{K}$. 
\begin{ass}[Unit Sphere]
\label{ass: bounded covariates}
$\norm{x_{j}(t)}_{\infty} < 1, \forall j \in [K], t \in [T].$
\end{ass}
This assumption is common in literature \citep{bastani2020online, li2023double, wang2023online} and is satisfied when normalization is applied.
\begin{ass}[Positive-Definiteness]
\label{ass: covariance matrix} Define $V = \mathbb{E}[XX\Tra|X \in \cD_{\cX}]$. Then there exists a deterministic constant $\phi_{0} \in \mathbb{R}^{+}$ such that for all $X \in \cD_{\cX}$ we have the minimum eigenvalue of the covariance matrix  $\lambda_{\min}(V) \geq \phi_{0}^{2}$.
\end{ass}

Assumption \ref{ass: covariance matrix} is referred to as the \textit{compatibility condition} in online statistical learning literature \citep{li2021online} and is to ensure that the online ridge estimate trained on samples $X \in \cD_{\cX}$ converges to the true preference parameter $\{\theta_{i,*}\}_{i=1}^{N}$ with high probability as the sample size grows.

\begin{ass}[Uniform Sub-optimal Minimal Gap Condition]
\label{ass: mirror Margin}
The difference in terms of the scaled matching score between the agent-optimal matching arm and arms from suboptimal arm $\K_{i, \text{sub}}(t)$ for all agents over $T$ is uniformly greater than $\rho > 0$. That is, $$\widetilde{\Delta}_{i, \min} = \min_{t\in [T]}\min_{j \in \K_{i, \text{sub}}(t)}\overline{\Delta}_{i, j}(t) /|| x_{\overline{m}_{t}(i)}(t)- x_{j}(t)||_{2}   > \rho, \forall i \in [N].$$
\end{ass}
Assumption \ref{ass: mirror Margin} assures the uniqueness of the agent-optimal match.
This assumption extends the fixed uniform sub-optimal minimal gap condition in static matching contexts \citep{liu2020competing} to the dynamic matching framework where the true matching score $\mu_{i,j}(t)$ varies.

Without loss of generality, we have the following assumption over the preference parameter.
\begin{ass}[Positive Preference]
\label{ass: positive theta}
$\theta_{i, *}^{(r)} > 0, \forall r \in [d], i \in [N]$.
\end{ass}
Assumption \ref{ass: positive theta} captures the fact that agents evaluate arms' attributes positively but with varying priorities based on the fitness of the arm to the agent.\footnote{In practice, if this assumption does not hold, the platform can initially estimate it, find that parameter entries are negative, and subsequently adjust the sign of the context.}

\subsection{Regret Upper Bound}
\label{sec: main upper bound}
In this section, we provide the agent-optimal regret upper bound of our algorithm as follows.
\begin{thm}
\label{thm: upper bound of cc-etc}
\textit{With Assumptions \ref{ass: noise ass} - \ref{ass: positive theta} and given the  learning length $h$, if the platform follows the \CCETC{}, agent $p_{i}$'s regret up to time $T$ is upper bounded by}
\begin{equation}
\label{eq: thm1 regret upper bound}
    \begin{aligned}
        R_{i}(T) 
        & \leq 
        \underbrace{\sum_{t=1}^{h} \overline{\Delta}_{i,m_{t}(i)}(t)}_\text{Part I Regret} + 
        2\underbrace{C_{0}(\lambda_{i})Nd 
        \Bigg[\sum_{t = h + 1}^{T}\overline{\Delta}_{i, \max}(t)(K-\min_{i\in [N]}\tau_{i}(t))\Bigg]
        \exp\Bigg[-h\frac{2\phi_{0}^{4}\rho^{2}}{d^2\sigma^2}
        \Bigg]}_\text{Part II Regret},
    \end{aligned}
\end{equation}
\textit{where $C_{0}(\lambda_{i}) =\exp\big[-4\lambda_{i}\phi_{0}^{2}\rho^{2}/d^2\sigma^2 \big]$, $\tau_{i}(t)$ is the agent $p_{i}$'s optimal matching object's ranking position $r_{i, \overline{m}_{t}(i)}(t)$, and $x_{i,\max} = \norm{\M{X}_{i}(h)}_{\infty}$ is the maximum absolute value of the context entry.}
\end{thm}

Theorem \ref{thm: upper bound of cc-etc} provides the decomposed regret upper bound of the dynamic matching. We split the regret into two parts, the learning step's regret (``Part I Regret") and the exploitation step's regret (``Part II Regret"). The detailed decomposition procedure can be found in Section \ref{sec: upper bound regret decomp}. With an optimized learning step $h$, next corollary shows that dynamic matching algorithm has a logarithmic regret.

\begin{cor}
\label{coro: upper order}
\textit{With $h = \left\lceil \underset{i \in [N]}{\max}  \frac{d^2\sigma^2}{ 2\phi_{0}^{4}\rho^{2} } \log \frac{4C_{0}(\lambda_{i})NK \phi_{0}^{4} \rho^{2}}{d \sigma^2 \overline{\Delta}_{i, \max}}T\right \rceil$, we have 
\begin{equation}
\label{upper bound order}
    \begin{aligned}
    R_{i}(T) 
    &\leq  C_{1}(1+\log{C_{2}T}) = \mathcal{\tilde{O}}\bigg(\frac{d^{2}\sigma^{2}}{\rho^{2}}\log(NKT)\bigg)
    \end{aligned}
\end{equation}
where $C_{1} = d^2 \sigma^2\overline{\Delta}_{i,\max}/(2\phi_{0}^{4}\rho^{2})$ and  $C_{2} = 4C_{0}(\lambda_{i})NK \phi_{0}^{4} \rho^{2}/(d \sigma^2\overline{\Delta}_{i,\max})$.
}
\end{cor}

For part II regret in Eq.\eqref{eq: thm1 regret upper bound}, it depends on the gap between the minimum optimal arm rank $\min_{i\in [N]}\tau_{i}(t)$ across all arms and the worst arm, which measures the difficulty of dynamic matching problem's characteristic. If there exists an agent's optimal arm rank $\tau_{i}(t)=1$, the gap is $K - \min_{i\in [N]}\tau_{i}(t)) =  K-1$.
Given the optimal learning length $h$ from Eq. \eqref{upper bound order}, we find the regret upper bound depends logarithmically over $T$, which means that it is a no-regret learning method.

We next discuss the dependence of the upper bound on several critical parameters. The quantity $\rho^{2}/d^{2}\sigma^{2}\log{NK}$ represents the signal-to-noise ratio for dynamic online matching problem. When the signal-to-noise ratio is high, the complexity level of the DMP is low; conversely, in the low signal-to-noise regime, the complexity of the DMP is high.
From another perspective, if the uniform sub-optimal minimal gap $\overline{\Delta}_{i, \min}$ is small, \CCETC{} faces a challenging task as it becomes difficult to distinguish between the optimal arm and the suboptimal arm. Consequently, the ranking provided by the platform is prone to errors, potentially leading to non-optimal stable matching results.
Moreover, we observe that both $N$ and $K$ increase at a logarithmic rate in terms of regret when the number of participants increases. We further provide an instance-dependent lower bound, which matches the order of the regret upper bound (see Section \ref{sec: theory-lowerbound}).

\subsubsection{Proof Outline}
\label{sec: upper bound regret decomp}
We provide key steps to prove Theorem \ref{thm: upper bound of cc-etc}.
We decompose the agent regret $R_{i}(T)$ into the learning step regret and the exploitation step regret as follows:
\begin{align*}
        R_{i}(T) = \sum_{t=1}^{T}\mu_{i,\overline{m}_{t}(i)}(t) - \mu_{i,m_{t}(i)}(t) \leq \underbrace{\sum_{t=1}^{h} \overline{\Delta}_{i,m_{t}(i)}(t)}_\text{Part I Regret}
        + 
        \underbrace{ 
        \Bigg[\sum_{t = h + 1}^{T}\overline{\Delta}_{i, \max}(t)
        (N \mathbb{P}(\widehat{r}_{i,t}  \text{ is invalid}))  
        \Bigg]}_\text{Part II Regret}.
\end{align*}
The ``Part I regret'' is the sum of gaps between the optimal arm and the arm recommended by \CCETC{} during the learning rounds. The Part II regret accumulates during the exploitation step. Based on Lemma \ref{lem: valid claim}, it is necessary to quantify the probability of an invalid ranking to calculate the instantaneous regret.
At time $t$, the instantaneous regret $\Delta_{i, m_{t}(i)}(t) N \mathbb{P}(\widehat{r}_{i,t} \text{ is invalid}) \leq \overline{\Delta}_{i, \max}(t) N \mathbb{P}(\widehat{r}_{i,t} \text{ is invalid})$. We quantify the probability of invalid ranking $\mathbb{P}(\widehat{r}_{i,t} \text{ is invalid})$ in the following Lemma \ref{lem: invalid prob}. We then sum all instantaneous regrets from time $h+1$ to $T$ to determine the ``Part II Regret," as shown in Equation \eqref{eq: thm1 regret upper bound}. Following this, we provide the upper bound of the invalid ranking probability.

\begin{lem}
\label{lem: invalid prob}
Assume all agents receive recommended arms from \CCETC{}, the invalid ranking probability's upper bound,
\begin{equation}
    \begin{aligned}
     \mathbb{P}(\widehat{r}_{i,t}  \text{ is invalid})  \leq 
     2d(K-\tau_{i}(t))\exp\bigg(-h\frac{2\lambda_{i}^{2}\rho^{2}\phi_{1}^{4}}{d^2\sigma^2}\bigg).
    \end{aligned}
\end{equation}
\end{lem}

\subsection{Matching Stability of Dynamic Matching Algorithm}
\label{sec: stability of CCETC}
In this section, we prove that our algorithm provides stable matching result with high probability.
While the DA algorithm can achieve stable matching based on two-sided true preferences, this scenario is typically unavailable in the initial decision rounds of the online setting, due to the uncertainty in the preference estimation. Our theory identifies the optimal minimum learning length in Corollary \ref{coro: upper order}, which guarantees that \CCETC{} delivers stable matching with high probability (Theorem \ref{thm: stability of ccetc}). This result connects the online learning techniques and offline matching algorithm design, providing key insights for designing more general online dynamic matching algorithms. 

\begin{thm}
\label{thm: stability of ccetc}
If
$t \geq \lceil{\frac{d^2 \sigma^2}{2\rho^2\phi_{0}^4}[\log(2d(K-1)) -  \log(1 - \Psi^{1/N})] \rceil},$ the \CCETC{} provides an agent-optimal stable matching solution with a probability at least $\Psi > 0$.
\end{thm}
This theorem indicates that no agents will deviate from the recommended matching arm with a probability of at least $\Psi$ after time $t$. The proof sketch of the stability property of \CCETC{} consists of two steps, naturally following the design of \CCETC{}. 
In the exploitation step, DA still produces a stable matching result based on estimated preferences and there are no blocked pairs during the matching procedure with high probability and the main proof follows Lemma \ref{lem: invalid prob} with a union bound of the valid ranking.

\subsection{Instance-Dependent Regret Lower Bound}
\label{sec: theory-lowerbound}
We next provide the instance-dependent regret lower bound over a two-agent, three-arm instance and demonstrate the matching lower bound of our algorithm.

In the following lower bound analysis, we consider that there are two agents and three arms in the platform. Contexts are generated from the uniform distribution, $x_{j}(t) \sim U(0, 1)^{d}, \forall t \in [T], \forall j \in [K]$.
We also assume the true preference parameter are designed as follows $\theta_{1,*} = (\sqrt{1-1/h}, 1/\sqrt{h}, 0, ..., 0)^{T}\in \mathbb{R}^{d}, \theta_{2,*} = (\sqrt{1-1/h}, 0, 1/\sqrt{h}, 0, ..., 0)^{T} \in \mathbb{R}^{d}$. Noise follows Gaussian distribution with variance $\sigma^{2}$. 
Then the estimator from Eq. (\ref{eq:ridge est}) satisfies,
\begin{equation*}
    \begin{aligned}
        \widehat{\theta}_{i}(h)|\mathcal{F}_{i}(h) \sim N(\bar{\theta}_{i},
        \sigma^{2} \M{M}_{i}), \quad i \in [N],
    \end{aligned}
\end{equation*}
where $\bar{\theta}_{i} = 
\big(\M{X}_{i} (h)\Tra \M{X}_{i} (h) + \lambda_{i} \M{I} \big)^{-1} 
\M{X}_{i}(h)\Tra \M{X}_{i}(h)
\theta_{i,*} \in \R^{d}$, and
$\text{Cov}[\widehat{\theta}_{i}(h)|\mathcal{F}_{i}(h)] = \sigma^{2}\big(\M{X}_{i} (h)\Tra \M{X}_{i} (h) + \lambda_{i} \M{I} \big)^{-1}
\M{X}_{i}(h)\Tra \M{X}_{i}(h) 
\big(\M{X}_{i} (h)\Tra \M{X}_{i} (h) + \lambda_{i} \M{I} \big)^{-1}
$ $\in \R^{d\times d}$.

\begin{thm}
\label{thm: lower bound}
Consider the designed two-agent three-arms instance above. The regret lower bound for agent $p_{i}$ is, 
\begin{equation}
    R_{i}(T) \geq 
        \sum_{t=1}^{h} \Delta_{i,m_{t}(i)}(t)
        +
         \sum_{t = h+1}^{T} \overline{\Delta}_{i, \min}
         \big[\mathcal{L}_{i}^{b}(t)\mathcal{L}_{j}^{b}(t)  + \mathcal{L}_{i}^{b}(t) \mathcal{L}_{j}^{g}(t) \big],
\end{equation} 
where $\mathcal{L}_{i}^{g}(t)= 1 -(3/c_{5}(t)\sqrt{2\pi h})\exp{(-
c_{5}^{2}(t)h/2)}$,
$\mathcal{L}_{i}^{b}(t) =(1/c_{7}(t)\sqrt{h} - 1/c_{7}^{3}(t)h^{3/2})
\exp{(-c_{7}^{2}(t)h/2)}$, and $c_{5}(t), c_{7}(t)$ are contextual time-dependent constants but independent of designing exploration rounds $h$. With the optimized $h$ provided by \CCETC{},
 the order of the regret lower bound is 
$R_{i}(T) = \Omega(\log(T))$.
\end{thm}

From Theorem \ref{thm: lower bound}, we find that our algorithm achieve a matching regret lower bound. This lower bound not only depends on both agents' incorrect ranking's probability lower bound $\mathcal{L}_{i}^{b}(t), i = 1,2$, but also the other agent's ($p_{j}$) correct ranking estimate's probability lower bound $\mathcal{L}_{j}^{g}(t)$.

\begin{rem}
    \cite{liu2020competing} provided a regret lower bound for the non-contextual non-dynamic setting by considering other agents  submitting truthful rather than strategic rankings to the platform and bounding the maximum number of pulls of non-optimal arms. %
    \cite{jagadeesan2021learning} presented a lower bound for the multi-armed bandit problem instance with money transfer, as opposed to the strict preference constraint like ours.
    \cite{li2022minimax} considered the minimax lower bound for the multi-agent Markov game where it shares the same action space for all agents. However, our DMP setting is different from their setting due to the exclusive action selection characteristic. In DMP, for agents, there is \textit{exclusivity} in action selection and this exclusivity is ubiquitous since one arm cannot be matched with two agents.
\end{rem}

\subsubsection{Proof Outline of Theorem \ref{thm: lower bound}}
\label{sec-main: regret decompose}
The agent-optimal regret $R_{i,t}$ for agent $p_{i}$ at time $t$ can be decomposed as 
\begin{equation}
\label{eq: regret decompose new}
\begin{aligned}
R_{i,t} &= \mathbb{E}[\mathbb{E}[R_{i,t}|\text{other agents' ranking status}]]\\
&= \mathbb{E}[R_{i,t}|\bigcap_{j\neq i} \{\widehat{r}_{j}(t) = r_{j}(t)\}]\mathbb{P}(\underbrace{\bigcap_{j \neq i }\{\widehat{r}_{j}(t) = r_{j}(t)\}}_{\text{Event I}}) + \mathbb{E}[R_{i,t}| \underbrace{\bigcup_{j\neq i}\{\widehat{r}_{j}(t) \neq r_{j}(t)\}}_{\text{Event II}}] \mathbb{P}(\bigcup_{j\neq i}\{\widehat{r}_{j}(t) \neq r_{j}(t)\}).
\end{aligned}
\end{equation} 
Here if we assume $i = 1$, Event I becomes $\{\widehat{r}_{2}(t) = r_{2}(t)\}$, $p_{2}$ has a correct ranking. Event II becomes $\{r_{2}(t) \neq r_{2}(t) \}$, $p_{2}$ has incorrect ranking. 
$R_{1,t}(\widehat{r}_{2}(t) = r_{2}(t))$
is the first component of Eq. \eqref{eq: regret decompose new}, which is the expected instantaneous regret for $p_{1}$ conditioning on  $p_{2}$ having correct ranking at time $t$. 
Similarly, $R_{1,t}(\widehat{r}_{2}(t) \neq r_{2}(t))$  is the second component of 
Eq.\eqref{eq: regret decompose new}, the expected instantaneous regret for $p_{1}$ conditioning on $p_{2}$ having incorrect ranking at time $t$. $R_{1,t}(\widehat{r}_{2}(t) = r_{2}(t))$ and $R_{1,t}(\widehat{r}_{2}(t) \neq r_{2}(t))$ are product of the Event I and II's probabilities and corresponding expected regret. Thus, the next step is to quantify the lower bound of two probabilities and expected regret, which we provide in Lemmas \ref{lem: good event: lower bound} and \ref{corr: uniform good lower bound and failure lower bound} of appendix. 

\section{Experiments}
\label{sec: experiments}
This section demonstrates the effectiveness and robustness of \CCETC{} in simulation and real data, where the simulation studies include five different settings, its robustness under different context distributions (S1 $\&$ S2). The additional experiments such as  different minimal margins (S3), different feature vector dimensions (S4), and different sizes of agents and arms (S5) are available at Section \ref{supp-sec: add sim} of appendix. In real data, we apply the \CCETC{} in an online job-seeking market.

\subsection{Simulation}
\label{sec: simulation}

From Scenario 1 to Scenario 4, we consider that there are two agents $N=2$ and three arms $K = 3$. In Scenario 5, we consider that $N=K=5$. The penalty parameters for all agents are set to be $\lambda = 0.1$ in all scenarios and $T = 1000$.

\textit{Scenario 1 (S1)}: Contexts are generated from a $d$-dimensional normal distribution with four different fluctuation variance $\zeta= [0.01, 0.05, 0.1, 0.2]$ and $d=2$, and normalized to have unit norm. $\{\theta_{i, *}\}_{i=1}^{2}$ are randomly generated from uniform distribution and scaled to have unit norm. 
The uniform minimal sub-optimal condition for this scenario is set to be  $\rho = 0.2$. 
In addition, the noise is generated from normal distribution with $\sigma = 0.05$.
We assume that arms to agents' preference $\pi$ are $a_{1}: p_{1}> p_{2},a_{2}: p_{2}> p_{1},a_{3}: p_{1}> p_{2}$. 
According to Corollary \ref{coro: upper order}, the optimal learning step length $h$ is 312.  
    
\textit{Scenario 2 (S2)}: Contextual features move with an \textit{angular velocity} $w_{t} = 0.005t$ which is different from S1, and $d=2$. Contexts for arms are still generated from normal distribution and normalized. But for $x_{1}(t)$, its mean is constantly increasing with a velocity $w_{t}$. 
The true parameters $\{\theta_{i, *}\}_{i=1}^{2}$ are the same as these in S1. The uniform minimal sub-optimal for this scenario is set to be $\rho = 0.2$.
The example of moving context with an angular velocity is illustrated in Figure \ref{fig:fig_4_cosine_fig}.
We consider three levels of noise $\sigma = [0.01, 0.02, 0.05]$ to test the robustness of our algorithm.
In S2, the agent-optimal matching is no longer fixed even when fluctuation level $\zeta = 0$ since context $x_{1}(t)$ is dynamic.  $h$ for three noise levels are $h = [24, 66, 312]$, correspondingly.  

Additional experiments settings and results are available in appendix.

\subsubsection{Results and Analysis}
\begin{figure}
    \centering
    \includegraphics[scale=0.5]{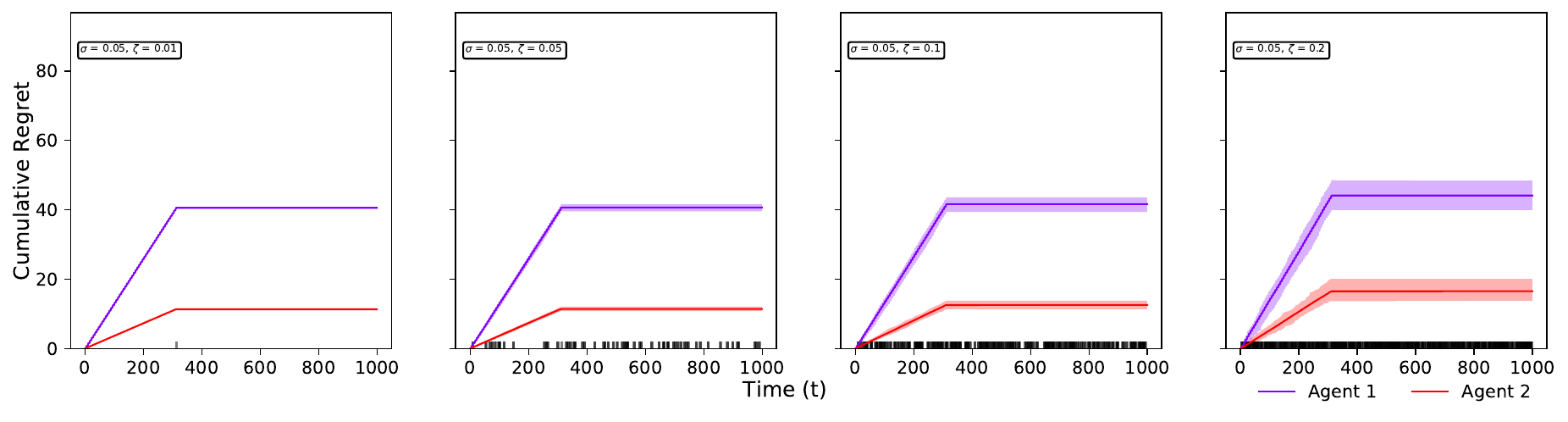}
    \caption{S1: Cumulative regret for different  context variation levels $\zeta$. Each black stick means a change of optimal matching.}
    \label{fig:fig_2_non_global_normal} 
\end{figure}

\begin{figure}
\centering
\includegraphics[scale=1]{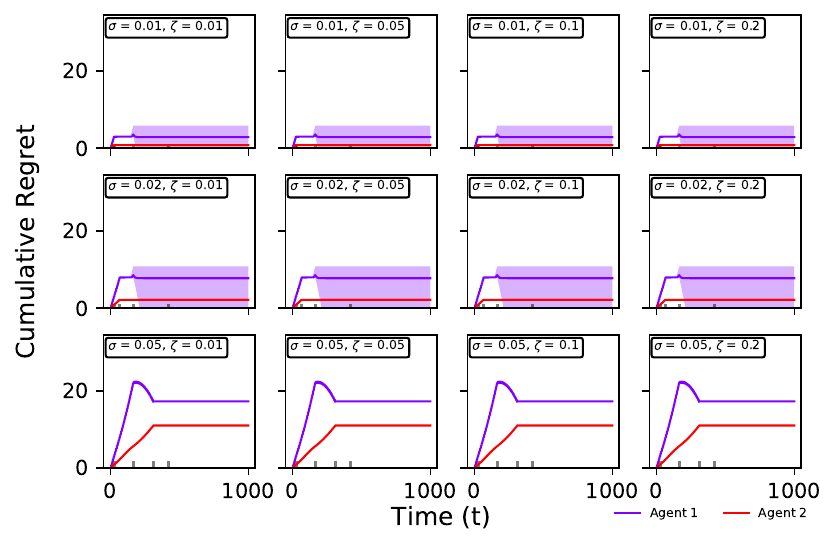}
\caption{Cumulative regret for different noise levels and context variation levels of mean shifting context in Scenario S2.}
\label{fig:fig_3_non_global_angle_vector_chi_0.2}
\end{figure}

In Figures \ref{fig:fig_2_non_global_normal} and \ref{fig:fig_3_non_global_angle_vector_chi_0.2}, the horizontal axis represents the time point and the vertical axis represents the cumulative regret. In all figures, we plot the maximum (worst) regret represented as the upper bound shaded line, mean regret represented as the solid line, and minimum (best) regret represented as the lower bound shaded line, over 100 replications. 

\textit{S1: \CCETC{} is robust to different contexts' variance levels $\zeta$.}  In Figure \ref{fig:fig_2_non_global_normal}, our \CCETC{} shows the logarithmic regret shape which demonstrates that it is robust to contexts' noise levels. 
When contexts' variance level $\zeta$ increases, the shaded area becomes wider, indicating the uncertainty of the regret increasing, and indicates that the complexity of the DMP is also larger.
In this figure and following figures, we use the short black sticks to represent the change of the optimal matching between two adjacent-time points due to the contextual information change. We mark the short black stick at time $t+1$ on the horizontal axis if $\overline{\V{m}}(t) \neq \overline{\V{m}}(t+1)$, which means that the optimal matching result is different on two continuous-time points.  
The denser the black stick is, the more frequently the agent-optimal matching changes over time.
In other words, when the contexts' fluctuation magnitudes increase, the optimal stable matching changes more frequently, and it exhibits the dynamic property of DMP.

\textit{S2: \CCETC{} is robust to mean shifting context distributions $w_{t}$ and different levels of observed score noise $\sigma$.}  In Figure \ref{fig:fig_3_non_global_angle_vector_chi_0.2}, we present the S2's results when the mean of arm $a_{1}$ changes with an angular velocity $w_{t}$.
In row 1, we find there is a small ``bump" in the mean regret of the cumulative regret in each plot, and the slight bump occurs at the exploitation step where the occurring time of the bump is greater than $h$. 

Two reasons cause the occurrence of the small bump. One is the coarse estimation of parameters. 
Another is that the context's angular velocity changes too slowly, violating the uniform sub-optimal minimal condition assumption. Both of these will cause the incorrect ranking estimated by the platform, resulting in regret.
A similar pattern can also be found in the second row of the figure. 
In order to demonstrate the conjecture of violating uniform sub-optimal minimal condition assumption, we decrease it $\overline{\Delta}_{i, \min}$ from 0.2 to 0.1 in S3 (Figure \ref{fig:fig_3_non_global_angle_vector_chi_0.1} in appendix), which indirectly extends the learning step $h$ and therefore increases the estimation accuracy because platform would gather more data to acquire more accurate estimates.
In addition, in the third row, the shaded area disappears because the learning step is long enough to accumulate sufficient data to get a reasonable estimate compared with the first row and the second row, which demonstrate our conjecture.

Another interesting finding is the decreasing regret phenomena after the bump. 
In the first and second rows of Figure \ref{fig:fig_3_non_global_angle_vector_chi_0.2}, the phenomenon of decreasing regret occurs because the agent, $p_{1}$, needs to \textit{recover} from the violation of the ``uniform sub-optimal minimal condition" assumption. This implies that agent $p_{1}$ is unable to distinguish the differences between arms when the uniform sub-optimal minimal condition is violated.
In the third row's, the regret decreases over a long period because agent $p_{1}$ is in the learning step 
and the \textit{super-optimal} arms have a much larger gain over \textit{sub-optimal} arms.
These significant gains will result in a negative regret. So the cumulative regret will decrease. This interesting phenomenon is only occurring in DMP when considering the contextual information. In all, we find that \CCETC{} is robust to changing the context format.

\subsection{Real Data}
\label{sec: real data}
We next apply the \CCETC{} in the job application market with job applicants' profile information and companies' job description information from LinkedIn. %

\subsubsection{Background}
Three job applicants and two companies are in the market. 

\noindent\textbf{Job Applicants' and Companies' Preferences}: Based on the profiles of the job applicants, three candidates with diverse backgrounds are seeking job opportunities in the market:
\begin{itemize}
    \item $a_{1}$: a data scientist (ds), 
    \item $a_{2}$: a software development engineer (sde),
    \item $a_{3}$: a quantitative researcher (qr).
\end{itemize}
In addition, two companies provides two job descriptions indicating that they are interested in hiring candidates with specific skill sets as follows:
\begin{itemize}
    \item Company $p_{1}$ is looking for a candidate with quantitative research skills,
    \item Company $p_{2}$ is looking for a candidate with software development skills.
\end{itemize}
Given this setup, the preferences of the three job applicants for companies can be described as:
\begin{itemize}
    \item For the data scientist, $a_{1}$, the preferences are $\pi_{ds}(a_{1}): p_{1}(\text{qr}) > p_{2}(\text{sde})$,
    \item For the software development engineer, $a_{2}$, the preferences are $\pi_{sde}(a_{2}): p_{2}(\text{sde}) > p_{1}(\text{qr})$,
    \item For the quantitative researcher, $a_{3}$, the preferences are $\pi_{qr}(a_{3}): p_{1}(\text{qr}) > p_{2}(\text{sde})$.
\end{itemize}
This indicates that each applicant prefers the company whose job description best matches their background and skills. Detailed description of the data is provided in Section \ref{app-sec: textual} of appendix.

\noindent
\textbf{Dynamic Contextual Information}:
To incorporate the dynamic contextual information, we take the following steps to construct the dynamic matching environment ($T=10800$):
\begin{enumerate}
\item Job applicants' dynamic contextual information: 
    \begin{itemize}
        \item At $t=0$: the job applicant $a_{j}$ has textual information $\V{w}_{j}(0)$, a sequence of words represented as $\V{w}_{j}(0) = \{w_{j}^{1}(0), w_{j}^{2}(0), ..., w_{j}^{q_{0}}(0)\}$ and $q_{0}$ is the length of the sequence of the words at time $t=0$, profile like \texttt{research projects on modeling of high-dimensional and multi-modal
        (partially observed), inputs for classification, regression and clustering tasks, leveraging a wide range of techniques}.
        \item At $t = 600z, z \in \mathbb{N}$: we assume that job applicants learn new skills, update profile like \texttt{(1)t=600, Strong interested in data science, (2)t=1200, machine learning, (3)t=1800, data visualization...}, and updates his profile, so the textual information becomes $\V{w}_{j}(t)$, the sequence of words is represented as $\V{w}_{j}(t) = \{w_{j}^{1}(t), w_{j}^{2}(t), ..., w_{j}^{q_{t}}(t)\}$ for all $t = 600z, z=1,2,...,18$.\footnote{One agent update profile every 1800 steps for different updating frequency.}\footnote{Here we create the streaming data is through adding additional textual information over time.}
    \end{itemize}
\item Companies' fixed job descriptions:
\begin{itemize}
    \item  The job descriptions from companies are fixed texts over time denoted by $\V{w}_{i} = \{w_{i}^{1}, w_{i}^{2}, ..., w_{i}^{p_{i}}\}$ where $p_{i}$ is the length of words for company $p_{i}$, job descriptions are like \texttt{Strong passion in quant finance, strong mathematical and statistical knowledge. Proficiency in programming languages like Python or R, etc.}
\end{itemize}
\end{enumerate}

The detailed text data is available in Tables \ref{table: job applicant profile} and \ref{table: job descript}  at Section \ref{app-sec: textual} of appendix.

\noindent\textbf{Text-to-Embedding}: We use the encoder of the Transformer model \citep{devlin2018bert} $f$ to generate the word embedding of these textual information from job applicants' profiles $\{\V{w}_{j}(t)\}_{j=1,2,3;t\in [T]}$, and companies' job descriptions $\{\V{w}_{i}\}_{i=1,2}$. 
\begin{equation}
    \begin{aligned}
        h_{j}(t) = f(\V{w}_{j}(t)), h_{i} = f(\V{w}_{i}),
    \end{aligned}
\end{equation}
where $h_{j}(t), h_{i} \in \mathbb{R}^{d_{raw}}$ and $d_{raw} = 768$ is the commonly output dimension of the transformer model \citep{devlin2018bert}. For simplicity, here we use PCA method \citep{pearson1901liii, jolliffe2016principal} to extract the most significant dimension from these word embedding vectors for job applicants and add a Gaussian noise to $h_{j}(t)$ at every time step to transform it into streaming data. So the observed contextual information for each job applicant is $x_{j}(t) = \text{PCA}(h_{j}(t)) + N(0, \zeta^{2}) \in \mathbb{R}^{d}$ where $d=3$ and $\zeta=1e-6$.

\noindent\textbf{True Response}: The true response $y_{i,j}(t)$ is determined by the similarity of the job applicant's profile and job description with an added Guassian noise, that is, $y_{i,j}(t) = h_{j}\Tra h_{i}(t) + \epsilon_{i,j}(t), \epsilon_{i,j}(t) \sim N(0, \sigma^2), t\in [T]$, $\sigma = 0.1, 0.2$.

\noindent\textbf{Comparison Methods}: Here we compare our algorithm with three methods:
\begin{itemize}
\item \textbf{Greedy method}: This approach constructs the ranking based purely on previously collected data to form estimate $\widehat{\mu}_{i,j}(t)$, without regard for exploration. 

\item \textbf{$\epsilon$-greedy method ($\epsilon=0.05$)}: This method exploits the ranking list based on previously collected data, but with a probability of $\epsilon$ exploration (5\% random matching in this case), it will randomly explore other options (randomly permute the ranking list). 

\item \textbf{$1/t$-greedy method with a decaying rate $1/t$}: This technique adjusts the balance between exploring and exploiting by decreasing the exploration rate over time, specifically using a rate that inversely decays with the number of matching $t$.  
\end{itemize}

\subsubsection{Results}
\begin{figure}
    \centering
    \includegraphics[scale=0.5]{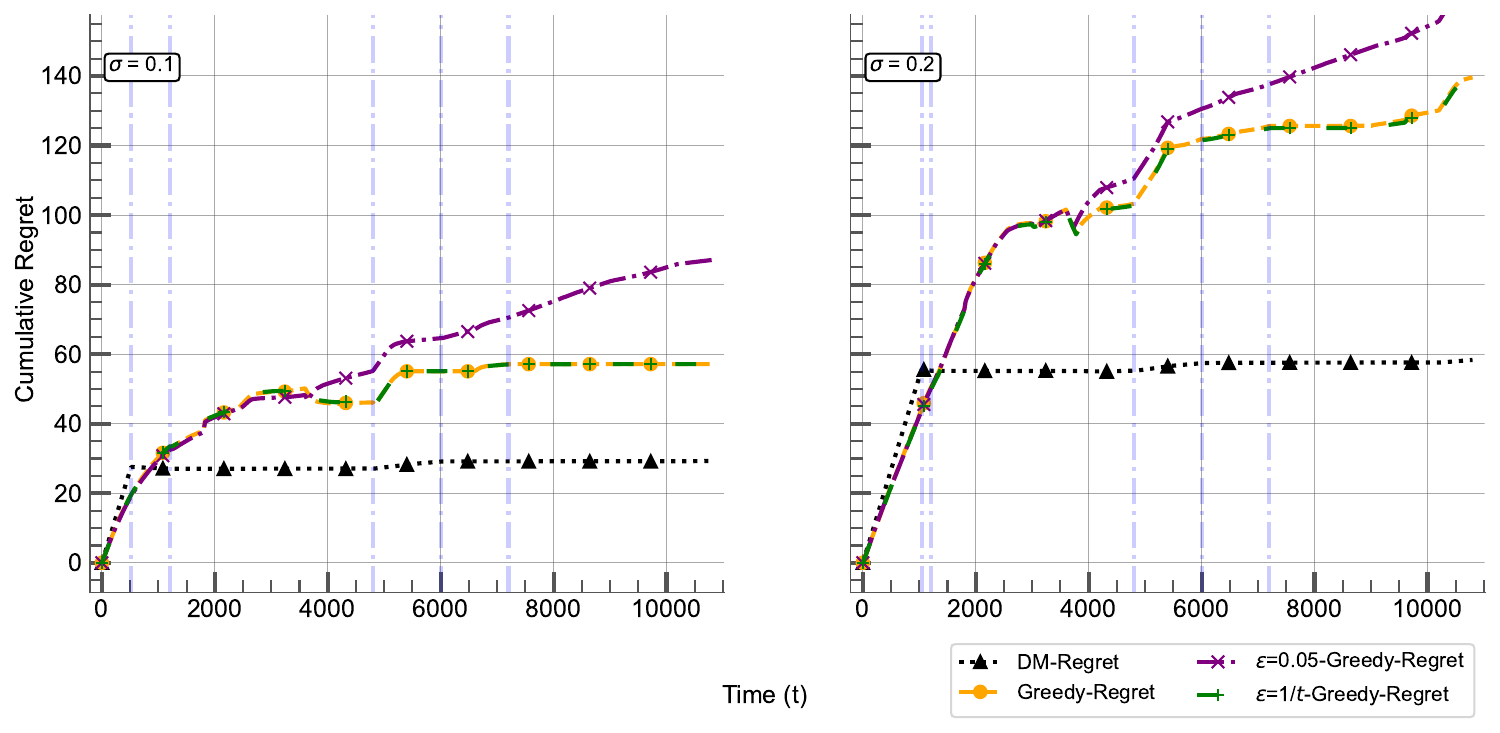}
    \caption{Total regret for agent $p_{1}$ and $p_{2}$ under noise $ \sigma = 0.1$ (Left) and $\sigma=0.2$ (Right) of methods \CCETC{}, greedy, 0.05-greedy, $1/t$-greedy.}
    \label{fig:real-data-2}
\end{figure}
In Figure \ref{fig:real-data-2}, we demonstrate the social welfare gap—a measure of the absolute difference between the optimal and actual total matching score across all agents—of different methods at various noise levels. The \CCETC{} consistently achieves the minimum social welfare gap under these conditions. The sub-optimality of other comparison methods can be attributed to their failure to utilize dynamic contextual information within the DMP to adaptively design the exploration rate $\epsilon$.

Additionally, we use a vertical dashed line to indicate the transition point of the optimal matching pattern. It is noteworthy that our findings underscore the robustness of our method in the face of changes in the optimal matching pattern, a characteristic that is absent in the greedy group method. For instance, examining the social welfare gap around the $t=5,000$ time step reveals a marked increase in the regret pattern associated with the comparison methods.

\bibliographystyle{Chicago}
\bibliography{references}

\newpage
\appendix

\setcounter{equation}{0}
\numberwithin{equation}{section}
\renewcommand\theequation{\thesection.\arabic{equation}}

\bigskip
\begin{center}
{\large\bf Appendix}
\end{center}
This appendix is organized as follows. In Section \ref{supp1:misc lemmas}, we provide the Bernstein concentration lemma and tail probability's upper bound and lower bound for the normal distribution. 
In Section \ref{app-sec:da algo}, the detail of the DA Algorithm \ref{algo:gs-algo} under the job application scenario is provided. 
In Section \ref{Appendix: Lemma 6 proof}, we prove that if agents can submit valid rankings to the platform, agents will acquire the matching which is as least as good as the stable matching.
In Section \ref{supp: Proof of Theorem 1 in appendix}, we provide detailed proof of the regret upper bound of \CCETC{}.
In Section \ref{app sec: proof of stability with high p}, we prove the stable matching holding with high probability.
In Section \ref{appendix: Example-6-cases}, the detailed instantaneous regret decomposition at time $t$ is available when we consider there are two agents and three arms in this online matching market. 
Finally, we provide detailed proof of the instance-dependent regret lower bound in Section \ref{supp: lower bound proof}.
In Section \ref{supp-sec: add sim}, we provide more experimental results of UCB method and \CCETC. In addition, various simulation settings' result is presented in Section \ref{sec-app: simulation} and real data related materials are available in Section \ref{sec-app: add real data}.

\section{Miscellaneous Lemmas}
\label{supp1:misc lemmas}
\begin{lem}[Bernstein Concentration]
    \label{lm:bernstein concentration}
    Let $\{D_{k}, \mathcal{F}_{k}\}_{k=1}^{\infty}$ be a martingale difference, and suppose that $D_{k}$ is a $\sigma$-subgaussian in an adapted sense, i.e., for all $\alpha \in \mathbb{R}$. $\mathbb{E}[e^{\alpha D_{k}}| \mathcal{F}_{k-1}] \leq e^{\frac{\alpha^2 \sigma^2}{2}}$ almost surely. Then, for all $t\geq 0$,
    \begin{equation}
        \mathbb{P}[|\sum_{k=1}^{n} D_{k}| \geq t] \leq 2e^{-\frac{t^2}{2n\sigma^2}}.
    \end{equation}
\end{lem}
Lemma \ref{lm:bernstein concentration} is from Theorem 2.3 of \citep{wainwright2019high} when $\alpha_{\ast} = \alpha_{k} = 0$ and $\nu_{k} = \sigma$ for all $k$.

\begin{lem}[Tails of Normal distribution]
\label{Tails of Normal distribution}
Let $g \sim N(0, 1)$. Then for all $t > 0$, we have
\begin{equation}
\label{tail of normal}
    \begin{aligned}
        (\frac{1}{t} - \frac{1}{t^{3}}) 
        \frac{1}{\sqrt{2\pi}} e^{(-t^2/2)} \leq \mathbb{P}(g \geq t) \leq 
        \frac{1}{t} \frac{1}{\sqrt{2\pi}} e^{(-t^2/2)}.
    \end{aligned}
\end{equation}
\end{lem}

\begin{lem}
\label{lem: covariance minimum eigen}
    With probability at most $\delta$, we have the sample covariance matrix minimum eigenvalue over $n \geq n_{0} = \log(d/\delta)/\tilde{C}_{2}(\phi_{0})$ i.i.d samples is bounded below by $\lambda_{i}/2h  + \phi_{0}^{2}/2$ with probability $1 - \delta$.
    \begin{equation}
    \begin{aligned}
        \text{Pr} \left[ \lambda_{\min}(\widehat{\Sigma}(\M{X}(n) ) > \frac{\lambda}{2n}+\frac{\phi_{0}^{2}}{2}\right] \geq 1 - 
        \exp\bigg[-\tilde{C}_{2}(\phi_{0})n + \log(d)\bigg]
    \end{aligned}
\end{equation}
where $\tilde{C}_{2}(\phi_{0}) = \min(1/2, \phi_{0}^{2}/8(x_{\max}^{2}+\lambda))$.
\end{lem}
\textit{Proof.}
First, note that
    \begin{equation}
        \begin{aligned}
            \lambda_{\max}\big(\widehat{\Sigma}(\M{X}(n))\big) &= \max_{\norm{u}=1} u \Tra  \widehat{\Sigma}(\M{X}(n)) u \\
            &= \max_{\norm{u}=1} \frac{1}{n} \sum_{t \in [n]} (X_{t}\Tra u)^{2} + \lambda\\
            &\leq x_{\max}^{2} + \lambda 
        \end{aligned}
    \end{equation}
    Then, it follows from the matrix Chernoff inequality, Corollary 5.2 in \cite{tropp2015introduction}, that
    \begin{equation}
        \begin{aligned}
            \text{Pr} \left[ \lambda_{\min}(\widehat{\Sigma}(\M{X}(n) ) > \frac{\lambda}{2n}+\frac{\phi_{0}^{2}}{2}\right]
            &\geq 
            1 - d\exp\bigg[-\frac{n\phi_{0}^{2}}{8(x_{\max}^{2}+\lambda)}\bigg]\\
            & \geq 1 -d\exp\bigg[-\tilde{C}_{2}(\phi_{0})n\bigg],
        \end{aligned}
    \end{equation}
if we take $\tilde{\delta}=1/2$ and $R = x_{\max}^{2} +\lambda$.

\section{Deferred Acceptance (DA) Algorithm}
\label{app-sec:da algo}
In algorithm \ref{algo:gs-algo}, we present the DA algorithm in the example of job seeking scenario.

\begin{algorithm}[t]
\label{algo:gs-algo}
\SetAlgoLined
	\DontPrintSemicolon
	\SetAlgoLined
	\SetKwInOut{Input}{Input}\SetKwInOut{Output}{Output}
	Input: Companies set $\N$, job applicants set $\K$, companies to job applicants' preferences, job applicants to companies' preferences.\\
	Initialize: An empty set $S$.\\
	\While{$\exists$ A company $p$ who is not matched and has not contacted to every job applicant}{
	    Let $a$ be the highest ranking job applicant in company $p$'s preference, to whom company $p$ has not yet contacted.\\
	    Now company $p$ contacts the job applicant $a$.\\
	    \uIf{Job applicant $a$ is free}{
	        $(p,a)$ become matched (add $(p, a)$ to $S$).\\
        }\uElse{
            Job applicant $a$ is matched to company $p'$ (add $(p', a)$ to $S$).\\
            \uIf{Job applicant $a$ prefers company $p'$ to company $p$}{
                Company $p$ remains free (remove ($p, a$) from $S$).\\
            }\uElse{
                Job applicant $a$ prefers company $p$ to company $p'$.\\
                Company $p'$ becomes free (remove ($p', a$) from $S$).\\
                ($p, a$) are paired (add ($p, a$) to $S$).\\
             }
        }
    }   
    Output: Matching result $S$.
	\caption{\texttt{DA Algorithm}}
\end{algorithm}

\section{Proof of Lemma \ref{lem: valid claim}}
\label{Appendix: Lemma 6 proof}
Lemma \ref{lem: valid claim} states that if all agents have valid rankings to the platform, the DA-Algorithm will provide a matching $m_{t}$ as least as good as $\overline{m}_{t}$.

\textit{Proof.}
First, we show that the agent-optimal matching $\overline{m}(t)$ is \textit{stable} according to the rankings submitted by agents when all those rankings are valid. 

Let $a_{j}$ be an arm such that $\widehat{r}_{i,j}(t) < \widehat{r}_{i, \overline{m}_{t}(i)}(t)$ for agent $p_{i}$. Since $\widehat{r}_{i,[K]}(t)$ is a valid ranking, which means that $p_{i}$ prefers $a_{j}$ over $\overline{m}_{t}(i)$ according to the true preference. However, since $\overline{m}(t)$ is \textit{stable} according to the true preference, arm $a_{j}$ must prefer agent $\overline{m}_{j}(t)^{-1}$ over $p_{i}$ because arm $a_{j}$ has no incentive to deviate the current matching $\overline{m}(t)$, where $\overline{m}_{j}(t)^{-1}$ is $a_{j}$'s matching object according to the agent-optimal $\overline{m}(t)$ or the empty set if $a_{j}$ does not have a match. Therefore, according to the ranking $\widehat{r}_{i,[K]}(t)$, $p_{i}$ has no incentive to deviate to arm $a_{j}$ because that arm $a_{j}$ would reject him. 

Since $\overline{m}(t)$ is a stable matching according to the valid ranking $\widehat{r}_{i,[K]}(t)$, we know that the DA-algorithm will output a matching which is at least as good as $\overline{m}(t)$ for all agents according to rankings $\widehat{r}_{i,[K]}(t)$ since this $\widehat{r}_{i,[K]}(t)$ ranking is an agent-optimal ranking if it were the true ranking. Since all rankings are valid rankings, it follows that the DA algorithm will output a matching $m(t)$ which is as least as good as $\overline{m}(t)$.

\section{Proof of Theorem \ref{thm: upper bound of cc-etc} - Regret Upper Bound}
\label{supp: Proof of Theorem 1 in appendix}

\subsection{Proof of Lemma \ref{lem: invalid prob}}
\label{Appendix: Lemma 2 proof}

\textit{Proof.}
We consider one time step $t$ at the exploitation step throughout this proof. 
We first show how to quantify the invalid ranking probability.

If the ranking $\widehat{r}_{i,[K]}(t)$ is \textit{invalid}, there must exist an arm $a_{j}$ where $j\neq \overline{m}_{t}(i)$ such that $\mu_{i,\overline{m}_{t}(i)}(t) > \mu_{i,j}(t)$, but $\widehat{r}_{i,j}(t) < \widehat{r}_{i, \overline{m}_{t}(i)}(t)$, due to the inaccurate estimation of the true parameter, which is equivalent to $\widehat{\mu}_{i,j}(t) > \widehat{\mu}_{i, \overline{m}_{t}(i)}(t)$. So we have  
\begin{equation}
    \begin{aligned}
    &\mathbb{P}(\widehat{\mu}_{i,j}(t) > \widehat{\mu}_{i, \overline{m}_{t}(i)}(t))\\
    &= \mathbb{P}\Big[
    \widehat{\mu}_{i,j}(t) 
    - \mu_{i,j}(t)
    - \widehat{\mu}_{i,\overline{m}_{t}(i)}(t) 
    +\mu_{i,\overline{m}_{t}(i)}(t) 
    \geq 
    \mu_{i,\overline{m}_{t}(i)}(t)
    - \mu_{i,j}(t)
    \Big]\\
    & =
    \mathbb{P}\Big[
    \widehat{\theta}_{i}(h)\Tra x_{j}(t)
    - \theta_{i, *}\Tra x_{j}(t)
    - \widehat{\theta}_{i}(h)\Tra x_{\overline{m}_{t}(i)}(t) 
    + \theta_{i, *}\Tra x_{\overline{m}_{t}(i)}(t) 
    \geq 
    \mu_{i,\overline{m}_{t}(i)}(t)
    - \mu_{i,j}(t)
    \Big]\\
    & = 
    \mathbb{P}
    \Big[
    (\widehat{\theta}_{i}(h) -  \theta_{i, *})\Tra(x_{j}(t)- x_{\overline{m}_{t}(i)}(t))
    \geq 
    \theta_{i,*}\Tra(x_{\overline{m}_{t}(i)}(t) - x_{j}(t))
    \Big]\\
    &\leq
    \mathbb{P}
    \Big[
    \norm{\widehat{\theta}_{i}(h) -  \theta_{i, *}}_{2}\norm{x_{j}(t)- x_{\overline{m}_{t}(i)}(t)}_{2} \geq 
    \theta_{i,*}\Tra(x_{\overline{m}_{t}(i)}(t) - x_{j}(t))
    \Big]\\
    &= \mathbb{P}
    \Big[
    \norm{\widehat{\theta}_{i}(h) -  \theta_{i, *}}_{2}
    \geq 
    \langle 
    \theta_{i,*},
    \frac{x_{\overline{m}_{t}(i)}(t) - x_{j}(t)}{\norm{ x_{\overline{m}_{t}(i)}(t)- x_{j}(t)}_{2}}
    \rangle
    \Big],
    \end{aligned}
\end{equation}
where in the inequality, we use the Cauchy inequality to upper bound the left inner product. Here we find an interesting term called, \textit{similarity difference} (SD), which is
\begin{equation}
    \begin{aligned}
        \text{SD} 
        &\overset{\Delta}{=} \langle \theta_{i,*}, \frac{x_{\overline{m}_{t}(i)}(t) - x_{j}(t)}{\norm{ x_{\overline{m}_{t}(i)}(t)- x_{j}(t)}_{2}} \rangle  \\
        & =
        \norm{\theta_{i,*}}_{2} \langle \frac{\theta_{i,*}}{\norm{\theta_{i,*}}_{2}}, \frac{x_{\overline{m}_{t}(i)}(t) - x_{j}(t)}{\norm{ x_{\overline{m}_{t}(i)}(t)- x_{j}(t)}_{2}} \rangle \\
        & =\norm{\theta_{i,*}}_{2}\cos({\phi_{i,j}(t)}), 
    \end{aligned}
\end{equation}
where $\phi_{i,j}(t)$ represents the \textit{angle} between the normalized true parameter $\theta_{i,*}$ and the normalized arms difference at time step $t$, which is the similarity difference between arm $a_{j}$ and arm $a_{\overline{m}_{t}(i)}$ from the viewpoint of agent $p_{i}$.

Here we discuss the boundary scenario of the similarity difference.
If SD = 0, there are three possible reasons. 
\begin{enumerate}
    \item The first possible reason is that if the true parameter $\theta_{i, *} = 0$. Since we assume all agents' true parameters are meaningful and positive, with Assumption \ref{ass: positive theta}, we can rule out this case.
    
    \item The second possible reason is that if arm $a_{j}$ and arm $a_{\overline{m}_{t}(i)}$ are identical such that $x_{j}(t) = x_{\overline{m}_{t}(i)}$. Since we assume all arms are different, we can also rule out this case.
    
    \item The third possible reason is that if $\cos({\phi_{i,j}(t)}) = 0$. That means from the view point of agent $p_{i}$ at time $t$, arm $a_{j}$ and arm $a_{\overline{m}_{t}(i)}$ are symmetric. Since we assume there are no ties in ranking over time, we can also rule out this scenario.
\end{enumerate}
The last case we also discussed in Assumption \ref{ass: mirror Margin} where we assume that the uniform sub-optimal minimal condition over time is greater than zero. That means there is no symmetric case for the agent to distinguish two arms between the agent-optimal and the sub-optimal arm—\textit{That is the key difference between the DMP and the MAB competing bandit problem. The MAB competing bandit only has one constant gap over time and no existence of the interesting symmetric arms.} 
We now restate the \textit{uniform sub-optimal minimal condition} $\overline{\Delta}_{i, \min}$ for agent $p_{i}$ over time $t$, that is
$\overline{\Delta}_{i, \min} = \underset{j \in [K], t \in [T]}{\min}\norm{\theta_{i,*}}_{2}\cos({\phi_{i,j}(t)}) > 0$.

With Assumption \ref{ass: mirror Margin}, we consider the estimation error of the true parameter is lower bounded by the \textit{uniform sub-optimal minimal condition} $\overline{\Delta}_{i, \min}$. So the probability of the invalid ranking is upper bounded by
\begin{equation}
\label{supp: lemma 1 - tail prob}
    \begin{aligned}
        \mathbb{P}
        \Bigg[
        \norm{\widehat{\theta}_{i}(h) -  \theta_{i, *}}_{2}
        \geq 
        \norm{\theta_{i,*}}_{2} \cos({\phi_{i,j}(t)})
        \Bigg]
        \leq 
        \mathbb{P}
        \Bigg[
        \norm{\widehat{\theta}_{i}(h) -  \theta_{i, *}}_{2}
        \geq 
        \overline{\Delta}_{i, \min}
        \Bigg].
    \end{aligned}
\end{equation}
To get the upper bound of this tail event's probability, we use the technique of quantifying the confidence ellipsoid from \citep{li2021online}. Notation $\widehat{\Sigma}(\M{X}_{i}(t))$ represents the normalized covariance matrix, so $\widehat{\Sigma}(\M{X}_{i}(t)) = \Phi_{i}(t)/t = (\M{X}_{i}(t)\Tra \M{X}_{i}(t) + \lambda_{i}\M{I}_{d})/t$ for $t\geq 1$, where we define $\Phi_{i}(0) = \lambda_{i}\M{I}_{d}$ and $\lambda_{i}$ is the prespecified penalty hyperparameter for agent $p_{i}$. Note that the event $\lambda_{\min}(\widehat{\Sigma}(\M{X}_{i}(t))) \geq \phi_{0}^{2}/2 + \lambda_{i}/2t$ holds for $t\geq 1$ with having that $\lambda_{\min}\big(\M{X}_{i}(t)\Tra \M{X}_{i}(t)\big)/t \geq \phi_{0}^{2}/2$, based on the high probability in exponential decay wrt $t$ (see Corollary 5.2 in \citep{tropp2015introduction} and Lemma \ref{lem: covariance minimum eigen}. Thus after the learning step, agents have already gathered length $h$ historical data, which include actions, rewards and contexts.  For notation simplicity, we use $\widehat{\theta}_{i}$ to replace $\widehat{\theta}_{i}(h)$ and $\M{X}_{i}$ to replace $\M{X}_{i}(h)$. So we have
\begin{equation}
    \begin{aligned}
        &\norm{\widehat{\theta}_{i} - \theta_{i, *}}_{2} \\
        &=\norm{(\M{X}_{i}\Tra \M{X}_{i} + \lambda_{i}\M{I})^{-1}\M{X}_{i}\Tra(\M{X}_{i} \theta_{i, *} + \epsilon) - \theta_{i, *} }_{2} 
        \\
        &=\norm{(\M{X}_{i}\Tra \M{X}_{i} + \lambda_{i}\M{I})^{-1}\M{X}_{i}\Tra \epsilon 
        + \theta_{i, *} 
        -\lambda_{i}(\M{X}_{i}\Tra \M{X}_{i} + \lambda_{i}\M{I})^{-1}\theta_{i, *} 
        - \theta_{i, *}
        }_{2}
        \\
        &=\norm{(\M{X}_{i}\Tra \M{X}_{i} + \lambda_{i}\M{I})^{-1}(\M{X}_{i}\Tra \epsilon - \lambda_{i}\theta_{i, *})
        }_{2}
        \\
        &\leq \frac{1}{\lambda_{i}+ h\phi_{0}^{2}} \norm{\M{X}_{i}\Tra \epsilon - \lambda_{i}\theta_{i, *}}_2.
    \end{aligned}
\end{equation}
Here we use a constant $\chi > 0$ to get the estimation error. So we have 
\begin{equation}
    \begin{aligned}
        &\text{Pr}\left[ \norm{\widehat{\theta}_{i} - \theta_{i, *}}_{2} \leq \chi \right]\\
        &\geq
        \text{Pr} \left[\bigg(\norm{\M{X}_{i} \Tra \epsilon - \lambda_{i}\theta_{i, *}}_{2} \leq 2\chi (\lambda_{i}+ h\phi_{0}^{2})\bigg) \cap \bigg(\lambda_{\min}(\widehat{\Sigma}(\M{X}_{i} ) > \frac{\lambda_{i}}{2h}+\frac{\phi_{0}^{2}}{2}\bigg)\right] \\
        &\geq  
        1 - \sum_{r=1}^{d} \underbrace{\text{Pr}\left[ \epsilon \Tra \M{X}^{(r)}_{i}  > \lambda_{i} \theta_{i, *}^{(r)} + \frac{2\chi (\lambda_{i}+ h\phi_{0}^{2})}{\sqrt{d}}\right]}_{\text{Part I}} - \underbrace{\text{Pr}\left[ \epsilon \Tra \M{X}^{(r)}_{i} < \lambda_{i} \theta_{i, *}^{(r)} - \frac{2\chi (\lambda_{i}+ h\phi_{0}^{2})}{\sqrt{d}}\right]}_{\text{Part II}} \\
        &\hspace{2cm} - \text{Pr}\left[ \lambda_{\min}(\widehat{\Sigma}(\M{X}_{i}) \leq \frac{\lambda_{i}}{2h}+\frac{\phi_{0}^{2}}{2} \right] 
    \end{aligned}
\end{equation}
where we let $\M{X}^{(r)}_{i}(t)$ denote the $r^{\text{th}}$ column of $\M{X}_{i}(t)$. 
To make $\text{Pr}[ \epsilon \Tra \M{X}^{(r)}_{i} < \lambda_{i} \theta_{i, *}^{(r)} - \frac{2\chi (\lambda_{i}+ h\phi_{0}^{2})}{\sqrt{d}}]$ have a relative small probability, based on the Assumption \ref{ass: positive theta} that $\theta_{i, *}^{(r)}$ is positive and let $\lambda_{i} < \frac{2\chi (\lambda_{i}+ h\phi_{0}^{2})}{\sqrt{d}\theta_{i, *}^{(r)}}$, with the analysis from Case B.2.3 from \citep{li2021online}, the part II's probability will be small. So when $\lambda_{i} < \frac{2\chi (\lambda_{i}+ h\phi_{0}^{2})}{\sqrt{d}\theta_{i, *}^{(r)}}$ is small, part I and part II's probability will be similar\footnote{Or we can follow \citep{li2021online}'s analysis for part I and part II separately. However, based on the Assumption \ref{ass: positive theta}, the probability difference is minor.}. So the previous probability lower bound will be 
\begin{equation}
    \begin{aligned}
        &\text{Pr}\left[ \norm{\widehat{\theta}_{i} - \theta_{i, *}}_{2} \leq \chi \right]\\
        &\geq  
        1 - \sum_{r=1}^{d} 2\text{Pr}\left[ \epsilon \Tra \M{X}^{(r)}_{i}  > \lambda_{i} \theta_{i, *}^{(r)} + \frac{2\chi (\lambda_{i}+ h\phi_{0}^{2})}{\sqrt{d}}\right] - \text{Pr}\left[ \lambda_{\min}(\widehat{\Sigma}(\M{X}_{i} ) \leq \frac{\lambda_{i}}{2h}+\frac{\phi_{0}^{2}}{2} \right] 
    \end{aligned}
\end{equation}
We can expand $\epsilon\Tra \M{X}^{(r)}_{i}(t) = \sum_{j \in [t]} \epsilon(j) x_{i,j}^{(r)}$, where we note that $D_{i, j, r} \equiv \epsilon(k) x_{i,j}^{(r)}$ is a $x_{i,\max} \sigma$-subgaussian random variable, where $x_{i,\max} = \norm{\M{X}_{i}(t)}_{\infty}$, conditioned on the sigma algebra $\mathcal{F}_{j-1}$ that is generated by random variable $X_1,..., X_{j-1}, Y_{1},..., Y_{j-1}$. Defining $D_{i, 0, r} = 0$, the sequence $D_{i, 0, r}, D_{i, 1, r},..., D_{(i,j,r)}$ is a martingale difference sequence adapted to the filtration $\mathcal{F}_{1} \subset \mathcal{F}_{2} \subset ... \mathcal{F}_{j}$, since $E[\epsilon(j)x_{j}^{(r)}|\mathcal{F}_{j-1}] = 0$. Using the Bernstein concentration inequality from Lemma \ref{lm:bernstein concentration},
\begin{equation}
    \begin{aligned}
        &\text{Pr}\left[ \norm{\widehat{\theta}_{i} - \theta_{i, *}}_{2} \leq \chi  \right]
        \\
        &\geq 
        1 - \sum_{r=1}^{d} 2\text{Pr}\left[ \epsilon \Tra \M{X}^{(r)}_{i}  > \lambda_{i} \theta_{i, *}^{(r)} + \frac{2\chi (\lambda_{i}+ h\phi_{0}^{2})}{\sqrt{d}}\right] - \text{Pr}\left[ \lambda_{\min}(\widehat{\Sigma}(\M{X}_{i} ) \leq \frac{\lambda_{i}}{2h}+\frac{\phi_{0}^{2}}{2} \right]  \\
        & \geq 
        1 - \sum_{r=1}^{d} 2\text{Pr}\left[ \epsilon \Tra \M{X}^{(r)}_{i} >  \frac{2
        h \chi \lambda_{i}\phi_{1}^{2}
        (h, \lambda_{i})}{\sqrt{d}}\right]  - \text{Pr}\left[ \lambda_{\min}(\widehat{\Sigma}(\M{X}_{i} ) \leq \frac{\lambda_{i}}{2h}+\frac{\phi_{0}^{2}}{2} \right] \\
        &\geq 1 - 2d\exp{\left[-\frac{2h\chi^{2} \lambda_{i}^{2}\phi_{1}^{4}}{d\norm{\M{X}_{i}}_{\infty}^{2}\sigma^2} \right]} -
        \text{Pr}\left[ \lambda_{\min}(\widehat{\Sigma}(\M{X}_{i} ) \leq \frac{\lambda_{i}}{2h}+\frac{\phi_{0}^{2}}{2} \right],
    \end{aligned}    
\end{equation}
where we denote $\phi_{1}^{2} := \phi_{1}^{2}(h, \lambda_{i}) = (\lambda_{i}+ h\phi_{0}^{2})/(h\lambda_{i}) = 1/h + \phi_{0}^{2}/\lambda_{i}$. 
So we have the probability upper bound for the estimation error for any constant $\chi > 0$,
\begin{equation}
    \begin{aligned}
         &\text{Pr}\left[ \norm{\widehat{\theta}_{i} - \theta_{i, *}}_{2} \geq \chi  \right]
        \\
        & \leq 2d\exp{\left[-\frac{2h\chi ^{2} \lambda_{i}^{2}\phi_{1}^{4}}{d\norm{\M{X}_{i}}_{\infty}^{2}\sigma^2} \right]} -
        \text{Pr}\left[ \lambda_{\min}(\widehat{\Sigma}(\M{X}_{i}) \leq \frac{\lambda_{i}}{2h}+\frac{\phi_{0}^{2}}{2} \right].
    \end{aligned}
\end{equation}
Now we replace $\chi$ with $\overline{\Delta}_{i, \min}$, and we have $x_{i, \max} \leq 1$ by Assumption \ref{ass: bounded covariates}.  So we get the following upper bound of the invalid ranking probability,
\begin{equation}
    \begin{aligned}
        &\mathbb{P}
        \Big[
        \norm{\widehat{\theta}_{i}(h) -  \theta_{i, *}}_{2}
        \geq 
        \overline{\Delta}_{i, \min}
        \Big] \\
        &\leq 
        \exp\Bigg[-h\frac{2\lambda_{i}^{2}\rho^{2} \phi_{1}^{4}}{d^2x_{i, \max}^{2}\sigma^2}  + \log(2d)\Bigg] -
        \mathbb{P}
        \Big[\lambda_{\min}(\widehat{\Sigma}(\M{X}_{i}(h)) 
        \leq 
        \frac{\lambda_{i}}{2h}+\frac{\phi_{0}^{2}}{2}
        \Big] \\
        &\lesssim  
        \exp\Bigg[-h\frac{2\lambda_{i}^{2}\rho^{2}\phi_{1}^{4}}{d^2\sigma^2} + \log(2d)
        \Bigg].
    \end{aligned}
\end{equation}
So the invalid ranking's probability created by agent $p_{i}$ at time $t$ is upper bounded by
\begin{equation}
    \begin{aligned}
     \mathbb{P}(\widehat{\mu}_{i,j}(t) > \widehat{\mu}_{i, \overline{m}_{t}(i)}(t)) \lesssim   
     2d\exp\Bigg[-h\frac{2\lambda_{i}^{2}\rho^{2}\phi_{1}^{4}}{d^2\sigma^2}\Bigg],
    \end{aligned}
\end{equation}
and because we consider all such sub-optimal arms $a_{j}$, we have the following upper bound of the invalid ranking probability,
\begin{equation}
    \begin{aligned}
     \mathbb{P}(\widehat{r}_{i,[K]}(t)  \text{ is invalid})  \leq 
     2d (K-\tau_{i}(t))\exp\Bigg[-h\frac{2\lambda_{i}^{2}\rho^{2}\phi_{1}^{4}}
     {d^2\sigma^2}
        \Bigg],
    \end{aligned}
\end{equation}
where we use $\tau_{i}(t)$ to represent the agent $p_{i}$'s optimal ranking position when matched with $\overline{m}_{t}(i)$. 

With Lemma \ref{lem: invalid prob}, we can quantify the regret at $t > h$. So the instantaneous regret for agent $p_{i}$ at time $t$ will be upper bounded by 
\begin{equation}
    \begin{aligned}
        R_{i,t}
        & \overset{\Delta}{=} \overline{\Delta}_{i,j}(t) \mathbb{P}(\text{at least one } \widehat{r}_{i,[K]}(t)  \text{ is invalid}, \forall i \in [N])\\
        & \leq N (K-\min_{i\in [N]}\tau_{i}(t))\overline{\Delta}_{i,\max}(t) \mathbb{P}(\widehat{r}_{i,[K]}(t) \text{ is invalid}).
    \end{aligned}
\end{equation} 
Then we add the part I regret and part II regret together and get the regret upper bound of \CCETC{}.
\begin{equation}
    \begin{aligned}
        R_{i}(n) &\leq 
        \sum_{t=1}^{h} \overline{\Delta}_{i,j}(t)    + 
        2Nd 
        \Bigg[\sum_{t = h + 1}^{T}\overline{\Delta}_{i, \max}(t)(K-\min_{i\in [N]}\tau_{i}(t))\Bigg]
        \exp\Bigg[-\frac{2\lambda_{i}^{2}\rho^{2}\phi_{1}^{4 }}{d^2\sigma^2}h 
        \Bigg].
    \end{aligned}
\end{equation}
By $\phi_{1}^{4} = (1/h + \phi_{0}^{2}/\lambda_{i})^{2} = \frac{1}{h^{2}} + \frac{2\phi_{0}^{2}}{\lambda_{i}h} + \frac{\phi_{0}^{4}}{\lambda_{i}^{2}}$,  we have $\phi_{1}^{4}h = \frac{1}{h} + \frac{2\phi_{0}^{2}}{\lambda_{i}} + h\frac{\phi_{0}^{4}}{\lambda_{i}^{2}} \geq \frac{2\phi_{0}^{2}}{\lambda_{i}} + h\frac{\phi_{0}^{4}}{\lambda_{i}^{2}}$,

the regret upper bound is
\begin{equation}
    \begin{aligned}
        R_{i}(n) &\leq 
        \sum_{t=1}^{h} \overline{\Delta}_{i,j}(t)    + 
        2Nd
        \Bigg[\sum_{t = h + 1}^{T}\overline{\Delta}_{i, \max}(t)(K-\min_{i\in [N]}\tau_{i}(t))\Bigg]
        \exp\Bigg[-\frac{2\lambda_{i}^{2}\rho^{2}\phi_{1}^{4}}{d^2\sigma^2}h
        \Bigg]\\
        &<
        \sum_{t=1}^{h} \overline{\Delta}_{i,j}(t)    + 
        2N d
        \Bigg[\sum_{t = h + 1}^{T}\overline{\Delta}_{i, \max}(t)(K-\min_{i\in [N]}\tau_{i}(t))\Bigg]
        \exp\Bigg[-\frac{2\lambda_{i}^{2}\rho^{2}}{d^2\sigma^2}(\frac{2\phi_{0}^{2}}{\lambda_{i}} + h\frac{\phi_{0}^{4}}{\lambda_{i}^{2}})
        \Bigg]\\
        &=
        \sum_{t=1}^{h} \overline{\Delta}_{i,j}(t)    + 
        2C_{0}(\lambda_{i}) N d
        \Bigg[\sum_{t = h + 1}^{T}\overline{\Delta}_{i, \max}(t)(K-\min_{i\in [N]}\tau_{i}(t))\Bigg]
        \exp\Bigg[-\frac{2\phi_{0}^{4} \rho^{2}}{d^2\sigma^2}h  
        \Bigg]
    \end{aligned}
\end{equation}
where $C_{0}(\lambda_{i}) = \exp\bigg[-\frac{4\lambda_{i}\phi_{0}^{2}\rho^{2}}{d^2\sigma^2} \bigg]$.

\subsection{Proof of Corollary \ref{coro: upper order}}
\label{Appendix: order analysis of upper bound of CC-ETC}
\textit{Proof.}
Moreover, in order to analyze the order of the regret upper bound, we optimize the  the exploration horizon,
\begin{equation}
    \label{eq:h_bar}
    \begin{aligned}
        R_{i}(n) 
        &\leq 
        \sum_{t=1}^{h} \overline{\Delta}_{i,j}(t)   + 
        2C_{0}(\lambda_{i})Nd
        \Bigg[\sum_{t = 1}^{T}\overline{\Delta}_{i, \max}(t)(K-\min_{i\in [N]}\tau_{i}(t))\Bigg]
        \exp\Bigg[-\frac{2\phi_{0}^{4}\rho^{2}}{d^2\sigma^2}h
        \Bigg]\\
        & \leq 
        h \overline{\Delta}_{i, \max}  + 
        2C_{0}(\lambda_{i}) Nd 
        \Bigg[\sum_{t = 1}^{T}\overline{\Delta}_{i, \max}(t)(K-\min_{i\in [N]}\tau_{i}(t))\Bigg]
        \exp\Bigg[-\frac{2\phi_{0}^{4}\rho^{2}}{d^2\sigma^2}h 
        \Bigg] \\
        & \leq 
        h \overline{\Delta}_{i, \max}  +
        2C_{0}(\lambda_{i}) N K d T \overline{\Delta}_{i, \max}
        \exp\Bigg[-\frac{2\phi_{0}^{4}\rho^{2}}{d^2\sigma^2}h 
        \Bigg],
    \end{aligned}
\end{equation}
where we know that $\overline{\Delta}_{i,j}(t) \leq \overline{\Delta}_{i, \max}(t) \leq \overline{\Delta}_{i, \max}, \forall t \in [T]$ and $K-\tau_{i}(t) < K$. Taking the derivative on the RHS of Eq. \eqref{eq:h_bar} with respect to $h$ to obtain the optimal $h$,
\begin{equation}
    \begin{aligned}
        \overline{\Delta}_{i, \max}  +
        2C_{0}(\lambda_{i}) N K d T \overline{\Delta}_{i, \max} 
        \exp\Bigg[-\frac{2\phi_{0}^{4}\rho^{2}}{d^2\sigma^2}h  \Bigg] 
        \times (-\frac{2\phi_{0}^{4}\rho^{2}}{d^2\sigma^2})= 0,
    \end{aligned}
\end{equation}
and get       
\begin{equation}
    \begin{aligned}
    h= \frac{d^2\sigma^2}{ 2\phi_{0}^{4}\rho^{2} } \log \frac{4C_{0}(\lambda_{i})TNK \phi_{0}^{4} \rho^{2} }{d \sigma^2\overline{\Delta}_{i, \max} },
    \end{aligned}
\end{equation}
when we set the optimal learning step to $h \leftarrow  \left\lceil h  \right\rceil$, we can achieve the minimum regret,
\begin{equation}
    \begin{aligned}
        R_{i}(T) &\leq \max \bigg\{h\overline{\Delta}_{i,\max}, \frac{d^2\sigma^2 \overline{\Delta}_{i,\max}}{ 2\phi_{0}^{4}\rho^{2} } \log \frac{4C_{0}(\lambda_{i})NK \phi_{0}^{4} \rho^{2}}{d \sigma^2\overline{\Delta}_{i,\max}}T\bigg\}
        + \frac{d^2 \sigma^2\overline{\Delta}_{i,\max}}{2\phi_{0}^{4}\rho^{2}}\\
        &= C_{1}(d, \sigma, \overline{\Delta}_{i, \min},\overline{\Delta}_{i,\max}, \lambda_{i}, x_{i,\max})
        \log \Bigg[C_{2}(N, K, d, \sigma, \overline{\Delta}_{i, \min}, \overline{\Delta}_{i,\max}, \lambda_{i}, x_{i,\max}) \times T \Bigg]\\
        &\quad+ C_{1}(d, \sigma, \overline{\Delta}_{i, \min}, \overline{\Delta}_{i,\max},\lambda_{i}, x_{i,\max})\\
        &=\mathcal{\tilde{O}}\bigg(\frac{d^2\sigma^2}{\rho^{2}}\log(NKT)\bigg)
    \end{aligned}
\end{equation}
where the constants $C_{1}, C_{2}$ are given by
\begin{equation}
C_{1} =  \frac{d^2 \sigma^2\overline{\Delta}_{i,\max}}{2\phi_{0}^{4}\rho^{2}},
        \qquad
C_{2} = \frac{4C_{0}(\lambda_{i})NK \phi_{0}^{4} \rho^{2}}{d \sigma^2\overline{\Delta}_{i,\max}}.
\end{equation}

\section{Proof of Theorem \ref{thm: stability of ccetc} - Stable Matching}
\label{app sec: proof of stability with high p}
\textit{Proof.} Based on Lemmas \ref{lem: valid claim} and \ref{lem: invalid prob}, as long as all agents have valid rankings, then the matching solution is stable. In order to have the $ \mathbb{P}(\text{matching solution is stable}) \geq \Psi$, we have
\begin{equation}
    \begin{aligned}
        \mathbb{P}(\text{Matching solution is stable}) &= \mathbb{P}(\text{all agents have valid rankings}) \\
        &\geq \prod_{i=1}^{N}\bigg[1 - 2d(K-\tau_{i}(t))\exp\bigg(-t\frac{2\lambda_{i}^{2}\rho^{2}\phi_{1,i}^{4}}{d^2\sigma^2}\bigg)\bigg]\\
        &\geq \bigg[1 - 2d(K-1)\exp\bigg(-t\frac{2\rho^{2}\phi_{0}^{4}}{d^2\sigma^2}\bigg)\bigg]^{N}. 
    \end{aligned}
\end{equation}
Thus, given $t \geq \lceil{\frac{d^2 \sigma^2}{2 \rho^2\phi_{0}^4}[\log(2d(K-1)) -  \log(1 - \Psi^{1/N})] \rceil}$ based on Corollary \ref{coro: upper order}, we have the matching solution provided by \CCETC{} is stable with probability at least $\Psi$.

\section{Detailed Regret Analysis for Two Agents and Three Arms}
\label{appendix: Example-6-cases}
\begin{figure}[t]
    \centering
    \includegraphics[scale=.2]{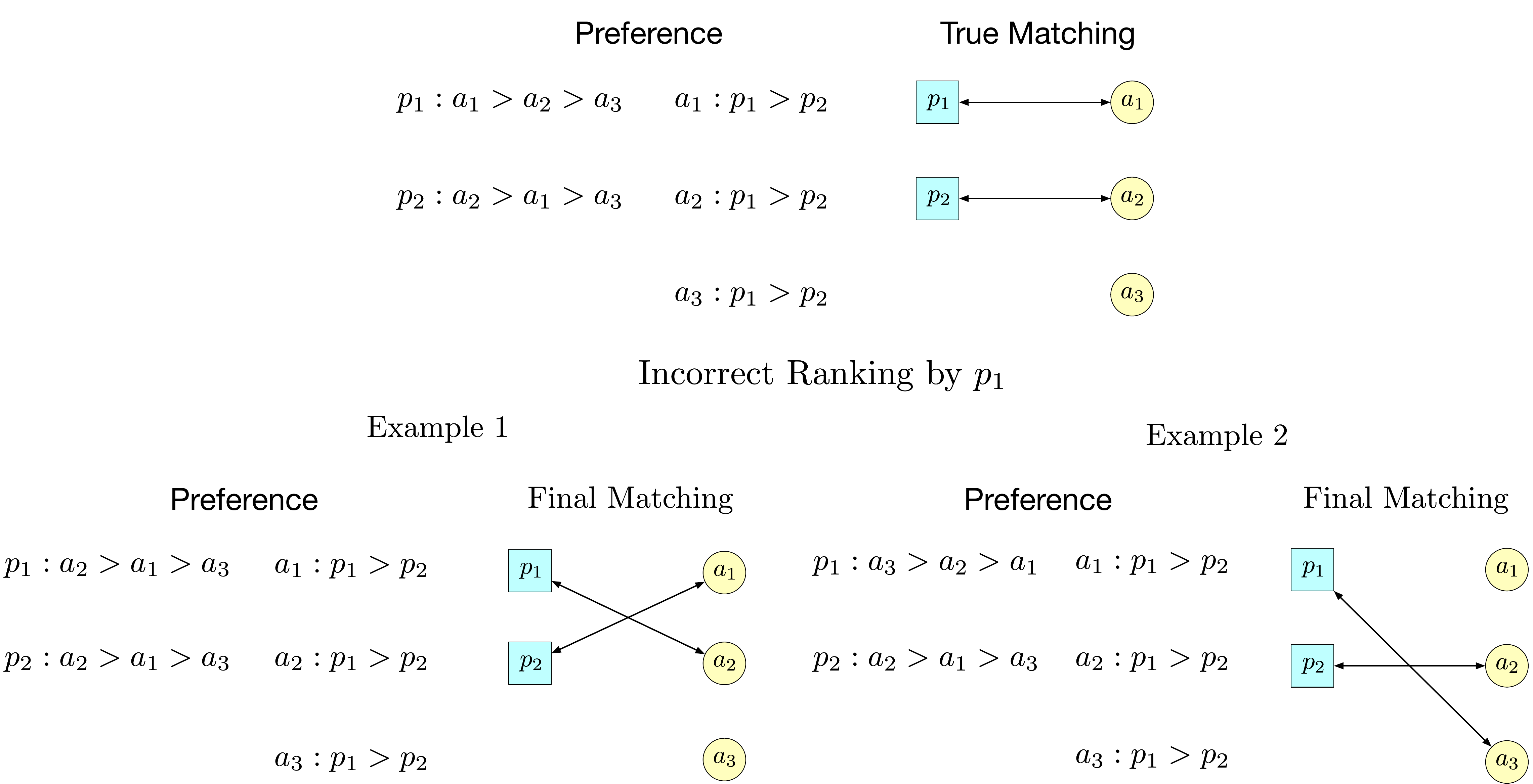}
    \caption{Examples of the matching result caused by the incorrect ranking provided by agent $p_{1}$ when agent $p_{2}$ submits the  correct ranking list under the global preference. In Example 1, Agent $p_{1}$ provides an incorrect ranking $p_{1}: a_{2} > a_{1} > a_{3}$. The final matching result is $\{(p_{1}, a_{2}), (p_{2}, a_{2})\}$. It creates a positive regret for both agents. In Example 2: Agent $p_{1}$ provides an incorrect ranking $p_{1}: a_{3} > a_{2} > a_{1}$. The final matching result is
    $\{(p_{1}, a_{3}), (p_{2}, a_{2})\}$. It creates a positive regret for $p_{1}$ and no regret for $p_{2}$.}
    \label{fig:example-global}
\end{figure}

\begin{figure}[h]
    \centering
    \includegraphics[scale=.2]{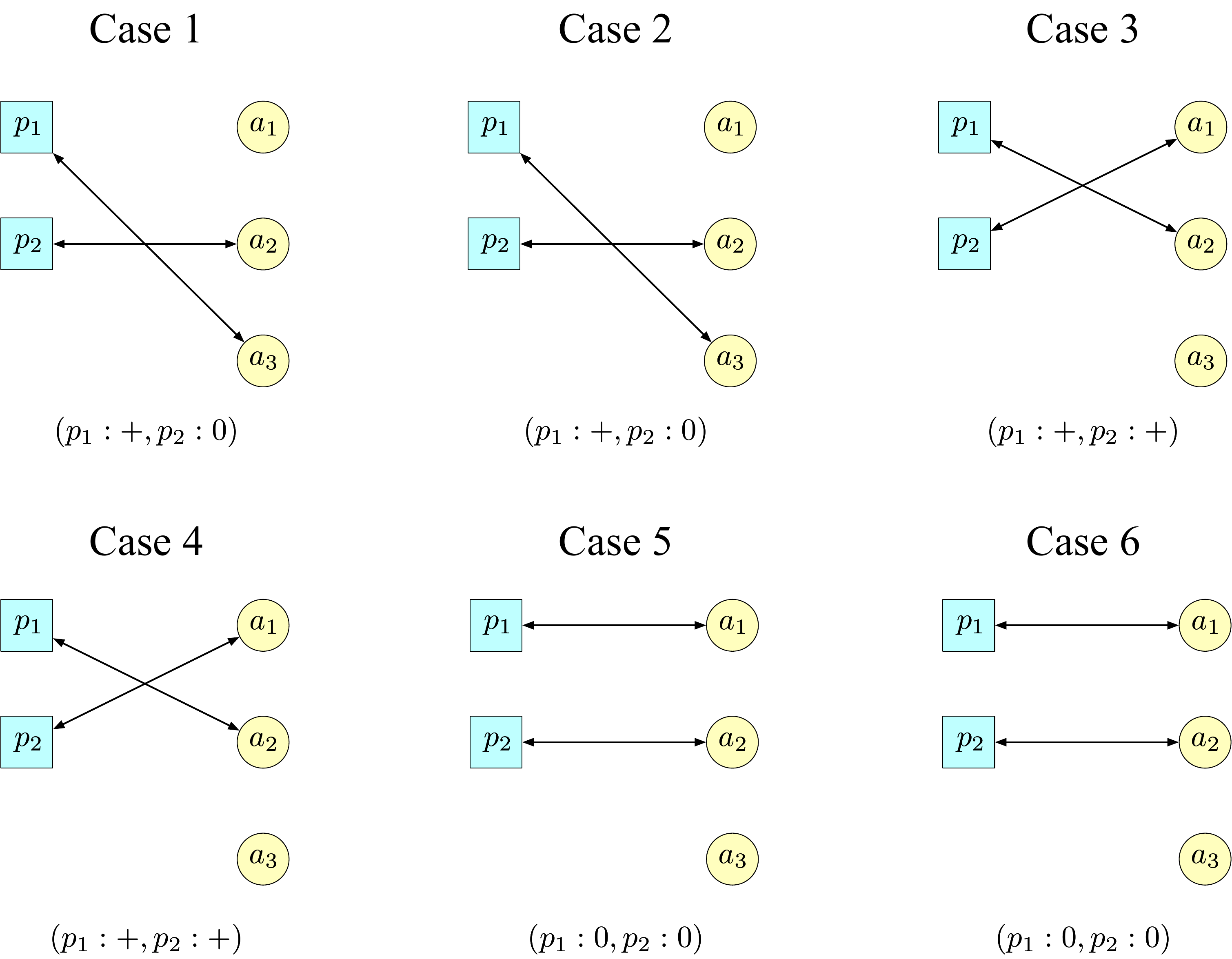}
    \caption{The corresponding matching results and regret status in six cases when agent $p_{1}$ submits an incorrect ranking. \textit{Single agent suffers regret}: Case 1 and Case 2. 
    \textit{Both agents suffer regret}: Case 3 and Case 4. 
    \textit{No regret}: Case 5 and Case 6.}
    \label{fig:6-cases-old}
\end{figure}

The expected instantaneous regret $R_{1,t}(\widehat{r}_{2}(t) = r_{2}(t)) = \mathbb{P}(\mathcal{G}_{2}(t))
\sum_{z=1}^{6} \mathbb{P}^{C_{z}}_{t} R_{1,t}^{C_{z}}(\widehat{r}_{2}(t) = r_{2}(t))$, 
where $\mathcal{G}_{2}(t)$ is the correct ranking. $\mathbb{P}^{C_{z}}_{t}$ is the probability of occurring matching case $z$ at time $t$. $R_{1,t}^{C_{z}}(\widehat{r}_{2}(t) = r_{2}(t))$ is the conditional instantaneous regret of occurring matching case $z$ at time $t$ if $p_{2}$ submits correct ranking list. 
Meanwhile $\sum_{z=1}^{6} \mathbb{P}^{C_{z}}_{t} R_{1,t}^{C_{z}}(\widehat{r}_{2}(t) = r_{2}(t))$ represents the expected regret for $p_{1}$ when $p_{2}$ submits the correct ranking list. 
We find that there are six cases in total if $p_{2}$ submits the correct ranking list shown in Figures \ref{fig:example-global} and \ref{fig:6-cases-old}.


After collecting all probabilities' lower bounds, we can compute the instantaneous regret for agent $p_{1}$ at time $t$. Then we can sum all instantaneous regret to get the regret lower bound. 

Due to the \textit{incorrect raking} from agent $p_{1}$, it creates six cases in total. In the following passage, we will analyze them case by case.

\textit{Case 1.} If agent $p_{1}$ wrongly estimates the ranking over arms as $p_{1}: a_{3} > a_{1} > a_{2}$, the matching result by DA Algorithm is shown in Figure \ref{fig:6-cases} Case 1.
Agent $p_{1}$ is matched with $a_{3}$ and agent $p_{2}$ is matched with $a_{2}$. In this case  $p_{1}$ suffers a positive regret. The instantaneous regret can be decomposed as
\begin{equation}
    \begin{aligned}
        R_{1,t}^{C_{1}} \overset{\Delta}{=} \theta_{1,*}\Tra x_{1}(t) 
        -\theta_{1,*}\Tra x_{3}(t) 
        = \theta_{1,*}\Tra(x_{1}(t) - x_{3}(t)),
    \end{aligned}
\end{equation}
where we define $R_{1,t}^{C_{1}}$ is the case 1 instantaneous regret for agent $p_{1}$ at time $t$. Here $C_{1}$ represents the case 1, ``1" in the subscript represents agent $p_{1}$, and $t$ in the subscript represents the time step. 
Similar definitions are used in the following analysis. In addition, agent $p_{2}$ does not suffer regret in case 1. 

This incorrect ranking's joint probability for agent $p_{1}$ is the product of two ranking probabilities $\mathbb{P}_{t}^{C_{1}} \overset{\Delta}{=} \mathbb{P}_{1}\big(\widehat{\mu}_{1,3}(t) > \widehat{\mu}_{1,1}(t) > \widehat{\mu}_{1,2}(t)\big)$ from agent $p_{1}$ and $\mathbb{P}(\mathcal{G}_{2}(t))  \overset{\Delta}{=} \mathbb{P}_{2}\big(\widehat{\mu}_{2,2}(t) > \widehat{\mu}_{2,1}(t) > \widehat{\mu}_{2,3}(t)\big)$ from agent $p_{2}$. Here we define $\mathbb{P}_{t}^{C_{1}}$ as the probability of occurring case 1 of agent $p_{1}$ and $\mathcal{G}_{2}(t)$ represents that agent $p_{2}$ submits correct ranking list to the centralized platform and we call this as the correct ranking in the following analysis, which is equivalent to agent submitting the correct ranking list to the platform. And the bad event is equivalent to agent submitting  the incorrect rankings.
So this $\{\widehat{\mu}_{2,2}(t) > \widehat{\mu}_{2,1}(t) > \widehat{\mu}_{2,3}(t)\}$ is a \textit{good} event because agent $p_{2}$ correctly estimate its preference scheme over arms. $\{\widehat{\mu}_{1,3}(t) > \widehat{\mu}_{1,1}(t) > \widehat{\mu}_{1,2}(t)\}$ is a \textit{bad} event because agent $p_{1}$ wrongly estimate its preference scheme over arms.  The decomposed instantaneous regret for agent $p_{1}$ is
\begin{equation}
    \begin{aligned}
        &\mathbb{P}(\mathcal{G}_{2}(t)) \mathbb{P}_{t}^{C_{1}}  R_{1,t}^{C_{1}} \\
        & = \mathbb{P}\big(\widehat{\mu}_{1,3} > \widehat{\mu}_{1,1}(t) > \widehat{\mu}_{1,2}(t)\big)   \mathbb{P}\big(\widehat{\mu}_{2,2}(t) > \widehat{\mu}_{2,1}(t) > \widehat{\mu}_{2,3}(t)\big)   \theta_{1,*}\Tra(x_{1}(t) - x_{3}(t)),
    \end{aligned}
\end{equation}
where the above instantaneous regret is greater than zero.
For agent $p_{2}$, the decomposed instantaneous regret is
\begin{equation}
    \begin{aligned}
        &\mathbb{P}(\mathcal{G}_{2}(t)) \mathbb{P}_{t}^{C_{1}}  R_{2,t}^{C_{1}} \\
        & = \mathbb{P}\big(\widehat{\mu}_{1,3} > \widehat{\mu}_{1,1}(t) > \widehat{\mu}_{1,2}(t)\big)   \mathbb{P}\big(\widehat{\mu}_{2,2}(t) > \widehat{\mu}_{2,1}(t) \geq \widehat{\mu}_{2,3}\big)   \theta_{2,*}\Tra(x_{2}(t) - x_{2}(t)) = 0.
    \end{aligned}
\end{equation}

\textit{Case 2.} If agent $p_{1}$ wrongly estimates the ranking over arms as $p_{1}: a_{3} > a_{2} > a_{1}$. The matching result by DA Algorithm is in  Figure \ref{fig:6-cases} Case 2.
Agent $p_{1}$ is matched with arm $a_{3}$ and agent $p_{2}$ is matched with arm $a_{2}$, where agent $p_{1}$ suffers a positive regret. The instantaneous regret is
\begin{equation}
    \begin{aligned}
       R_{1,t}^{C_{2}} = \theta_{1,*}\Tra x_{1}(t) - \theta_{1,*}\Tra x_{3}(t) = \theta_{1,*}\Tra(x_{1}(t) - x_{3}(t)).
    \end{aligned}
\end{equation}
In addition, agent $p_{2}$ does not suffer regret in case 2.

This   bad event's joint probability is the product of two ranking probabilities $\mathbb{P}_{t}^{C_{2}} = \mathbb{P}\big(\widehat{\mu}_{1,3}(t) > \widehat{\mu}_{1,2}(t) > \widehat{\mu}_{1,1}(t)\big)$ by agent $p_{1}$ and  $\mathbb{P}(\mathcal{G}_{2}(t))$ by agent $p_{2}$.  $\{\widehat{\mu}_{1,3}(t) > \widehat{\mu}_{1,2}(t) > \widehat{\mu}_{1,1}(t)$ is the \textit{bad} event that agent $p_{1}$ wrongly estimate its preference scheme over arms. 
The decomposed instantaneous regret for agent $p_{1}$ is
\begin{equation}
    \begin{aligned}
     &\mathbb{P}(\mathcal{G}_{2}(t)) \mathbb{P}_{t}^{C_{2}}  R_{1,t}^{C_{2}} 
     \\
    &=\mathbb{P}\big(\widehat{\mu}_{1,3}(t) > \widehat{\mu}_{1,2}(t) > \widehat{\mu}_{1,1}(t)\big)   \mathbb{P}\big(\widehat{\mu}_{2,2}(t) > \widehat{\mu}_{2,1}(t) > \widehat{\mu}_{2,3}(t)\big)   \theta_{1,*}\Tra(x_{1}(t) - x_{3}(t)) > 0.
    \end{aligned}
\end{equation}
For agent $p_{2}$, the decomposed instantaneous regret is
\begin{equation}
    \begin{aligned}
        &\mathbb{P}(\mathcal{G}_{2}(t)) \mathbb{P}_{t}^{C_{2}}  R_{2,t}^{C_{2}}  \\
        &=
        \mathbb{P}\big(\widehat{\mu}_{1,3}(t) > \widehat{\mu}_{1,2}(t) > \widehat{\mu}_{1,1}(t)\big)   \mathbb{P}\big(\widehat{\mu}_{2,2}(t) > \widehat{\mu}_{2,1}(t) > \widehat{\mu}_{2,3}(t)\big)   \theta_{2,*}\Tra(x_{2}(t) - x_{2}(t)) = 0.
    \end{aligned}
\end{equation}
This case is the same as the case 1.

\textit{Case 3.} If agent $p_{1}$ wrongly estimates the ranking over arms as $p_{1}: a_{2} > a_{3} > a_{1}$. The matching result by DA Algorithm is in  Figure \ref{fig:6-cases} Case 3.
Agent $p_{1}$ is matched with arm $a_{2}$ and agent $p_{2}$ is matched with arm $a_{1}$. The decomposed instantaneous regret for agent $p_{1}$ is
\begin{equation}
    \begin{aligned}
        R_{1,t}^{C_{3}} = \theta_{1,*}\Tra x_{1}(t) - \theta_{1,*}\Tra x_{2}(t) = \theta_{1,*}\Tra(x_{1}(t) - x_{2}(t)) > 0.
    \end{aligned}
\end{equation}
In addition, agent $p_{2}$ suffers a regret.
The decomposed instantaneous regret for agent $p_{2}$ is
\begin{equation}
    \begin{aligned}
        R_{2,t}^{C_{3}} = \theta_{2,*}\Tra x_{2}(t) - \theta_{2,*}\Tra x_{1}(t) = \theta_{2,*}\Tra(x_{2}(t) - x_{1}(t)) > 0.
    \end{aligned}
\end{equation}

This   bad event's joint probability is the product of two ranking probabilities $\mathbb{P}_{t}^{C_{3}} = \mathbb{P}\big(\widehat{\mu}_{1,2}(t) > \widehat{\mu}_{1,3}(t) > \widehat{\mu}_{1,1}(t)\big)$ by agent $p_{1}$ and  $\mathbb{P}(\mathcal{G}_{2}(t))$ by agent $p_{2}$. $\{\widehat{\mu}_{1,2}(t) > \widehat{\mu}_{1,3}(t) > \widehat{\mu}_{1,1}(t)\}$ is the \textit{bad} event that agent $p_{1}$ wrongly estimate its preference scheme over arms. The decomposed instantaneous regret for agent $p_{1}$ is
\begin{equation}
    \begin{aligned}
     &\mathbb{P}(\mathcal{G}_{2}(t)) \mathbb{P}_{t}^{C_{3}}  R_{1,t}^{C_{3}} \\
     &=\mathbb{P}\big(\widehat{\mu}_{1,2}(t) > \widehat{\mu}_{1,3}(t) > \widehat{\mu}_{1,1}(t)\big)   \mathbb{P}\big(\widehat{\mu}_{2,2}(t) > \widehat{\mu}_{2,1}(t) > \widehat{\mu}_{2,3}(t)\big)   \theta_{1,*}\Tra(x_{1}(t) - x_{2}(t)) > 0.
    \end{aligned}
\end{equation}
For agent $p_{2}$, the decomposed instantaneous regret is
\begin{equation}
    \begin{aligned}
        &\mathbb{P}(\mathcal{G}_{2}(t)) \mathbb{P}_{t}^{C_{3}}  R_{2,t}^{C_{3}} \\
     &=
        \mathbb{P}\big(\widehat{\mu}_{1,3}(t) > \widehat{\mu}_{1,1} > \widehat{\mu}_{1,2}(t)\big)   \mathbb{P}\big(\widehat{\mu}_{2,2}(t) > \widehat{\mu}_{2,1}(t) > \widehat{\mu}_{2,3}(t)\big)   \theta_{2,*}\Tra(x_{2}(t) - x_{1}(t)) > 0.
    \end{aligned}
\end{equation}

\textit{Case 4.} If agent $p_{1}$ wrongly estimates the ranking over arms as $p_{1}: a_{2} > a_{1} > a_{3}$. The matching result by DA Algorithm is in  Figure \ref{fig:6-cases} Case 4.
Agent $p_{1}$ is matched with arm $a_{2}$ and agent $p_{2}$ is matched with arm $a_{1}$. The decomposed instantaneous regret for agent $p_{1}$ is
\begin{equation}
    \begin{aligned}
        R_{1,t}^{C_{4}} = \theta_{1,*}\Tra x_{1}(t) - \theta_{1,*}\Tra x_{2}(t) = \theta_{1,*}\Tra(x_{1}(t) - x_{2}(t)) > 0.
    \end{aligned}
\end{equation}
In addition, agent $p_{2}$ suffers a positive regret. The decomposed instantaneous regret for agent $p_{2}$ is
\begin{equation}
    \begin{aligned}
        R_{2,t}^{C_{4}} = \theta_{2,*}\Tra x_{2}(t) - \theta_{2,*}\Tra x_{1}(t) = \theta_{2,*}\Tra(x_{2}(t) - x_{1}(t)) > 0.
    \end{aligned}
\end{equation}

This  bad event's joint probability is the product of two ranking probabilities $\mathbb{P}_{t}^{C_{4}} = \mathbb{P}\big(\widehat{\mu}_{1,2}(t)  > \widehat{\mu}_{1,1}(t) > \widehat{\mu}_{1,3}(t)\big)$ by agent $p_{1}$ and $\mathbb{P}(\mathcal{G}_{2}(t))$ by agent $p_{2}$.
$\{\widehat{\mu}_{1,2}(t) > \widehat{\mu}_{1,1}(t) > \widehat{\mu}_{1,3}(t)\}$ is the \textit{bad} event that agent $p_{1}$ wrongly estimate its preference scheme over arms. The decomposed instantaneous regret for agent $p_{1}$ is
\begin{equation}
    \begin{aligned}
     &\mathbb{P}(\mathcal{G}_{2}(t)) \mathbb{P}_{t}^{C_{4}}  R_{1,t}^{C_{4}} \\
     &=
        \mathbb{P}\big(\widehat{\mu}_{1,2}(t)  > \widehat{\mu}_{1,1}(t) > \widehat{\mu}_{1,3}(t)\big)   \mathbb{P}\big(\widehat{\mu}_{2,2}(t) > \widehat{\mu}_{2,1}(t) > \widehat{\mu}_{2,3}(t)\big)   \theta_{1,*}\Tra(x_{1}(t) - x_{2}(t)) > 0.
    \end{aligned}
\end{equation}
For agent $p_{2}$, the decomposed instantaneous regret is
\begin{equation}
    \begin{aligned}
            &\mathbb{P}(\mathcal{G}_{2}(t)) \mathbb{P}_{t}^{C_{4}}  R_{2,t}^{C_{4}} \\
     &=
        \mathbb{P}\big(\widehat{\mu}_{1,2}(t)  > \widehat{\mu}_{1,1}(t) > \widehat{\mu}_{1,3}(t)\big)   \mathbb{P}\big(\widehat{\mu}_{2,2}(t) > \widehat{\mu}_{2,1}(t) > \widehat{\mu}_{2,3}(t)\big)   \theta_{2,*}\Tra(x_{2}(t) - x_{1}(t)) > 0.
    \end{aligned}
\end{equation}

\textit{Case 5.} If agent $p_{1}$ wrongly estimates the ranking over arms as  $p_{1}: a_{1} > a_{3} > a_{2}$. The matching result by DA Algorithm is in  Figure \ref{fig:6-cases} Case 5.
Agent $p_{1}$ is matched with arm $a_{1}$ and agent $p_{2}$ is matched with arm $a_{2}$. This pair will not suffer regret. The decomposed instantaneous regret for agent $p_{1}$ is
\begin{equation}
    \begin{aligned}
        R_{1,t}^{C_{5}} = \theta_{1,*}\Tra x_{1}(t) - \theta_{1,*}\Tra x_{1}(t) = \theta_{1,*}\Tra(x_{1}(t) - x_{1}(t)) = 0.
    \end{aligned}
\end{equation}
In addition, agent $p_{2}$ will not suffer a regret. The decomposed instantaneous regret for agent $p_{2}$ is
\begin{equation}
    \begin{aligned}
        R_{2,t}^{C_{5}} = \theta_{2,*}\Tra x_{2}(t) - \theta_{2,*}\Tra x_{2}(t) = \theta_{2,*}\Tra(x_{2}(t) - x_{2}(t)) = 0.
    \end{aligned}
\end{equation}

This bad event's joint probability is the product of two ranking probabilities $\mathbb{P}_{t}^{C_{5}} = \mathbb{P}\big(\widehat{\mu}_{1,1}(t) > \widehat{\mu}_{1,3}(t) > \widehat{\mu}_{1,2}(t)\big) $ by agent $p_{1}$ and  $\mathbb{P}(\mathcal{G}_{2}(t))$ by agent $p_{2}$. 
$\{\widehat{\mu}_{1,1}(t) > \widehat{\mu}_{1,3}(t) > \widehat{\mu}_{1,2}(t)\}$ is the \textit{bad} event that agent $p_{1}$ wrongly estimate its preference scheme over arms. The decomposed instantaneous regret for agent $p_{1}$ is
\begin{equation}
    \begin{aligned}
         &\mathbb{P}(\mathcal{G}_{2}(t)) \mathbb{P}_{t}^{C_{5}}  R_{1,t}^{C_{5}} \\
     &=
        \mathbb{P}\big(\widehat{\mu}_{1,1}(t) > \widehat{\mu}_{1,3}(t) > \widehat{\mu}_{1,2}(t)\big)   \mathbb{P}\big(\widehat{\mu}_{2,2}(t) > \widehat{\mu}_{2,1}(t) > \widehat{\mu}_{2,3}(t)\big)   \theta_{1,*}\Tra(x_{1}(t) - x_{1}(t)) = 0.
    \end{aligned}
\end{equation}
For agent $p_{2}$, the decomposed instantaneous regret is
\begin{equation}
    \begin{aligned}
            &\mathbb{P}(\mathcal{G}_{2}(t)) \mathbb{P}_{t}^{C_{5}}  R_{2,t}^{C_{5}} \\
     &=
        \mathbb{P}\big(\widehat{\mu}_{1,1}(t) > \widehat{\mu}_{1,3}(t) > \widehat{\mu}_{1,2}(t)\big)   \mathbb{P}\big(\widehat{\mu}_{2,2}(t) > \widehat{\mu}_{2,1}(t) > \widehat{\mu}_{2,3}(t)\big)   \theta_{2,*}\Tra(x_{2}(t) - x_{2}(t)) = 0.
    \end{aligned}
\end{equation}
This setting will not create any regret.

\textit{Case 6.} If agent $p_{1}$ correctly estimates the ranking over arms as  $p_{1}: a_{1} > a_{2} > a_{3}$. The matching result by DA Algorithm is in  Figure \ref{fig:6-cases} Case 6.
Agent $p_{1}$ is matched with arm $a_{1}$ and agent $p_{2}$ is matched with arm $a_{2}$. This pair will not suffer regret. The decomposed instantaneous regret for agent $p_{1}$ is
\begin{equation}
    \begin{aligned}
        R_{1,t}^{C_{6}} = \theta_{1,*}\Tra x_{1}(t) - \theta_{1,*}\Tra x_{1}(t) = \theta_{1,*}\Tra(x_{1}(t) - x_{1}(t)) = 0.
    \end{aligned}
\end{equation}
In addition, agent $p_{2}$ will not suffer a regret. The decomposed instantaneous regret for agent $p_{1}$ is
\begin{equation}
    \begin{aligned}
        R_{2,t}^{C_{6}} =\theta_{2,*}\Tra x_{2}(t) - \theta_{2,*}\Tra x_{2}(t) = \theta_{2,*}\Tra(x_{2}(t) - x_{2}(t)) = 0.
    \end{aligned}
\end{equation}

This   bad event's joint probability is the product of two ranking probabilities $\mathbb{P}_{t}^{C_{6}} = \mathbb{P}\big(\widehat{\mu}_{1,1}(t) > \widehat{\mu}_{1,2}(t) > \widehat{\mu}_{1,3}(t)\big) $ by agent $p_{1}$, which in fact is a good event and  $\mathbb{P}(\mathcal{G}_{2}(t))$ by agent $p_{2}$. $\{\widehat{\mu}_{1,1}(t) > \widehat{\mu}_{1,2}(t) > \widehat{\mu}_{1,3}(t)\}$ is the \textit{good} event that agent $p_{1}$ correctly estimate its preference scheme over arms. The decomposed instantaneous regret for agent $p_{1}$ is
\begin{equation}
    \begin{aligned}
         &\mathbb{P}(\mathcal{G}_{2}(t)) \mathbb{P}_{t}^{C_{6}}  R_{1,t}^{C_{6}} \\
        &=
        \mathbb{P}\big(\widehat{\mu}_{1,1}(t) > \widehat{\mu}_{1,2}(t) > \widehat{\mu}_{1,3}(t)\big)   \mathbb{P}\big(\widehat{\mu}_{2,2}(t) > \widehat{\mu}_{2,1}(t) > \widehat{\mu}_{2,3}(t)\big)   \theta_{1,*}\Tra(x_{1}(t) - x_{1}(t)) = 0,
    \end{aligned}
\end{equation}
For agent $p_{2}$, the decomposed instantaneous regret is
\begin{equation}
    \begin{aligned}
        &\mathbb{P}(\mathcal{G}_{2}(t)) \mathbb{P}_{t}^{C_{6}}  R_{2,t}^{C_{6}} \\
        &=
        \mathbb{P}\big(\widehat{\mu}_{1,1}(t) > \widehat{\mu}_{1,3}(t) > \widehat{\mu}_{1,2}(t)\big)   \mathbb{P}\big(\widehat{\mu}_{2,2}(t) > \widehat{\mu}_{2,1}(t) > \widehat{\mu}_{2,3}(t)\big)   \theta_{2,*}\Tra(x_{2}(t) - x_{2}(t)) = 0.
    \end{aligned}
\end{equation}
This setting will also not create any regret.

In summary, for agent $p_{1}$, the four regret occurred cases are represented in Case 1 to Case 4, two regret vanished cases happen at Case 5 and Case 6. For agent $p_{2}$, the two regret occurred cases are represented in Case 3 and Case 4, four regret vanishing cases happen at Case 1, Case 2, Case 5, and Case 6. 
These six cases represent all the possible regret occurring cases when $p_{1}$ submits incorrect ranking and $p_{2}$ submits correct ranking.

\section{Proof of Theorem \ref{thm: lower bound} - Instance-Dependent Lower Bound}
\label{supp: lower bound proof}

Based on the setting constructed in Section \ref{sec: theory-lowerbound}, we conduct the regret analysis to get the lower bound.
After $h$ rounds of exploration, for agent $p_{i}$, its estimator $\widehat{\theta}_{i}(t)$ is acquired through the penalized linear regret. Thus at time step $t$, the estimated mean reward for arm $a_{j}$ from the viewpoint of agent $p_{i}$  is $\widehat{\mu}_{i,j}(t) = \widehat{\theta}_{i}(t)\Tra x_{j}(t)$, which provides the basis to construct the ranking list $\widehat{r}_{i,[K]}(t)$. Besides, since all contexts are from uniform distribution, conditioning on all previous information $\mathcal{F}_{i}(h)$ and contextual information of $x_{j}(t)$, we have the distribution of the estimated mean reward $\widehat{\mu}_{i,j}(t)$ following the normal distribution 
\begin{equation}
    \begin{aligned}
        \widehat{\mu}_{i,j}(t) = \widehat{\theta}_{i}(t)\Tra x_{j}(t) | \mathcal{F}_{i}(h) \sim N(\bar{\theta}_{i}\Tra x_{j}(t), \sigma^{2} x_{j}(t)\Tra\M{M}_{i} x_{j}(t)), \quad \forall j \in [K], 
    \end{aligned}
\end{equation}
where $\mathbb{E}[\widehat{\theta}_{i}(t)|\mathcal{F}_{i}(h)] = \bar{\theta}_{i}= 
\big(\M{X}_{i} (h)\Tra \M{X}_{i} (h) + \lambda_{i} \M{I} \big)^{-1} 
\M{X}_{i}(h)\Tra \M{X}_{i}(h)
\theta_{i,*} \in \R^{d}$, 
and $\text{Cov}[\widehat{\theta}_{i}(t)|\mathcal{F}_{i}(h)] = \sigma^{2}\M{M}_{i}= \sigma^{2}\big(\M{X}_{i} (h)\Tra \M{X}_{i} (h) + \lambda_{i} \M{I} \big)^{-1}
\M{X}_{i}(h)\Tra \M{X}_{i}(h) 
\big(\M{X}_{i} (h)\Tra \M{X}_{i} (h) + \lambda_{i} \M{I} \big)^{-1}
\in \R^{d\times d}$.

Denote the true preference for $p_{i}$ at $t$ is $a_{j_{1}} <_{i}^{t} a_{j_{2}} <_{i}^{t} a_{j_{3}}$ and the correct ranking event and partial correct ranking rank event as 
$\mathcal{G}_{i}(t) = \{\widehat{\mu}_{i,j_{1}}(t) > \widehat{\mu}_{i,j_{2}}(t) > \widehat{\mu}_{i,j_{3}}(t)\}$ and $\mathcal{G}_{i}^{c}(t) = \{\widehat{\mu}_{i,j_{1}}(t) > \widehat{\mu}_{i,j_{2}}(t) > \widehat{\mu}_{i,j_{3}}(t)\}^{c}$.
The lower bound probability of the correct ranking estimate (good event) and partial correct ranking estimate (bad event) is provided as follows.
\begin{lem}
\label{lem: good event: lower bound}
(1) Define  
$\M{M}_{i} = \big(\M{X}_{i} (h)\Tra \M{X}_{i} (h) + \lambda_{i} \M{I} \big)^{-1}
\M{X}_{i}(h)\Tra \M{X}_{i}(h) 
\big(\M{X}_{i} (h)\Tra \M{X}_{i} (h) + \lambda_{i} \M{I} \big)^{-1}
\in \R^{d\times d}$, 
and $\Sigma_{i, (j, k)}(t) = \sigma^{2} [x_{j}(t)\Tra\M{M}_{i}x_{j}(t) + x_{k}(t)\Tra\M{M}_{i}x_{k}(t)]$. 
If the true preference for $p_{i}$ over arms is $a_{j_{1}} < a_{j_{2}} < a_{j_{3}}$ at time step $t$, the probability of $\mathcal{G}_{i}(t)$ is lower bounded by
\begin{equation}
    \begin{aligned}
        \mathbb{P}(\mathcal{G}_{i}(t))
        &\geq 
        1 - \frac{1}{\sqrt{2\pi}}\big[\Psi_{i,t}(j_{1}, j_{2}) + \Psi_{i,t}(j_{2}, j_{3}) + \Psi_{i,t}(j_{1}, j_{3})\big],
    \end{aligned}
\end{equation}
where $\Psi_{i,t}(j, k) = \exp{(-\nu^{2}_{i,(j,k)}(t)/2)}/\nu_{i,t}(j,k)$ and $\nu_{i,t}(j,k) = \bar{\theta}_{i}\Tra [x_{j}(t) - x_{k}(t)]/\Sigma_{i, (j, k)}(t)$  represents the scaled mean difference of $a_{j}$ and  $a_{k}$ from the perspective of $\bar{\theta}_{i}$ at time $t$. 

(2) Define $\tilde{\nu}_{i,t}(j,k) = \bar{\theta}_{i}\Tra [x_{j}(t) - x_{k}(t)]/\tilde{\Sigma}_{i, t}(j,k)$
and
$\tilde{\Sigma}_{i, t}(j,k) = \sigma^2[x_{j}(t)\Tra\M{M}_{i} x_{j}(t) + x_{k}(t)\Tra\M{M}_{i} x_{k}(t) - 2x_{j}(t)\Tra\M{M}_{i}x_{k}(t)]$. If the true preference for $p_{i}$ over arms is $a_{j_{1}} < a_{j_{2}} < a_{j_{3}}$ at time step $t$, the $\mathcal{G}_{i}^{c}(t)$ probability lower bound is,
\begin{equation}
    \begin{aligned}
        \mathbb{P}\big(\mathcal{G}_{i}^{c}(t)\big)
        \geq \min\big\{ \Gamma_{i, t}(j_{1},j_{2}), \Gamma_{i, t}(j_{2},j_{3})\big\}
    \end{aligned}
\end{equation}
where $\Gamma_{i, t}(j,k) = (1/\tilde{\nu}_{i,t}(j, k) - 1/\tilde{\nu}^{3}_{i,t}(j,k))\exp{(-\tilde{\nu}^{2}_{i,t}(j,k)/2)}$.
\end{lem}

Lemma \ref{lem: good event: lower bound} is used to getting the $\mathcal{G}_{i}(t)$ and $\mathcal{G}_{i}^{c}(t)$'s lower bounds via the sharp Gaussian tail probability lower bound at each time step. In addition, the conditional expectation regret is provided in Section \ref{appendix: Example-6-cases}.  The following lemma provides the order of lower bounds of $\mathcal{G}_{i}(t)$ and $\mathcal{G}_{i}^{c}(t)$.

\begin{lem}
\label{corr: uniform good lower bound and failure lower bound}
Considering the problem instance in appendix, the order of the probability's lower bound are 
\begin{equation}
    \begin{aligned}
        \mathbb{P}(\mathcal{G}_{i}(t)) 
        \geq
        \mathcal{L}_{i}^{g}(t) \text{ and } 
        \mathbb{P}\big(\mathcal{G}_{i}^{c}(t)\big)
       \geq
       \mathcal{L}_{i}^{b}(t),
    \end{aligned}
\end{equation}
where $\mathcal{L}_{i}^{g}(t)= 1 -(3/c_{5}(t)\sqrt{2\pi h})\exp{(-
c_{5}^{2}(t)h/2)}$,
$\mathcal{L}_{i}^{b}(t) =(1/c_{7}(t)\sqrt{h} - 1/c_{7}^{3}(t)h^{3/2})
\exp{(-c_{7}^{2}(t)h/2)}$, and $c_{5}(t), c_{7}(t)$ are contextual time-dependent constants but independent of designing exploration rounds $h$.
\end{lem}

With the distribution of $\widehat{\mu}_{i,j}(t)$, to derive the regret lower bound, we provide the proof of good events $\mathcal{G}_{1}(t)$ and $\mathcal{G}_{2}(t)$'s probability lower bound in Section \ref{Appendix: Proof of good event: lower bound}, and bad events $\mathcal{G}_{1}^{c}(t)$ and $\mathcal{G}_{2}^{c}(t)$'s lower bound in Section \ref{Appedix: proof of k-1 step good evnet probability}. 
In addition, we provide these events' probability lower bounds'  order at time $t$, which is provided in Section
\ref{Appendix: proof of lower bound order}. 
Finally, with the previous technical lemmas, 
we provide the final instance-dependent regret lower bound as a whole.

To get the regret of agent $p_{1}$, we first assume that  $p_{2}$ correctly estimates its preference at time step $t$ in the exploitation step. So the instantaneous regret $R_{1,t}(\widehat{r}_{2}(t) = r_{2}(t))$ for agent $p_{1}$, if agent $p_{2}$ submits correct ranking, can be decomposed as follows,
\begin{equation}
    \begin{aligned}
        R_{1,t}(\widehat{r}_{2}(t) = r_{2}(t))
        & = \mathbb{P}\big(\mathcal{G}_{2}(t)\big) \mathbb{E}[R_{1}]
        \\
        &=\mathbb{P}\big(\mathcal{G}_{2}(t)\big) \sum_{z=1}^{6} \mathbb{P}^{C_{z}}_{t} R_{1,t}^{C_{z}}(\widehat{r}_{2}(t) = r_{2}(t)).
    \end{aligned}
\end{equation}
where these six cases' regret analysis can be found at Appendix \ref{appendix: Example-6-cases}. So we can decompose these six cases' regrets into 
\begin{equation}
    \begin{aligned}
        \sum_{z=1}^{6} \mathbb{P}^{C_{z}}_{t} R_{1,t}^{C_{z}}(\widehat{r}_{2}(t) = r_{2}(t)) 
        & = \theta_{1,*}\Tra
        \Bigg[
        \mathbb{P}\big(\widehat{\mu}_{1,3}(t) > \widehat{\mu}_{1,1}(t) > \widehat{\mu}_{1,2}(t)\big) (x_{1}(t) - x_{3}(t))
        \\
        &\quad\quad\quad +  \mathbb{P}\big(\widehat{\mu}_{1,3}(t) > \widehat{\mu}_{1,2}(t) > \widehat{\mu}_{1,1}(t)\big) (x_{1}(t) - x_{3}(t))\\
        &\quad\quad\quad
        + \mathbb{P}\big(\widehat{\mu}_{1,2}(t) > \widehat{\mu}_{1,3}(t) > \widehat{\mu}_{1,1}(t)\big) (x_{1}(t) - x_{2}(t))\\
        &\quad\quad\quad +  \mathbb{P}\big(\widehat{\mu}_{1,2}(t) > \widehat{\mu}_{1,1}(t) > \widehat{\mu}_{1,3}(t)\big) (x_{1}(t) - x_{2}(t))
        \Bigg],
    \end{aligned}
\end{equation}
because there are four cases suffering regret and two cases without suffering regret. Combining case 1 and case 2 as a whole, and case 3 and case 4 together, we obtain
\begin{equation}
    \begin{aligned}
        &= \theta_{1,*}\Tra 
        \Bigg[
        \bigg(\mathbb{P}\big(\widehat{\mu}_{1,3}(t) > \widehat{\mu}_{1,1}(t) > \widehat{\mu}_{1,2}(t)\big) + \mathbb{P}\big(\widehat{\mu}_{1,2}(t) > \widehat{\mu}_{1,3}(t) > \widehat{\mu}_{1,1}(t)\big)\bigg)(x_{1}(t) - x_{3}(t)) \\
        &\quad\quad
        +\bigg(\mathbb{P}\big(\widehat{\mu}_{1,2}(t) > \widehat{\mu}_{1,3}(t) > \widehat{\mu}_{1,1}(t)\big) + \mathbb{P}\big(\widehat{\mu}_{1,2}(t) > \widehat{\mu}_{1,1}(t) > \widehat{\mu}_{1,3}(t)\big)\bigg)
        (x_{1}(t) - x_{2}(t))
        \Bigg].
    \end{aligned}
\end{equation}
With Lemma \ref{lem: good event: lower bound},
we have the bad event's probability lower bound, and define $\overline{\Delta}_{1,\min}(t) = \underset{j \in [3], \overline{\Delta}_{1,j}(t) > 0}{\min} \overline{\Delta}_{1,j}(t) = \underset{j \in [K], \overline{\Delta}_{1,j}(t) > 0}{\min} \langle\theta_{1,*},x_{\overline{m}_{t}(1)}(t) - x_{j}(t) \rangle$, we can get this instantaneous regret as follows
\begin{equation}
    \begin{aligned}
        &\geq
        \Bigg[
        \bigg(\mathbb{P}\big(\widehat{\mu}_{1,3}(t) > \widehat{\mu}_{1,1}(t) > \widehat{\mu}_{1,2}(t)\big) + \mathbb{P}\big(\widehat{\mu}_{1,2}(t) > \widehat{\mu}_{1,3}(t) > \widehat{\mu}_{1,1}(t)\big)\bigg)
        \overline{\Delta}_{1, \min}(t) \\
        &\quad\quad
        +\bigg(\mathbb{P}\big(\widehat{\mu}_{1,2}(t) > \widehat{\mu}_{1,3}(t) > \widehat{\mu}_{1,1}(t)\big) + \mathbb{P}\big(\widehat{\mu}_{1,2}(t) > \widehat{\mu}_{1,1}(t) > \widehat{\mu}_{1,3}(t)\big)\bigg)
        \overline{\Delta}_{1, \min}(t)
        \Bigg]\\
        &=
        \mathbb{P}\big(\mathcal{G}_{1}^{c}(t)\big)\overline{\Delta}_{1, \min}(t)
    \end{aligned}
\end{equation}
So the regret for agent $p_{1}$ is lower bounded by 
\begin{equation}
    \begin{aligned}
        R_{1,t}(\widehat{r}_{2}(t) = r_{2}(t)) &\geq \mathbb{P}\big(\mathcal{G}_{2}(t)\big) \mathbb{P}\big(\mathcal{G}_{1}^{c}(t)\big)\overline{\Delta}_{1, \min}(t) 
        \geq 
        \mathbb{P}\big(\mathcal{G}_{2}(t)\big) 
        \mathcal{L}_{1}^{b}(t) 
        \overline{\Delta}_{1, \min}(t).
    \end{aligned}
\end{equation}
Based on Lemma \ref{lem: good event: lower bound}, we have the good event $\mathcal{G}_{2}(t)$'s probability lower bound and get
\begin{equation}
    \begin{aligned}
        R_{1,t}(\widehat{r}_{2}(t) = r_{2}(t))\geq  \mathcal{L}_{2}^{g}(t) \mathcal{L}_{1}^{b}(t)\overline{\Delta}_{1, \min}(t).
    \end{aligned}
\end{equation}
With the same rule, we obtain similar result when $p_{2}$ is incorrect,
\begin{equation}
    \begin{aligned}
        R_{1,t}(\widehat{r}_{2}(t) \neq r_{2}(t)) \geq \overline{\Delta}_{1, \min}(t) \prod_{i=1}^{2}\mathcal{L}_{i}^{b}(t).
    \end{aligned}
\end{equation}
By considering agent $p_{2}$'s preference at time $t$, the regret for agent $p_{1}$ at time $t$ is lower bounded by 
\begin{equation}
    \begin{aligned}
        R_{1}(t) \geq \overline{\Delta}_{1, \min}(t) \bigg(\prod_{i=1}^{2}\mathcal{L}_{i}^{b}(t)  + \mathcal{L}_{2}^{g}(t) \mathcal{L}_{1}^{b}(t) \bigg).
    \end{aligned}
\end{equation}
The agent $p_{2}$ gets similar regret lower bound by symmetry. So the overall lower bound regret for agent $p_{1}$ is
\begin{equation}
    \begin{aligned}
        R_{1}(T) \geq 
        \sum_{t=1}^{h} \Delta_{i,m_{t}(i)}(t)
        +
        \sum_{t = h+1}^{T}
        \overline{\Delta}_{1, \min}(t) \bigg(\prod_{i=1}^{2}\mathcal{L}_{i}^{b}(t)  + \mathcal{L}_{2}^{g}(t) \mathcal{L}_{1}^{b}(t) \bigg).
    \end{aligned}
\end{equation}
Besides, we analyze the order of the two probability lower bounds' product.
\begin{equation}
    \begin{aligned}
        &\mathcal{L}_{2}^{g}(t) \mathcal{L}_{1}^{b}(t) 
        =
        (\frac{1}{\sqrt{h}} - \frac{1}{\sqrt{h^3}})
        e^{-\frac{h}{2}}(1- 
        \frac{1}{\sqrt{h}}e^{-\frac{h}{2}})
        =
        \frac{c_{8}(t)}{\sqrt{h}}
        e^{-\frac{h}{2}},
    \end{aligned}
\end{equation}
where $c_{8}(t)$ is a context-dependent constant, but independent of $h$.
And the product order of these bad events' probability lower bounds between two agents is,
\begin{equation}
    \begin{aligned}
        \mathcal{L}_{1}^{b}(t) \mathcal{L}_{2}^{b}(t)
        =
        (\frac{1}{\sqrt{h}} - \frac{1}{\sqrt{h^3}})
        e^{-\frac{h}{2}}
        (\frac{1}{\sqrt{h}} - \frac{1}{\sqrt{h^3}})
        e^{-\frac{h}{2}}
        = \frac{1}{h}
        e^{-c_{9}(t)h},
    \end{aligned}
\end{equation}
where $c_{9}(t)$ is a context-dependent constant, but independent of $h$. So the sum of $\mathcal{L}_{2}^{g}(t) \mathcal{L}_{1}^{b}(t)$ and 
$\mathcal{L}_{1}^{b}(t) \mathcal{L}_{2}^{b}(t)$ will be $\frac{c_{10}(t)}{\sqrt{h}}e^{-h}$. By $1/\sqrt{h} > 1/\sqrt{T}$,
the $R_{1}(T)$ will be lower bounded by $h\overline{\Delta}_{i, \min} + \sqrt{T}\overline{\Delta}_{1, \min}c_{10}(t)e^{-h}$. Then by similar analysis derived in the upper bound order analysis of dynamic matching in Appendix \ref{Appendix: order analysis of upper bound of CC-ETC}, we find that the order of the regret lower bound will be $\Omega(\log(T))$.

\subsection{Proof of Lemma \ref{lem: good event: lower bound} - Good Event} 
\label{Appendix: Proof of good event: lower bound}
\textit{Proof.} 
First, without loss of generality, suppose that the true preference from agent $p_{i}$ to all arms is $a_{j_{1}} <_{i}^{t} a_{j_{2}} <_{i}^{t} a_{j_{3}}$ at time $t$.
In order to present the competing status of those agents, we need to quantify the probability of the good event $\mathcal{G}_{i}(t) = \{\widehat{\mu}_{i,j_{1}}(t) > \widehat{\mu}_{i,j_{2}}(t) > \widehat{\mu}_{i,j_{3}}(t)\}$. Here we denote $\mathcal{A}(t) = \{\widehat{\mu}_{i,j_{1}}(t) > \max(\widehat{\mu}_{i,j_{2}}(t), \widehat{\mu}_{i,j_{3}}(t))\}$ as the \textit{1st-good event} and $\mathcal{B}(t) = \{\widehat{\mu}_{i,j_{2}}(t) > \widehat{\mu}_{i,j_{3}}(t)\}$ as the \textit{2nd-good event}, where $\mathcal{G}_{i}(t) = \mathcal{A}(t) \cap \mathcal{B}(t)$. Here we omit the index 'i' in \textit{1st-good event} and \textit{2nd-good event}. The \textit{1st-good event} $\mathcal{A}$ can also be divided into to the event $\mathcal{A}_{1}(t) = \{\widehat{\mu}_{i,j_{1}}(t) > \widehat{\mu}_{i,j_{2}}(t)\}$ and the event $\mathcal{A}_{2}(t) = \{\widehat{\mu}_{i,j_{1}}(t) > \widehat{\mu}_{i,j_{3}}(t)\}$ happening simultaneously, where $\mathcal{A}(t) = \mathcal{A}_{1}(t) \cap \mathcal{A}_{2}(t)$.
By the property of $\mathbb{P}(\mathcal{A}(t) \cap \mathcal{B}(t)) \geq \mathbb{P}(\mathcal{A}(t)) + \mathbb{P}(\mathcal{B}(t)) - \mathbb{P}(\mathcal{A}(t) \cup \mathcal{B}(t)) \geq \mathbb{P}(\mathcal{A}(t)) + \mathbb{P}(\mathcal{B}(t)) - 1$,  we use the same technique again and have $\mathbb{P}(\mathcal{A}(t)) \geq \mathbb{P}(\mathcal{A}_{1}(t)) + \mathbb{P}(\mathcal{A}_{2}(t)) - 1$.
So the event $\mathcal{A}(t)$'s probability lower bound is,
\begin{equation}
\label{eq:first-step succ event prob}
    \begin{aligned}
        \mathbb{P}(\mathcal{A}(t))
        &=\mathbb{P}(\widehat{\mu}_{i,j_{1}}(t)> \max(\widehat{\mu}_{i,j_{2}}(t), \widehat{\mu}_{i,j_{3}}(t))) \\
        &= \mathbb{P}(\{\widehat{\mu}_{i,j_{1}}(t)> \widehat{\mu}_{i,j_{2}}(t)\} \cap \{\widehat{\mu}_{i,j_{1}}(t)> \widehat{\mu}_{i,j_{3}}(t)\}) \\
        &\geq \mathbb{P}(\widehat{\mu}_{i,j_{1}}(t)> \widehat{\mu}_{i,j_{2}}(t)) + \mathbb{P}(\widehat{\mu}_{i,j_{1}}(t)> \widehat{\mu}_{i,j_{3}}(t)) - 1 \\
        &=\mathbb{P}(\mathcal{A}_{1}(t)) + \mathbb{P}(\mathcal{A}_{2}(t)) - 1.
    \end{aligned}
\end{equation}
Now we have to quantify the event $\mathcal{A}_{1}(t)$ and event $\mathcal{A}_{2}(t)$'s probabilities' lower bound. We first define the estimated mean reward difference for agent $p_{i}$ at time $t$ between arm $a_{j_{1}}$ and arm $a_{j_{2}}$ as $\widehat{Z}_{i,(j_{1},j_{2})} = \widehat{\mu}_{i,j_{1}}(t)- \widehat{\mu}_{i,j_{2}}(t)$. Given all contextual information at time $t$,  we get 
\begin{equation}
    \begin{aligned}
         \widehat{Z}_{i,(j_{1},j_{2})} |\mathcal{F}_{i}(h)
         \sim N(\bar{\theta}_{i}\Tra [x_{j_{1}}(t) - x_{j_{2}}(t)], \tilde{\Sigma}_{i, (j_{1}, j_{2})}(t)),
    \end{aligned}
\end{equation}
where $\tilde{\Sigma}_{i, (j_{1}, j_{2})}(t) = \sigma^{2} [x_{j_{1}}(t)\Tra\M{M}_{i} x_{j_{1}}(t) + x_{j_{2}}(t)\Tra\M{M}_{i} x_{j_{2}}(t) -  2x_{j_{1}}(t)\Tra\M{M}_{i} x_{j_{2}}(t)]\in \R$ is the variance of the estimated mean reward $\widehat{\mu}_{i,j_{1}}(t)- \widehat{\mu}_{i,j_{2}}(t)$. 
We know that $\widehat{\mu}_{i,j_{1}}(t)$ and $\widehat{\mu}_{i,j_{2}}(t)$ are positively correlated because $\M{M}_{i}$ is positive semi-definite since $x_{j_{1}}$ and $x_{j_{2}}$'s coordinates are follow uniform distribution $U(0,1)^{d}$. 
So the variance of the difference  $\widehat{\mu}_{i,j_{1}}(t)- \widehat{\mu}_{i,j_{2}}(t)$ of the two correlated normal random variables is less than the variance of the difference of two independent normal random variables by the property $var(\varpi_{1}-\varpi_{2}) \leq var(\varpi_{1}) + var(\varpi_{2})$ if $\varpi_{1}$ and $\varpi_{2}$ are positively correlated random variables. 
Besides, we know if two normal random variables $\widehat{\mu}_{i,j_{1}}(t)$ and $\widehat{\mu}_{i,j_{2}}(t)$ are independent,
\begin{equation}\label{Eq: normal variance comparison}
    \begin{aligned}
        \tilde{\Sigma}_{i, (j_{1}, j_{2})}(t) \leq
        \sigma^{2} [x_{j_{1}}(t)\Tra\M{M}_{i} x_{j_{1}}(t) + x_{j_{2}}(t)\Tra\M{M}_{i} x_{j_{2}}(t)],
    \end{aligned}
\end{equation}
where $x_{j_{1}}(t)\Tra\M{M}_{i} x_{j_{1}}(t)$ and $x_{j_{2}}(t)\Tra\M{M}_{i} x_{j_{2}}(t)$ are the variances of $\widehat{\mu}_{i,j_{1}}(t)$ and $\widehat{\mu}_{i,j_{2}}(t)$ correspondingly, and we define $\Sigma_{i, (j_{1}, j_{2})}(t) = \sigma^{2} [x_{j_{1}}(t)\Tra\M{M}_{i} x_{j_{1}}(t) + x_{j_{2}}(t)\Tra\M{M}_{i} x_{j_{2}}(t)]$. We use the \textit{proxy} random variable $Z_{i,(j_{1},j_{2})}$ to define the difference of two independent Gaussian random variables. $Z_{i,(j_{1},j_{2})}$'s distribution follows the normal distribution
\begin{equation}
    \begin{aligned}
        Z_{i,(j_{1},j_{2})}|\mathcal{F}_{i}(h) \sim N(\bar{\theta}_{i}\Tra [x_{j_{1}}(t) - x_{j_{2}}(t)], \Sigma_{i, (j_{1}, j_{2})}(t)),
    \end{aligned}
\end{equation}
where $\bar{\theta}_{i}\Tra [x_{j_{1}}(t) - x_{j_{2}}(t)]$ is the $Z_{i,(j_{1},j_{2})}$'s expectation and $\Sigma_{i, (j_{1}, j_{2})}(t)$ is the variance of $Z_{i,(j_{1},j_{2})}$. In the following passage, we omit the filtration $|\mathcal{F}_{i}(h)$ in argument. 
Then we can obtain the probability lower bound of arm $a_{j_{1}}$ is ranked higher than the arm $a_{j_{2}}$ at time step $t$ from the viewpoint of agent $p_{i}$ via the proxy random variable $Z_{i,(j_{1},j_{2})}$, that is
\begin{equation}
    \begin{aligned}
        \mathbb{P}(\widehat{\mu}_{i,j_{1}}(t)> \widehat{\mu}_{i,j_{2}}(t))
        &= \mathbb{P}(\widehat{\mu}_{i,j_{1}}(t)- \widehat{\mu}_{i,j_{2}}(t) >0)\\
        &= \mathbb{P}(\widehat{Z}_{i,(j_{1},j_{2})} >0) \\
        &\geq  \mathbb{P}(Z_{i,(j_{1},j_{2})} >0), \; \text{by the inequality }  \eqref{Eq: normal variance comparison} \\
        &= \mathbb{P}\Bigg(\frac{Z_{i,(j_{1},j_{2})} - \bar{\theta}_{i}\Tra [x_{j_{1}}(t) - x_{j_{2}}(t)]}{\Sigma_{i, (j_{1}, j_{2})}(t)} \geq -\nu_{i,(j_{1},j_{2})}(t)\Bigg),
    \end{aligned}
\end{equation}
where $\nu_{i,(j_{1},j_{2})}(t) = \frac{\bar{\theta}_{i}\Tra [x_{j_{1}}(t) - x_{j_{2}}(t)]}{\Sigma_{i, (j_{1}, j_{2})}(t)}$ greater than zero, is based on the true preference's setting that $a_{j_{1}} >_{i}^{t} a_{j_{2}} >_{i}^{t} a_{j_{3}}$ for agent $p_{i}$ at time $t$.
The aim of the last equality is to transform the proxy random variable to the standard normal variable and quantify the event $\mathcal{A}_{1}(t)$'s probability lower bound.
Now this event $\mathcal{A}_{1}(t)$'s probability lower bound is 
\begin{equation}
\label{eq:good event one step lower bound}
    \begin{aligned}
       \mathbb{P}(\widehat{\mu}_{i,j_{1}}(t)> \widehat{\mu}_{i,j_{2}}(t))
       &\geq 
       1 - \mathbb{P}\Bigg(\frac{Z_{i,(j_{1},j_{2})} - \bar{\theta}_{i}\Tra [x_{j_{1}}(t) - x_{j_{2}}(t)]}{\Sigma_{i, (j_{1}, j_{2})}(t)} \geq \nu_{i,(j_{1},j_{2})}(t)\Bigg) \\
        &\geq 1 - 
        \frac{1}{\nu_{i,(j_{1},j_{2})}(t)} \frac{1}{\sqrt{2\pi}} e^{\bigg(-\frac{\nu^{2}_{i,(j_{1},j_{2})}(t)}{2}\bigg)},
    \end{aligned}
\end{equation}
where the last inequality is by Lemma \ref{Tails of Normal distribution}, which provides the tail probability of the normal distribution since $\nu_{i,(j_{1},j_{2})}(t)$ is positive. With the same technique, we can acquire the event $\mathcal{A}_{2}(t)$'s  probability's lower bound and we also define $\nu_{t,(j_{1},j_{3})} = \frac{\bar{\theta}_{i}\Tra [x_{j_{1}}(t) - x_{j_{3}}(t)]}{\Sigma_{i, (j_{1}, j_{3})}(t)}$, which is greater than zero. Then we have  
\begin{equation}
    \begin{aligned}
        \mathbb{P}(\widehat{\mu}_{i,j_{1}}(t)> \widehat{\mu}_{i,j_{3}}(t)) 
        \geq 
        1 - \frac{1}{\nu_{t,(j_{1},j_{3})}} \frac{1}{\sqrt{2\pi}} e^{\bigg(-\frac{\nu_{t,(j_{1},j_{3})}^2}{2}\bigg)}.
    \end{aligned}
\end{equation}
So the \textit{1st-good event}'s lower bound probability is,
\begin{equation}
    \begin{aligned}
        &\mathbb{P}(\widehat{\mu}_{i,j_{1}}(t)> \max(\widehat{\mu}_{i,j_{2}}(t), \widehat{\mu}_{i,j_{3}}(t)))\\ 
        &\geq 
        1 - \frac{1}{\nu_{i,(j_{1},j_{2})}(t)} \frac{1}{\sqrt{2\pi}} e^{\bigg(-\frac{\nu^{2}_{i,(j_{1},j_{2})}(t)}{2}\bigg)}
        + 
        1 - \frac{1}{\nu_{t,(j_{1},j_{3})}} \frac{1}{\sqrt{2\pi}} e^{\bigg(-\frac{\nu_{t,(j_{1},j_{3})}^2}{2}\bigg)}
        - 
        1\\
        & = 1 -  \frac{1}{\nu_{i,(j_{1},j_{2})}(t)} \frac{1}{\sqrt{2\pi}} e^{\bigg(-\frac{\nu^{2}_{i,(j_{1},j_{2})}(t)}{2}\bigg)} - \frac{1}{\nu_{t,(j_{1},j_{3})}} \frac{1}{\sqrt{2\pi}} e^{\bigg(-\frac{\nu_{t,(j_{1},j_{3})}^2}{2}\bigg)}.
    \end{aligned}
\end{equation}
And the \textit{2nd-good event}'s lower bound probability is,
\begin{equation}
    \begin{aligned}
        \mathbb{P}(\widehat{\mu}_{i,j_{2}}(t)> \widehat{\mu}_{i,j_{3}}(t))\geq 
        1 - \frac{1}{\nu_{i,(j_{2},j_{3})}(t)} \frac{1}{\sqrt{2\pi}} e^{\bigg(-\frac{\nu^{2}_{i,(j_{2},j_{3})}(t)}{2}\bigg)},
    \end{aligned}
\end{equation}
where $\nu_{t,(j_{2},j_{3})} = \frac{\bar{\theta}_{i}\Tra [x_{j_{2}}(t) - x_{j_{3}}(t)]}{\Sigma_{i, (j_{2}, j_{3})}(t)} > 0$ and we define $\Sigma_{i, (j_{2}, j_{3})}(t) = \sigma^{2} [x_{j_{2}}(t)\Tra\M{M}_{i}x_{j_{2}}(t) + x_{j_{3}}(t)\Tra\M{M}_{i}x_{j_{3}}(t)]$.

Here we provide all definitions of  $\Sigma_{i, (j_{1}, j_{2})}(t)$, $\Sigma_{i, (j_{2}, j_{3})}(t)$, and $\Sigma_{i, (j_{1}, j_{3})}(t)$,
\begin{equation}
    \begin{aligned}
        &\nu_{i,(j_{1},j_{2})}(t) = \frac{\bar{\theta}_{i}\Tra [x_{j_{1}}(t) - x_{j_{2}}(t)]}{\Sigma_{i, (j_{1}, j_{2})}(t)},
        \Sigma_{i, (j_{1}, j_{2})}(t)= \sigma^2[x_{j_{1}}(t)\Tra\M{M}_{i} x_{j_{1}}(t) + x_{j_{2}}(t)\Tra\M{M}_{i} x_{j_{2}}(t)]\\
        &\nu_{i,(j_{2},j_{3})}(t) = \frac{\bar{\theta}_{i}\Tra [x_{j_{2}}(t) - x_{j_{3}}(t)]}{\Sigma_{i, (j_{2}, j_{3})}(t)},
        \Sigma_{i, (j_{2}, j_{3})}(t)= \sigma^2[x_{j_{2}}(t)\Tra\M{M}_{i} x_{j_{2}}(t) + x_{j_{3}}(t)\Tra\M{M}_{i} x_{j_{3}}(t)]\\
        &\nu_{i,(j_{1},j_{3})}(t) = \frac{\bar{\theta}_{i}\Tra [x_{j_{1}}(t) - x_{j_{3}}(t)]}{\Sigma_{i, (j_{1}, j_{3})}(t)},
        \Sigma_{i, (j_{1}, j_{3})}(t)= \sigma^2[x_{j_{1}}(t)\Tra\M{M}_{i} x_{j_{1}}(t) + x_{j_{3}}(t)\Tra\M{M}_{i} x_{j_{3}}(t)]
    \end{aligned}
\end{equation}

So the final good event $\mathcal{G}_{i}(t)$'s probability lower bound is  
\begin{equation}
    \begin{aligned}
        \mathbb{P}(\mathcal{G}_{i}(t))
        &\geq 
        1 - \frac{1}{\nu_{i,(j_{1},j_{2})}(t)} \frac{1}{\sqrt{2\pi}} e^{(-\frac{\nu^{2}_{i,(j_{1},j_{2})}(t)}{2})}\\
        & \qquad -
        \frac{1}{\nu_{i,(j_{1},j_{3})}(t)} \frac{1}{\sqrt{2\pi}} e^{(-\frac{\nu^2_{i,(j_{1},j_{3})}(t)}{2})}
        - \frac{1}{\nu_{i,(j_{2},j_{3})}(t)} \frac{1}{\sqrt{2\pi}} e^{(-\frac{\nu^2_{i,(j_{2},j_{3})}(t)}{2})}.
    \end{aligned}
\end{equation}

\subsection{Proof of Lemma \ref{lem: good event: lower bound} - Bad Event}
\label{Appedix: proof of k-1 step good evnet probability}
\textit{Proof.} 
To get the probability lower bound of the bad event $\mathcal{G}_{i}^{c}(t)$, we can obtain the upper bound of the good event $\mathcal{G}_{i}(t)$ probability first. The proof path is similar to the proof of Lemma \ref{lem: good event: lower bound} but with the upper bound of the tail probability of the normal distribution and exists some nuances. We have
\begin{equation}
    \begin{aligned}
        &\mathbb{P}\big(\mathcal{G}_{i}(t)\big)\\
        &= \mathbb{P}\big(\widehat{\mu}_{i,j_{1}}(t)- \widehat{\mu}_{i,j_{2}}(t) > 0,  \widehat{\mu}_{i,j_{2}}(t) -  \widehat{\mu}_{i,j_{3}}(t) > 0\big) \\
        &= \mathbb{P}\big(\widehat{\mu}_{i,j_{1}}(t)- \widehat{\mu}_{i,j_{2}}(t) > 0|\widehat{\mu}_{i,j_{2}}(t) -  \widehat{\mu}_{i,j_{3}}(t) > 0 \big) \mathbb{P}\big(\widehat{\mu}_{i,j_{2}}(t) -  \widehat{\mu}_{i,j_{3}}(t) > 0\big)\\
        &\leq  \mathbb{P}\big(\widehat{\mu}_{i,j_{2}}(t) -  \widehat{\mu}_{i,j_{3}}(t) > 0\big),
    \end{aligned}
\end{equation}
where the last inequality holds because $\mathbb{P}\big(\widehat{\mu}_{i,j_{1}}(t)- \widehat{\mu}_{i,j_{2}}(t) > 0|\widehat{\mu}_{i,j_{2}}(t) -  \widehat{\mu}_{i,j_{3}}(t) > 0 \big) \leq 1$. Similarly we have
\begin{equation}
    \begin{aligned}
        &\mathbb{P}\big(\mathcal{G}_{i}(t)\big)\\
        &= \mathbb{P}\big(\widehat{\mu}_{i,j_{2}}(t)- \widehat{\mu}_{i,j_{3}}(t) > 0|\widehat{\mu}_{i,j_{1}}(t) -  \widehat{\mu}_{i,j_{2}}(t) > 0 \big) \mathbb{P}\big(\widehat{\mu}_{i,j_{1}}(t) -  \widehat{\mu}_{i,j_{2}}(t) > 0\big)\\
        &\leq  \mathbb{P}\big(\widehat{\mu}_{i,j_{1}}(t) -  \widehat{\mu}_{i,j_{2}}(t) > 0\big),
    \end{aligned}
\end{equation}
where the last inequality holds because $\mathbb{P}\big(\widehat{\mu}_{i,j_{2}}(t)- \widehat{\mu}_{i,j_{3}}(t) > 0|\widehat{\mu}_{i,j_{1}}(t) -  \widehat{\mu}_{i,j_{2}}(t) > 0 \big) \leq 1$.

To get the upper bound of $\mathbb{P}\big(\mathcal{G}_{i}(t)\big)$, we need to quantify the maximum value of $\mathbb{P}\big(\widehat{\mu}_{i,j_{1}}(t) -  \widehat{\mu}_{i,j_{2}}(t) > 0\big)$ and $\mathbb{P}\big(\widehat{\mu}_{i,j_{2}}(t) -  \widehat{\mu}_{i,j_{3}}(t) > 0\big)$. Here we use the same definition in Lemma \ref{lem: good event: lower bound}, $\mathcal{A}_{1}(t) =\{\widehat{\mu}_{i,j_{1}}(t) > \widehat{\mu}_{i,j_{2}}(t)\}$ and $\mathcal{B}(t) = \widehat{\mu}_{i,j_{2}}(t) > \widehat{\mu}_{i,j_{3}}(t)$.
In the following, we provide the proof of getting upper bound probability of $\mathcal{B}(t)$ and $\mathcal{A}_{1}(t)$. The proof of getting the probability upper bound of two quantities is similar, so we get the upper bound of
$\mathbb{P}\big(\mathcal{B}(t)\big)$ first. 

Let's use the similar notation defined in Lemma \ref{lem: good event: lower bound}.
\begin{equation}
    \begin{aligned}
        \widehat{Z}_{i,(j_{2},j_{3})} = \widehat{\mu}_{i,j_{2}}(t) - \widehat{\mu}_{i,j_{3}}(t)|\mathcal{F}_{i}(h) \sim N(\bar{\theta}_{i}\Tra [x_{j_{2}}(t) - x_{j_{3}}(t)], \tilde{\Sigma}_{i, (j_{2}, j_{3})}),
    \end{aligned}
\end{equation}
where $\tilde{\Sigma}_{i, (j_{2}, j_{3})} = \sigma^2[x_{j_{2}}(t)\Tra\M{M}_{i} x_{j_{2}}(t) + x_{j_{3}}(t)\Tra\M{M}_{i} x_{j_{3}}(t) - 2x_{j_{2}}(t)\Tra\M{M}_{i}x_{j_{3}}(t)]$, greater than zero,  is the true variance of $\widehat{Z}_{i,(j_{2},j_{3})}$. 
So 
\begin{equation}
    \begin{aligned}
        \mathbb{P}\big(\mathcal{B}(t)\big)
        &= \mathbb{P}\bigg(\widehat{Z}_{i,(j_{2},j_{3})} > 0\bigg)\\
        &= \mathbb{P}\bigg(\frac{\widehat{Z}_{i,(j_{2},j_{3})} - \bar{\theta}_{i}\Tra [x_{j_{2}}(t) - x_{j_{3}}(t)]}{\tilde{\Sigma}_{i, (j_{2}, j_{3})}} \geq -\tilde{\nu}_{i,(j_{2}, j_{3})}(t) \bigg)\\
        & = 1 - \mathbb{P}\bigg(\frac{\widehat{Z}_{i,(j_{2},j_{3})} - \bar{\theta}_{i}\Tra [x_{j_{2}}(t) - x_{j_{3}}(t)]}{\tilde{\Sigma}_{i, (j_{2}, j_{3})}} \geq \tilde{\nu}_{i,(j_{2}, j_{3})}(t) \bigg)
    \end{aligned}
\end{equation}
where the last equality holds by the symmetry property of normal distribution and define
$\tilde{\nu}_{i,(j_{2}, j_{3})}(t) = \frac{\bar{\theta}_{i}\Tra [x_{j_{2}}(t) - x_{j_{3}}(t)]}{\tilde{\Sigma}_{i, (j_{2}, j_{3})}}$, greater than zero.
Besides, $\tilde{\nu}_{i,(j_{1}, j_{2})}(t)$ can be defined similarly,
\begin{equation}
    \begin{aligned}
        &\tilde{\nu}_{i,(j_{1}, j_{2})}(t) = \frac{\bar{\theta}_{i}\Tra [x_{j_{1}}(t) - x_{j_{2}}(t)]}{\tilde{\Sigma}_{i, (j_{1}, j_{2})}(t)},\\
        &\tilde{\Sigma}_{i, (j_{1}, j_{2})}(t) = \sigma^2[x_{j_{1}}(t)\Tra\M{M}_{i} x_{j_{1}}(t) + x_{j_{2}}(t)\Tra\M{M}_{i} x_{j_{2}}(t) - 2x_{j_{1}}(t)\Tra\M{M}_{i}x_{j_{2}}(t)].
    \end{aligned}
\end{equation}

So by the Lemma \ref{Tails of Normal distribution}'s lower bound of normal tail probability, we have the upper bound probability of $\mathcal{B}(t)$, 
\begin{equation}
    \begin{aligned}
        \mathbb{P}\big(\mathcal{B}(t)\big)\leq 
        1 - (\frac{1}{\tilde{\nu}_{i,(j_{2}, j_{3})}(t)} - \frac{1}{\tilde{\nu}^{3}_{i,(j_{2}, j_{3})}(t)})
        e^{\bigg(-\frac{\tilde{\nu}^{2}_{i,(j_{2},j_{3})}(t)}{2}\bigg)}.
    \end{aligned}
\end{equation}
So the similar result can be obtained for $\mathcal{A}_{1}(t)$,
\begin{equation}
    \begin{aligned}
        \mathbb{P}\big(\mathcal{A}_{1}(t)\big)\leq 
        1 - (\frac{1}{\tilde{\nu}_{i,(j_{1}, j_{2})}(t)} - \frac{1}{\tilde{\nu}^{3}_{i,(j_{1}, j_{2})}(t)})
        e^{\bigg(-\frac{\tilde{\nu}^{2}_{i,(j_{1},j_{2})}(t)}{2}\bigg)}.
    \end{aligned}
\end{equation}
Since 
\begin{equation}
    \begin{aligned}
        \mathbb{P}\big(\mathcal{G}_{i}(t)\big)
        &\leq  \max\Big\{\mathbb{P}\big(\widehat{\mu}_{i,j_{2}}(t) -  \widehat{\mu}_{i,j_{3}}(t) > 0\big), \mathbb{P}\big(\widehat{\mu}_{i,j_{1}}(t) -  \widehat{\mu}_{i,j_{2}}(t) > 0\big)\Big\},\\
        &\leq 
        1- \min
        \Big\{ 
        (\frac{1}{\tilde{\nu}_{i,(j_{2}, j_{3})}(t)} - \frac{1}{\tilde{\nu}^{3}_{i,(j_{2}, j_{3})}(t)})
        e^{\bigg(-\frac{\tilde{\nu}^{2}_{i,(j_{2},j_{3})}(t)}{2}\bigg)},
        \\
        &\quad\quad\quad\quad\quad (\frac{1}{\tilde{\nu}_{i,(j_{1}, j_{2})}(t)} - \frac{1}{\tilde{\nu}^{3}_{i,(j_{1}, j_{2})}(t)})
        e^{\bigg(-\frac{\tilde{\nu}^{2}_{i,(j_{1},j_{2})}(t)}{2}\bigg)}
        \Big\}.
    \end{aligned}
\end{equation}
Meanwhile we get the lower bound of $\mathcal{G}_{i}^{c}(t)$ as follows,
\begin{equation}
    \begin{aligned}
        \mathbb{P}\big(\mathcal{G}_{i}^{c}(t)\big)
        &\geq  
        \min
        \Big\{ 
        (\frac{1}{\tilde{\nu}_{i,(j_{2}, j_{3})}(t)} - \frac{1}{\tilde{\nu}^{3}_{i,(j_{2}, j_{3})}(t)})
        e^{\bigg(-\frac{\tilde{\nu}^{2}_{i,(j_{2},j_{3})}(t)}{2}\bigg)},\\
        &\quad\quad\quad
        (\frac{1}{\tilde{\nu}_{i,(j_{1}, j_{2})}(t)} - \frac{1}{\tilde{\nu}^{3}_{i,(j_{1}, j_{2})}(t)})
        e^{\bigg(-\frac{\tilde{\nu}^{2}_{i,(j_{1},j_{2})}(t)}{2}\bigg)}      
        \Big\}
    \end{aligned}
\end{equation}

\subsection{Proof of Lemma \ref{corr: uniform good lower bound and failure lower bound}
}
\label{Appendix: proof of lower bound order}
\textit{Proof.} 
In order to get the good event $\mathcal{G}_{i}(t)$ and bad event $\mathcal{G}_{i}^{c}(t)$'s probability order. We first need to analyze the order of $\nu_{i,(j_{1},j_{2})}(t)$, $\tilde{\nu}_{i,(j_{1},j_{2})}(t)$, $\Sigma_{i,(j_{2}, j_{3})}(t)$, $\tilde{\Sigma}_{i,(j_{2}, j_{3})}(t)$ and other similar terms. 

Based on the definition of $\nu_{i,(j_{1},j_{2})}(t) = \frac{\bar{\theta}_{i}\Tra [x_{j_{1}}(t) - x_{j_{2}}(t)]}{\Sigma_{i,(j_{1},j_{2})}}$, we know that the context difference at time $t$, which is $x_{j_{1}}(t) - x_{j_{2}}(t)$, independent of $h$. 
The expected ridge parameter is $\bar{\theta}_{i} = \big(\M{X}_{i} (h)\Tra \M{X}_{i} (h) + \lambda_{i} \M{I} \big)^{-1} 
\M{X}_{i}(h)\Tra \M{X}_{i}(h)
\theta_{i,*}$. Based on the problem design in Section. \ref{sec: theory-lowerbound}, $\bar{\theta}_{i}$ can be rewritten as $\{\sqrt{1-1/h}\V{c}_{1}+ 1/\sqrt{h}\V{c}_{2}\} = 1/\sqrt{h}\V{c}_{3}$, where $\V{c}_{1}, \V{c}_{2}$ are time-dependent constants based on $(\M{X}_{i} (h)\Tra \M{X}_{i} (h) + \lambda_{i} \M{I} \big)^{-1} \M{X}_{i}(h)\Tra \M{X}_{i}$ but independent of $h$, and $\V{c}_{3}$ is also a context-constant, but independent of $h$. 
Besides, we know that $x_{j_{1}}(t)\Tra\M{M}_{i}x_{j_{1}}(t) \leq \lambda_{\max}(\M{M}_{i})||x_{j_{1}}(t)||^{2}_{2} \leq \lambda_{\max}(\M{M}_{i})L$ by the property $||x_{j_{1}}(t)||^{2}_{2} \leq L$, where $L$ is a constant and we assume $L=1$. So $\lambda_{\max}(\M{M}_{i}) = c_{4}/h$ by Chapter 4 from \citep{vershynin2018high}, where $c_{4}$ can be viewed as a context-constant, independent of $h$. Thus $\nu_{i,(j_{1},j_{2})}(t) = \frac{\V{c}_{3}\Tra  [x_{j_{1}}(t) - x_{j_{2}}(t)]/\sqrt{h}}{c_{4}/h} = c_{5}(t)\sqrt{h}$, where $c_{5}(t)$ is a context-dependent constant, but independent of $h$.

From Lemma \ref{lem: good event: lower bound}, we get the lower bound of of probability $\mathbb{P}(\mathcal{G}_{i}(t))$ such as 
\begin{equation}
    \begin{aligned}
        &\mathbb{P}(\mathcal{G}_{i}(t))\\
        &\geq 
        1 - \frac{1}{\nu_{i,(j_{1},j_{2})}(t)} \frac{1}{\sqrt{2\pi}} e^{\bigg(-\frac{\nu^{2}_{i,(j_{1},j_{2})}(t)}{2}\bigg)}\\
        & \qquad -
        \frac{1}{\nu_{i,(j_{1},j_{3})}(t)} \frac{1}{\sqrt{2\pi}} e^{\bigg(-\frac{\nu^2_{i,(j_{1},j_{3})}(t)}{2}\bigg)}
        - \frac{1}{\nu_{i,(j_{2},j_{3})}(t)} \frac{1}{\sqrt{2\pi}} e^{\bigg(-\frac{\nu^2_{i,(j_{2},j_{3})}(t)}{2}\bigg)}\\
        &= 
        1- 
        3\max \bigg\{\frac{1}{\nu_{i,(j_{1},j_{2})}(t)} \frac{1}{\sqrt{2\pi}} e^{\bigg(-\frac{\nu^{2}_{i,(j_{1},j_{2})}(t)}{2}\bigg)},\\
        &\qquad \frac{1}{\nu_{i,(j_{1},j_{3})}(t)} \frac{1}{\sqrt{2\pi}} e^{\bigg(-\frac{\nu^2_{i,(j_{1},j_{3})}(t)}{2}\bigg)}, 
        \frac{1}{\nu_{i,(j_{2},j_{3})}(t)} \frac{1}{\sqrt{2\pi}} e^{\bigg(-\frac{\nu^2_{i,(j_{2},j_{3})}(t)}{2}\bigg)}
        \bigg\},
    \end{aligned}
\end{equation}
and its corresponding order,
\begin{equation}
    \begin{aligned}
        \mathbb{P}\big(\mathcal{G}_{i}(t)\big) 
         \geq
        \mathcal{L}_{i}^{g}(t),
    \end{aligned}
\end{equation}
where we define $\mathcal{L}_{i}^{g}(t) \overset{\Delta}{=} 1 - \frac{3}{\sqrt{2\pi}} \frac{1}{c_{5}(t)\sqrt{h}}e^{\big(-
\frac{c_{5}^{2}(t)}{2}h\big)}$ as the good event $\mathcal{G}_{i}(t)$'s probability lower bound.

Based on Lemma \ref{lem: good event: lower bound}, we can get the bad event $\mathcal{G}_{i}^{c}(t)$' probability lower bound, which is 
\begin{equation}
    \begin{aligned}
        \mathbb{P}\big(\mathcal{G}_{i}^{c}(t)\big)
        & \geq  
        \min
        \Big\{ 
        (\frac{1}{\tilde{\nu}_{i,(j_{2}, j_{3})}(t)} - \frac{1}{\tilde{\nu}^{3}_{i,(j_{2}, j_{3})}(t)})
        e^{\bigg(-\frac{\tilde{\nu}^{2}_{i,(j_{2},j_{3})}(t)}{2}\bigg)},\\
        &\qquad\qquad (\frac{1}{\tilde{\nu}_{i,(j_{1}, j_{2})}(t)} - \frac{1}{\tilde{\nu}^{3}_{i,(j_{1}, j_{2})}(t)})
        e^{\bigg(-\frac{\tilde{\nu}^{2}_{i,(j_{1},j_{2})}(t)}{2}\bigg)}
        \Big\}
    \end{aligned}
\end{equation}
where $\tilde{\nu}_{i,(j_{1}, j_{2})}(t), \tilde{\nu}_{i,(j_{2}, j_{3})}(t)$ and $\tilde{\Sigma}_{i, (j_{1}, j_{2})}(t), \tilde{\Sigma}_{i, (j_{2}, j_{3})}(t)$ are defined in Lemma \ref{lem: good event: lower bound}.

In addition, we know $\bar{\theta}_{i} = \sqrt{h}\V{c}_{3}$ by the instance design. Since $x_{j_{2}}(t)\Tra\M{M}_{i}x_{j_{2}}(t) \geq \lambda_{\min}(\M{M}_{i})||x_{j_{2}}(t)||^{2}_{2} \geq \lambda_{\min}(\M{M}_{i})c_{\min,j_{2}}(t)$
where $c_{\min,j_{2}}(t) = \underset{t \in [h, T]}{\min}||x_{j_{2}}(t)||^{2}_{2}$ and we assume contexts are meaningful, so $||x_{j_{2}}(t)|| \neq 0$.
Because we know that $\langle x_{j_{2}}(t), x_{j_{3}}(t) \rangle \geq 0$,  $2x_{j_{2}}(t)\Tra\M{M}_{i}x_{j_{3}}(t) \geq 2\lambda_{\min}(\M{M}_{i}) \langle x_{j_{2}}(t), x_{j_{3}}(t) \rangle \geq 2\lambda_{\min}(\M{M}_{i})c_{\min,(j_{2}, j_{3})}(t)$, where $\\ c_{\min,(j_{2}, j_{3})}(t) = \underset{t \in [h+1, T]}{\min}\langle x_{j_{2}}(t), x_{j_{3}}(t)\rangle$.
Then $\tilde{\Sigma}_{i, (j_{2}, j_{3})} \geq 2\sigma^2L\lambda_{\min}(\M{M}_{i}) - 2\sigma^{2}\lambda_{\max}(\M{M}_{i})\\ c_{\min,(j_{2}, j_{3})}(t) = c_{6, (j_{2}, j_{3})}(t)/h$. Thus $\tilde{\nu}_{i,(j_{2}, j_{3})}(t)$ is less $\frac{\V{c}_{3}\Tra  [x_{j_{2}}(t) - x_{j_{3}}(t)]/\sqrt{h}}{c_{6, (j_{2}, j_{3})}(t)/h} \overset{\Delta}{=} c_{7}(t)\sqrt{h}$, where $c_{7}(t)$ is a context-dependent constant, but independent of $h$.

So we get the lower bound order of $\mathbb{P}\big(\mathcal{G}_{i}^{c}(t)\big)$,
\begin{equation}
    \begin{aligned}
       \mathbb{P}\big(\mathcal{G}_{i}^{c}(t)\big)
       &\geq \mathcal{L}_{i}^{b}(t),
    \end{aligned}
\end{equation}
where we define $\mathcal{L}_{i}^{b}(t) \overset{\Delta}{=} (\frac{1}{c_{7}(t)\sqrt{h}} - \frac{1}{c_{7}^{3}(t)h^{3/2}})
e^{\big(-\frac{c_{7}^{2}(t)}{2}h\big)}$ as the bad event $\mathcal{G}_{i}^{c}(t)$'s probability lower bound.

\section{More Simulations}
\label{supp-sec: add sim}

\subsection{Section \ref{sec: ETC vs UCB} Example - Incapable Exploration}
\label{app: ucb vs ts}

We set the true matching reward for three firms to $(0.8, 0.4, 0.2)$, $(0.5, 0.7, 0.2)$, $(0.6, 0.3, 0.65)$. All preferences from companies over workers can be derived from the true matching reward. As we can view, company $p_{3}$ has a similar preference over $a_{1}$ (0.6) and $a_{3}$ (0.65). Thus, the small difference can lead the incapable exploration as described in Section \ref{sec: ETC vs UCB} by the UCB algorithm.

\noindent Next we present the experiment settings of S3, S4, and S5.
\subsection{More Simulation Settings}
\label{sec-app: simulation}
\textit{Scenario 3 (S3)}:
The uniform sub-optimal minimal condition for this scenario is set to be $\overline{\Delta}_{i, \min} = 0.05, \forall i \in [N]$.
The time horizon is set to be $T = 5000$ to have a long enough learning length since we decrease the uniform sub-optimal minimal condition. 
The learning length $h$ for the three noise levels are $h = [264, 876, 4014]$, correspondingly. Thus the difference between S3 and S2 is the time horizon $T$ and different hyperparameters. The data generation process for S3 and S2 are the same.

\textit{Scenario 4 (S4)}:
    The difference between S4 and S1 is that the context dimension changes from $d =2$ to $d =10$.
    The time horizon is set to be $T = 10000$ to accommodate the large dimension.
    Besides, the contextual features $\mu_{j} \in \mathbb{R}^{10}, \forall j \in [3]$, follow similar data generation process as it in S1.
    Here we consider the global preference, i.e, 
    we assume that arms to agents' preference is the global preference, $a_{1}: p_{1}> p_{2},a_{2}: p_{1}> p_{2},a_{3}: p_{1}> p_{2}$. 
    When contexts are noiseless ($\rho = 0$), the true optimal matching is $\{(p_{1}, a_{1}), (p_{2}, a_{2})\}$. 
    The uniform sub-optimal minimal condition for this scenario is set as $\overline{\Delta}_{i, \min} = 0.2, \forall i \in [N]$.
    The learning step length $h$ is 5856. 
    
\textit{Scenario 5 (S5)}:
    The setting in S5 is the same as the setting in S4 except $N=5, K=5$, and $d=5$.
    The time horizon is set to be $T = 15000$ to accommodate the increasing number of participants.
    The uniform sub-optimal minimal condition for this scenario is set to be  $\overline{\Delta}_{i, \min} = 0.1, \forall i \in [N]$.
    The learning step length $h$ is 1975.

\subsection{Additional Simulation Results}
Here we present the experimental analysis of S3 - S5.

\begin{figure}
\centering
\includegraphics[scale=1]{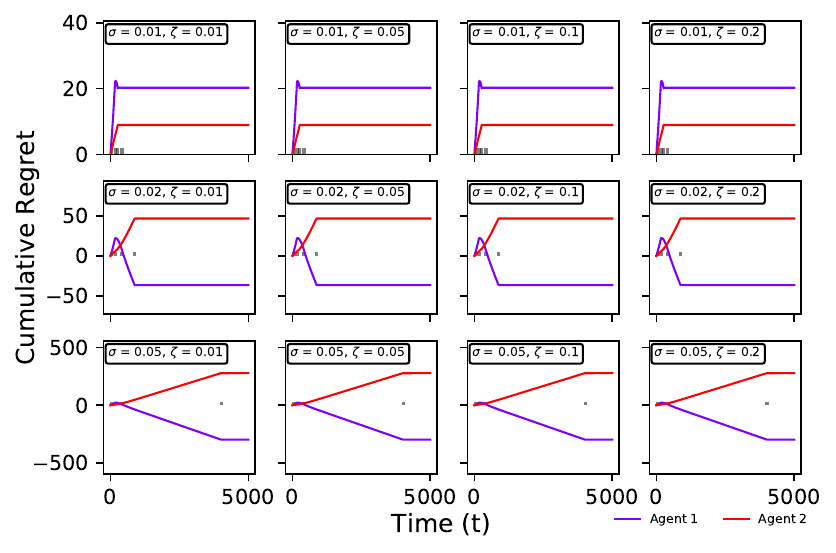}
\caption{Cumulative regret for different noise levels and context variation levels in Scenario S3.}
\label{fig:fig_3_non_global_angle_vector_chi_0.1}
\end{figure}

\textit{Scenario 3 (S3): \CCETC{} is robust to different uniform minimal margin scenarios.}  In Figure \ref{fig:fig_3_non_global_angle_vector_chi_0.1}, we present the result of S3.  
As we change $\overline{\Delta}_{i, \min}=0.2$ in S2 to $\overline{\Delta}_{i, \min}=0.05, \forall i \in [1,2]$, the learning length becomes larger and the estimation becomes better.
Compared with S2's first row and second row, the shaded area in S3's first row and second row becomes narrower, which substantiates our conjecture.
In the second row and third row, we find that agent $p_{1}$ achieves the negative cumulative regret mainly because in the learning step, agent $p_{1}$ is periodically matched with the super-optimal arm with a huge (in absolute value) negative regret. 

\begin{figure}
    \centering
    \includegraphics[scale=0.4]{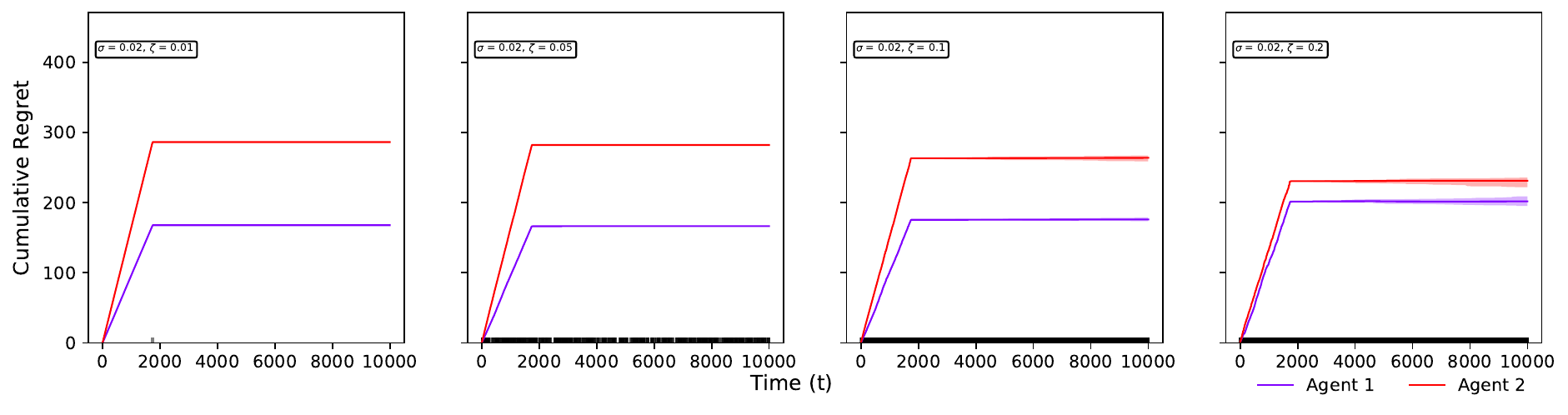}
    \caption{Cumulative regret for different context dimensions in Scenario S4.}
    \label{fig:fig_4_global_d_10}
\end{figure}

\textit{Scenario 4 (S4): \CCETC{} is robust to different dimensions.}  
In Figure \ref{fig:fig_4_global_d_10}, we present the result of S4. Compared with all previous results, we find that when dimension $d$ increases, the regret increases, and the logarithm regret pattern indicates that \CCETC{} is still robust to the dimension.

\begin{figure}
    \centering
    \includegraphics[scale=0.4]{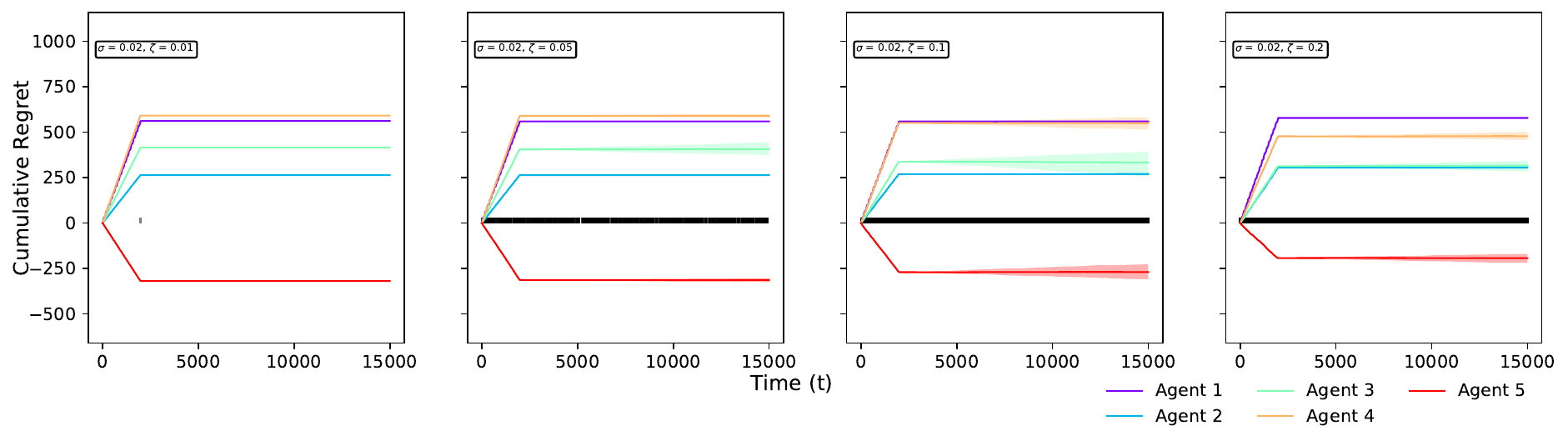}
    \caption{Cumulative regret for different number of agents and arms in Scenario S5.}
    \label{fig:fig_5_multiple_participants}
\end{figure}

\textit{Scenario 5 (S5): \CCETC{} is robust to multiple participants.} 
In Figure \ref{fig:fig_5_multiple_participants}, we present the result of S5, which includes five agents and five arms. Based on the analysis from previous figures and results, \CCETC{} is robust to the choice of preference, context dimension, and context changing format (fixed mean and dynamic mean). Furthermore, we find that dynamic matching is also robust to multiple participants. The cumulative regret still shows the logarithmic shape. 


\subsection{Additional Real Data Result}
\label{sec-app: add real data}
\begin{figure}
    \centering
    \includegraphics[scale=0.4]{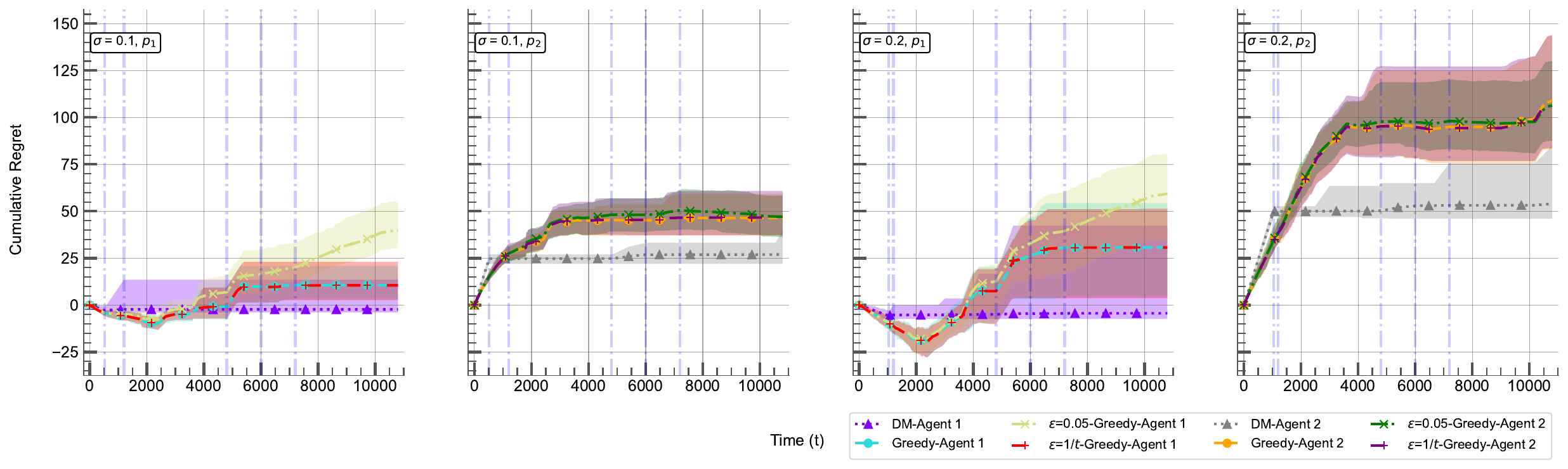}
    \caption{Individual regret for agent $p_{1}$ and $p_{2}$ under noise $ \sigma = 0.1$ (Left two) and $\sigma=0.2$ (Right two) of methods \CCETC{}, greedy, 0.05-greedy, $1/t$-greedy.}
    \label{fig:real-data-1}
\end{figure}

In Figure \ref{fig:real-data-1},
we exhibit the regret of two companies and find that \CCETC{}'s individual regret is superior over all comparison methods under different noise levels for both agents. The shaded area represents the upper and lower bound regret over 100 replications. Lines are used to represent the regret mean over these replications.

\subsection{Textual Information of job applicants and job description}
\label{app-sec: textual}

\begin{table}[!ht]
    \centering\makegapedcells
    \begin{tabular}{|c|c|}\hline
        \textbf{Candidate Profile} & \textbf{Text} \\
        \hline
        \makecell{DS profile} & \makecell{research projects on modeling of high-dimensional and multi-modal \\ (partially observed) inputs for classification, \\ regression and clustering tasks, leveraging a wide range of techniques.}  \\
        \hline
        \makecell{DS update \\ info (18)} &\makecell{
        (1) Strong interested in data science, 
        (2) machine learning,\\ 
        (3) data visualization,  
        (4) data analysis, \\ 
        (5) statistical model, 
        (6) deep learning, \\
        (7) natural language processing,
        (8) coding, \\ 
        (9) options, 
        (10) derivatives, \\
        (11) futures,
        (12) analyze investments, \\ 
        (13) assess risk, 
        (14) assess return profiles, \\
        (15) knowledge in math, 
        (16) statistical model, \\
        (17) programming python,
        (18) R.} \\ \hline
        \makecell{SDE profile} & \makecell{Experienced Software Engineer working at Cisco, skilled in Go, \\ Java, and C++, (1) Working on APIC (Application Policy \\ Infrastructure Controller) and a virtualization project of CMTS \\(Cable Modem Terminal System). (2) Working on a Cloud-native system \\ utilizing containerized microservices using Kubernetes, Docker, etc.}\\ \hline
        \makecell{SDE update \\ info (18)} & \makecell{(1) Algorithms, 
        (2) data structures, \\
        (3) Architecture, 
        (4) Artificial Intelligence, \\
        (5) Machine Learning,
        (6) Compilers, \\
        (7) Database, 
        (8) Distributed Systems,\\ 
        (9) Networking, 
        (10) Systems, \\
        (11) C, 
        (12) C++, \\
        (13) C\#, 
        (14) Java, \\
        (15) JavaScript, 
        (16) go, \\
        (17) Python, 
        (18) objective C.}\\ \hline
        \makecell{Quant profile} & \makecell{Strong passion in quant finance. Well-coordinated skill sets consisting \\ of math, finance, statistics and programming.
        Industrial experiences \\ in equity space including both linear and non-linear products. \\
        Pricing desk quant covering equity exotic derivatives, hybrid derivatives.} \\ \hline
        \makecell{Quant update \\ info (6)} & \makecell{(1) strong math, 
        (2) statistics modeling, \\ 
        (3) Programming, 
        (4) Python,\\
        (5) R, 
        (6) economics.}\\ \hline
        \hline
    \end{tabular}
    \caption{Job applicants' profile}
    \label{table: job applicant profile}
\end{table}

\begin{table}[!ht]
    \centering\makegapedcells
    \begin{tabular}{|c|c|}\hline
        \textbf{Job description} & \textbf{Text} \\
        \hline
        \makecell{Quantitative \\ job description} & \makecell{Strong passion in quant finance, strong mathematical \\ and statistical knowledge. Proficiency in programming languages \\ like Python or R. Data analysis and visualization skills. \\ Understanding of quantitative modeling \\ and statistical methods. Domain-specific knowledge (e.g., finance, \\ economics). know equitable product and derivatives.} \\ \hline
        \makecell{SDE \\ job description} & \makecell{Research experience in Algorithms, Architecture, Artificial Intelligence, \\ Compilers, Database, Data Mining, Distributed Systems, Machine Learning, \\ Networking, or Systems. Programming experience in one or more \\ of the following: C/C++, C\#, Java, JavaScript, Python Objective C, Go, \\ or similar.  Experience in computer science,  with competencies in data \\structures, algorithms and software design.} \\ \hline
    \end{tabular}
    \caption{Job description}
    \label{table: job descript}
\end{table}

\end{document}